%% file: main.tex
\documentclass[onecolumn]{fairmeta}

\usepackage{amsmath,amssymb,amsfonts}%
\usepackage{mathrsfs}%
\usepackage{mathtools}
\usepackage{bbm}

\usepackage{algorithm}%
\usepackage{algorithmicx}%
\usepackage{algpseudocode}%
\usepackage{listings}%

\usepackage{array}
\usepackage{longtable}
\usepackage{pdflscape}
\usepackage{scalerel}
\usepackage{enumitem}
\usepackage{xspace}
\usepackage{soul}
\usepackage{colortbl}
\usepackage{csquotes}
\usepackage{siunitx}
\usepackage{threeparttable}
\usepackage{tabularx,ragged2e}
\usepackage{bbding}
\usepackage{pifont}
\usepackage{rotating}
\usepackage{etoc}
\usepackage{tikz}
\usepackage{wasysym}
\usepackage{makecell}
\usepackage{adjustbox}
\usepackage{tipa}
\usepackage{float}
\usepackage{wrapfig}
\usepackage{nicefrac}
\usepackage[switch]{lineno}

\setcitestyle{numbers,square,comma}

\crefname{section}{Section}{Sections}
\Crefname{section}{Section}{Sections}
\crefname{table}{Table}{Tables}
\Crefname{table}{Table}{Tables}
\crefname{figure}{Fig.}{Figs.}
\Crefname{figure}{Fig.}{Figs.}

\newcommand{\extfigref}[1]{Extended Data Fig.~\ref{#1}}
\newcommand{\exttabref}[1]{Extended Data Table~\ref{#1}}

\newcommand{\suppfigref}[1]{Supplementary Fig.~\ref{#1}}
\newcommand{\supptabref}[1]{Supplementary Table~\ref{#1}}
\newcommand{\suppnoteref}[1]{Supplementary Note~\ref{#1}}

\newcommand{\beginextendeddata}{%
  \setcounter{figure}{0}%
  \setcounter{table}{0}%
  \renewcommand{\figurename}{Extended Data Fig.}%
  \renewcommand{\thefigure}{\arabic{figure}}%
  \renewcommand{\tablename}{Extended Data Table}%
  \renewcommand{\thetable}{\arabic{table}}%
  \crefname{figure}{Extended Data Fig.}{Extended Data Figs.}%
  \Crefname{figure}{Extended Data Fig.}{Extended Data Figs.}%
  \crefname{table}{Extended Data Table}{Extended Data Tables}%
  \Crefname{table}{Extended Data Table}{Extended Data Tables}%
}

\graphicspath{{figures/}}

\definecolor{darkpastelgreen}{rgb}{0.13, 0.55, 0.13}
\definecolor{darkpastelred}{rgb}{0.55, 0.13, 0.13}
\definecolor{mygray}{rgb}{0.85, 0.85, 0.85}
\definecolor{lightpink}{RGB}{255,235,238}
\definecolor{lightgreen}{RGB}{232,245,233}
\definecolor{Watermelon_Red}{RGB}{218, 91, 110}
\definecolor{Watermelon_Green}{RGB}{137, 152, 83}
\definecolor{Teal_Blue}{RGB}{0, 128, 128}
\definecolor{Coral_Red}{RGB}{217, 83, 79}
\definecolor{Cerulean}{RGB}{42, 122, 175}
\definecolor{Royal_Purple}{RGB}{106, 13, 173}
\definecolor{Olive_Drab}{RGB}{107, 142, 53}
\definecolor{our_red}{RGB}{232,157,160}
\definecolor{our_blue}{RGB}{136,206,230}
\definecolor{our_orange}{RGB}{246,200,168}
\definecolor{our_green}{RGB}{178,211,164}
\definecolor{token_blue}{RGB}{84, 120, 140}
\definecolor{green}{HTML}{009000}
\definecolor{red}{HTML}{ea4335}
\definecolor{codegreen}{rgb}{0,0.6,0}
\definecolor{codegray}{rgb}{0.5,0.5,0.5}
\definecolor{codepurple}{rgb}{0.58,0,0.82}
\definecolor{backcolour}{rgb}{0.95,0.95,0.92}
\definecolor{framecolor}{rgb}{0.8,0.8,0.8}
\definecolor{ourdarkblue}{rgb}{0.0, 0.0, 0.55}

\newcommand{\method}{ClinFusion\xspace}

\newlength\savewidth

\newcolumntype{x}[1]{>{\centering\arraybackslash}p{#1pt}}
\newcolumntype{y}[1]{>{\raggedright\arraybackslash}p{#1pt}}
\newcolumntype{z}[1]{>{\raggedleft\arraybackslash}p{#1pt}}
\newcolumntype{Y}{>{\RaggedRight\arraybackslash}X}

\lstdefinestyle{prettyjson}{
    backgroundcolor=\color{backcolour},
    commentstyle=\color{codegreen},
    keywordstyle=\color{blue}\bfseries,
    numberstyle=\tiny\color{codegray},
    stringstyle=\color{codepurple},
    basicstyle=\ttfamily\small,
    breakatwhitespace=false,
    breaklines=true,
    captionpos=b,
    keepspaces=true,
    numbers=left,
    numbersep=8pt,
    showspaces=false,
    showstringspaces=false,
    showtabs=false,
    tabsize=2,
    frame=single,
    frameround=tttt,
    framerule=0.5pt,
    rulecolor=\color{framecolor},
    xleftmargin=15pt,
    xrightmargin=15pt,
    aboveskip=15pt,
    belowskip=15pt,
    columns=flexible,
    escapeinside={(*@}{@*)}
}

\title{\method: A Vision-Centric Multimodal LLM System for Holistic Medical Understanding}

\author[1,3,4,\star]{Hangjie Yuan}
\author[1,3,\star]{Yichen Qian}
\author[1,3,\star]{Zhiwei Tang}
\author[2,3,\star]{Xianzhe Xu}
\author[1,\star]{Sicheng Yang}
\author[1,3]{Lirong Wu}
\author[1,3]{Jinwang Wang}
\author[1,3]{Pengju Wang}
\author[2]{Zhitao Zeng}
\author[2,3]{Yizeng Han}
\author[1,3]{Yan Xing}
\author[2,3]{Shengxuan Luo}
\author[5]{Tao Feng}
\author[6]{Qing Xie}
\author[6]{Weigen Yao}
\author[4]{Yi Yang}
\author[4,7]{Zuozhu Liu}
\author[1,3]{Jiasheng Tang}
\author[8]{Xiaomao Yu}
\author[8]{Shaocheng Wang}
\author[8,\dagger]{Jitao Wang}
\author[8,\dagger]{Jiahong Dong}
\author[2,3,\dagger]{Weihua Chen}
\author[9,10,\dagger]{Feng Xu}
\author[1,\dagger]{Fan Wang}

\affiliation[1]{DAMO Academy, Alibaba Group, Hangzhou, China}
\affiliation[2]{DAMO Academy, Alibaba Group, Beijing, China}
\affiliation[3]{Hupan Laboratory, Hangzhou, China}
\affiliation[4]{College of Computer Science and Technology, Zhejiang University, Hangzhou, China}
\affiliation[5]{Department of Computer Science and Technology, Tsinghua University, Beijing, China}
\affiliation[6]{Department of Radiology, The Affiliated Yangming Hospital of Ningbo University, Yuyao, China}
\affiliation[7]{Zhejiang University-University of Illinois Urbana-Champaign Institute, Zhejiang University, Haining, China}
\affiliation[8]{Hepato-Pancreato-Biliary Center, Beijing Tsinghua Changgung Hospital, School of Clinical Medicine, Tsinghua Medicine, Tsinghua University, Beijing, China}
\affiliation[9]{School of Software, Tsinghua University, Beijing, China}
\affiliation[10]{Beijing National Research Center for Information Science and Technology, Tsinghua University, Beijing, China}

\contribution[\star]{Equal contribution}
\contribution[\dagger]{Corresponding authors}

\abstract{\input{content/00_Abstract}
}

\begin{document}

\maketitle



\input{content/01_Introduction}

\input{content/02_Results}

\input{content/03_Discussion}

\clearpage
\input{content/04_Methodology}

\bibliographystyle{unsrtnat}
\bibliography{references}


\clearpage

\section*{Acknowledgements}

We thank all the board-certified radiologists who participated in the blinded expert evaluation study.

\section*{Author Contributions}

H.Y., Y.Q., Z.T., X.X., P.W., W.C. and S.Y. conceived and designed the study. 
J.W. (Jitao Wang), J.D., W.C., F.X., F.W., Y.Y. and J.T. supervised the project. 
H.Y., Y.Q., Z.T., X.X., Y.X., P.W. and Y.H. designed the algorithm. 
H.Y., X.X., Z.T., Q.X., W.Y., P.W. and Y.X. contributed to data interpretation and visualization. 
H.Y., Y.Q., Z.T., X.X., J.W. (Jinwang Wang), S.Y., Z.L., S.W. and X.Y. contributed to the initial drafting and revision of the manuscript. 
X.X., H.Y., Z.T., L.W., Z.Z. and S.L. conducted the data collection. 
H.Y., Y.Q., S.Y., J.W. (Jinwang Wang) and T.F. participated in data analysis. 
All authors discussed the results and approved the paper.

\section*{Competing Interests}

H.Y., Y.Q., Z.T., X.X., L.W., J.W. (Jinwang Wang), P.W., Y.H., Y.X., S.L., J.T., W.C., and F.W. are employees of DAMO Academy, Alibaba Group. 
The remaining authors declare no competing interests.

\section*{Data Availability}

The public datasets used for model training and evaluation are listed in \exttabref{ext-tab:data_sources_v260} and \exttabref{ext-tab:benchmark_summary}. 
Their download links and access terms are documented in \exttabref{tab:clinfusion_data_access}, and all datasets are available from their respective original sources. 
Data from the SEER cancer registry, used to construct part of the MedIF-Bench evaluation, are available at \url{https://seer.cancer.gov/data/}. 
The knowledge tool draws on publicly available medical knowledge sources, including UMLS (\url{https://www.nlm.nih.gov/research/umls/}), PrimeKG (\url{https://github.com/mims-harvard/PrimeKG}), PubMed (\url{https://pubmed.ncbi.nlm.nih.gov/}), StatPearls (\url{https://www.statpearls.com/}), Wikipedia (\url{https://www.wikipedia.org/}), and standard medical textbooks. 
The in-house knowledge base of clinical guidelines and journal articles, together with the hospital-collected CT data used in this study, are not publicly available because of patient privacy, ethical restrictions, and institutional data-use agreements. 
De-identified research data may be made available for academic and non-commercial purposes upon reasonable request to the corresponding authors, subject to institutional approval and a data-use agreement.

\section*{Code Availability}

The source code for \method, including all evaluation scripts, benchmark implementations, and the proposed MedIF-Bench, is available at \url{https://github.com/alibaba-damo-academy/ClinFusion}.
Model weights for both \method-8B and \method-32B are available at \url{https://huggingface.co/collections/Alibaba-DAMO-Academy/clinfusion}.




\clearpage
\input{content/07_Figures}

\clearpage
\setcounter{page}{1}
\input{content/09_ExtendedData}

\clearpage
\setcounter{page}{1}

\input{content/06_Appendix}

\end{document}

%% file: content/00_Abstract.tex
Multimodal large language models (MLLMs) hold immense potential to revolutionize clinical practice, yet deploying them in the medical domain is fundamentally a vision-centric challenge: 
models must absorb knowledge from heterogeneous 2D and 3D medical images, and 
evaluation protocols must align with radiologists' clinical practice and provide an accurate, fine-grained and factualness-driven assessment.
In this paper, we introduce \method, a vision-centric MLLM designed for holistic medical understanding that systematically addresses these limitations.
We propose a compositional and cascaded vision encoder architecture featuring a Cascade Spatial-Aware Locality Fusion operator that unifies diverse 2D and native 3D medical image understanding within a fused encoder. 
We further introduce a vision-grounded evaluation framework, including MedIF-Bench for instruction-following assessment and a region-of-interest-grounded method for clinically aligned and factualness-driven report generation evaluation.
We show that \method sets a new state-of-the-art across a comprehensive suite of 2D and 3D multimodal medical benchmarks---spanning visual question answering, report generation, and instruction following---as well as textual medical tasks, outperforming leading open-source medical MLLMs (\textit{e.g.}, Hulu-Med, Lingshu) on 20 out of 24 benchmarks and demonstrating multimodal capabilities better than powerful proprietary models such as GPT-5.2 and Gemini-3-Flash on 13 out of 16 benchmarks, and can be further augmented with agentic tool use for retrieval-augmented and tool-assisted clinical workflows.
A blinded evaluation by board-certified radiologists confirms that \method produces the highest-ranked reports, and validates our RoI-grounded metric as achieving the strongest correlation with expert judgment among all automatic evaluation metrics examined.

\textbf{Github}: \url{https://github.com/alibaba-damo-academy/ClinFusion}
\vspace{-.5cm}

\textbf{Huggingface}: \url{https://huggingface.co/collections/Alibaba-DAMO-Academy/clinfusion}

%% file: content/01_Introduction.tex
\section{Introduction}

The recent proliferation of Multimodal Large Language Models (MLLMs) has unlocked unprecedented capabilities in integrating and reasoning over diverse data modalities~\cite{bai2025qwen2.5-vl,qwen3-vl,liu2023llava,hurst2024gpt-4o,liu2024improved-llava}. 
Consequently, the medical domain has emerged as a fertile ground for MLLM-powered intelligent systems, which hold immense potential to revolutionize tasks such as automated diagnostics, clinical report generation, and interactive decision support. 

Because clinical decisions are predominantly driven by visual evidence---from radiographs and pathology slides to volumetric CT and MRI scans---successfully deploying MLLMs in the high-stakes medical field is fundamentally a \textbf{vision-centric} challenge. 
This challenge manifests along two tightly coupled dimensions:
\textbf{1)} \textit{Visual knowledge absorption}: the architecture must natively process and internalize the heterogeneous visual knowledge unique to medicine, spanning diverse 2D and 3D imaging modalities;
and \textbf{2)} \textit{Vision-grounded evaluation}: the evaluation system must go beyond surface-level text matching and instead assess a model's visual understanding through clinically meaningful, region-level analysis that reflects real-world diagnostic practice.
Underpinning both dimensions is a principled data-curation pipeline that provides the diverse, high-quality visual training signal necessary for robust medical perception.
Indeed, recent advancements in Medical MLLMs have made strides in addressing these aspects~\cite{xu2025lingshu,jiang2025hulu-med,sellergren2025medgemma,li2023llava-med,pan2025medvlm-r1,lai2025med-r1,chen2024huatuogpt-vision,zhang2024biomedgpt}. 

Despite promising results on established benchmarks for multimodal~\cite{hu2024omnimedvqa,liu2021slake,hamamci2024CT-Rate} and text-based~\cite{hendrycks2021mmlu,zuo2025medxpertqa,jin2019pubmedqa} medical tasks, critical gaps remain in both vision-centric dimensions identified above---visual knowledge absorption and vision-grounded evaluation---that hinder the transition of MLLMs to in-the-wild clinical practice.
The first critical challenge lies in \textbf{Heterogeneous Visual Knowledge Absorption}:
A single, monolithic vision encoder is often inadequate for capturing the diverse knowledge embedded in heterogeneous medical data. 
Medical imaging is fundamentally multimodal, spanning a wide array of 2D (\textit{e.g.}, X-rays, pathology slides, fundus images) and 3D (\textit{e.g.}, CT and MRI scans) modalities. 
This diversity presents a far greater challenge for visual feature extraction and understanding than that of general-domain applications.
The prevailing approach in medical MLLMs has been predominantly \textit{data-centric}: models such as Lingshu~\cite{xu2025lingshu}, Hulu-Med~\cite{jiang2025hulu-med}, LLaVA-Med~\cite{li2023llava-med}, and HuatuoGPT-Vision~\cite{chen2024huatuogpt-vision} adapt general-purpose vision-language backbones through stage-wise fine-tuning on meticulously curated image-text datasets, yet their underlying vision architectures remain largely unchanged---typically a single, monolithic encoder.
Existing architectural strategies can be broadly categorized into two groups:
a 2D-centric approach that treats all visual data as 2D images by sampling dense slices from 3D volumes~\cite{sun2025holistic_3d_brain_ct,xu2025lingshu,jiang2025hulu-med,sellergren2025medgemma}, which inevitably sacrifices critical 3D structural information; 
a 3D-centric approach that employs representative encoders for 3D data, requiring extensive and complex alignment procedures~\cite{xin2025med3dvlm,wu2025RadFM,pai2024foundation_cancer}.
While multi-encoder designs have been explored in the general vision-language domain~\cite{shi2024eagle,li2025eagle-2,tong2024cambrian-1,li2025mini-gemini,karamcheti2024prismatic_vlms,luo2024Mixture-of-Resolution-Adaptation}, they target multi-resolution inputs or stronger visual acuity rather than the unique challenges of medical imaging. 
The most related work, Cambrian-1~\cite{tong2024cambrian-1}, uses a parallel design to aggregate visual information from multiple vision encoders, but lacks a native 3D encoder and the guided fusion mechanism we propose.

The second critical challenge lies in \textbf{Vision-Grounded Evaluation}.
Current evaluation protocols for medical MLLMs have two critical gaps. 
The first is that existing benchmarks overlook instruction-following capability, which is often degraded by domain-specific fine-tuning yet remains a prerequisite for any meaningful downstream evaluation and clinical deployment.
The second centers on report generation.
Report generation is among the most clinically important applications of medical MLLMs, yet its evaluation remains fundamentally flawed.
A common practice to evaluate report generation capability is by 1) providing a general instruction for MLLMs to generate a comprehensive report and then 2) comparing the generated report against a reference report. 
However, this paradigm is misaligned with clinical practice: radiologists interpret images guided by the patient's clinical context (\textit{e.g.}, referral indication, medical history), which directs their attention to specific anatomical regions and shapes the structure of their reports—yet current protocols force models to produce comprehensive reports without any focus on the patient.
Moreover, existing metrics compare generated reports against references through either surface-level lexical or semantic matching (\textit{e.g.}, BLEU, ROUGE~\cite{lin2004rouge}, METEOR~\cite{banerjee2005METEOR}, CIDEr~\cite{vedantam2015cider}, BERTScore~\cite{Zhang2020BERTScore}), entity-level extraction and matching that relies on rules or small-scale models with limited accuracy (\textit{e.g.}, RadGraph-F1~\cite{delbrouck-etal-2022-improving}, RaTEScore~\cite{zhao2024ratescore}), or holistic LLM-based scoring that conflates accuracy and completeness (\textit{e.g.}, GREEN~\cite{ostmeier2024GREEN})—none of which provide an accurate, fine-grained and factualness-driven decomposition of matched, missed, and hallucinated clinical findings.
Beyond these two core vision-centric challenges, a further practical consideration is that MLLMs are inherently passive and isolated systems: their parametric knowledge is static, their reasoning lacks traceability, and they face perception bottlenecks on specialized tasks where dedicated AI models remain superior~\cite{qiu2024agentic,wange2025baymax,wang2025medagentpro,li2024mmedagent}.

To address these fundamental challenges, we present \method, a vision-centric multimodal large language model system designed for holistic medical understanding. 
As illustrated in~\cref{fig:contributions_overview}, our work introduces a principled framework that systematically tackles each of the aforementioned limitations through a contribution spanning architecture, evaluation, and system-level integration.

First, to overcome the challenge of heterogeneous visual knowledge absorption, we propose a Compositional Vision Encoder. 
Instead of relying on a monolithic encoder, \method integrates a foundational and well-aligned vision transformer with an ensemble of specialist 2D encoders. 
More critically, we introduce Cascade Spatial-Aware Locality (CaSL\footnote{CaSL is pronounced as ``castle'' (UK: \textipa{/`kA:s@l/}, US: \textipa{/`k\ae s@l/}).}) Fusion, a hierarchical operator that enables a cascaded interplay among the 2D encoders, which progressively enriches a foundational visual representation while preserving its alignment.
To natively handle volumetric data, we further incorporate a dedicated 3D encoder and design a 2D-anchored depth-aware CaSL Fusion, where well-aligned 2D feature representations guide the alignment of 3D features. 
In contrast to the parallel aggregation designs of prior work~\cite{tong2024cambrian-1}, \method uniquely incorporates a native 3D vision encoder alongside a compositional ensemble of 2D encoders, coordinated by the CaSL Fusion operator that enables both a cascaded interplay among the 2D encoders for progressive feature enrichment and a unique 2D-anchored interplay between the 2D and 3D modalities.

Second, to bridge the gap between evaluation and clinical practice, we introduce a vision-grounded evaluation framework centered on two key innovations. 
We propose {MedIF-Bench}, a comprehensive benchmark that addresses the critical yet overlooked issue of instruction-following in medical contexts, assessing whether MLLMs can accurately execute complex, practical medical instructions---a prerequisite for reliable clinical interaction. 
More fundamentally, for vision-intensive tasks like report generation, we introduce a {Region-of-Interest (RoI)-grounded evaluation} methodology that aligns generation with clinical practice by conditioning on patient-specific clinical context (\textit{i.e.}, clinically salient anatomical areas) and employs an LLM-as-a-judge to decompose diagnostic claims into matched, missed, and hallucinated findings, which provides an accurate and factualness-driven assessment.

To validate both our model and our evaluation methodology against expert clinical judgment, we conduct a blinded expert evaluation in which six board-certified radiologists independently rank reports generated by \method, Gemini-3-Flash, and Hulu-Med across 300 clinical cases spanning CT and X-ray modalities.
This study serves a dual purpose: it confirms that \method produces the highest-ranked reports across factual accuracy, completeness, and clinical utility, and it demonstrates that our RoI-grounded metric achieves the strongest correlation with expert judgment among all eleven automatic metrics examined---providing direct clinical evidence that our evaluation framework measures what clinicians value.

Furthermore, to enhance the practical deployment of \method in real-world clinical settings, we develop an agentic tool use system that extends the model's capabilities beyond its parametric knowledge. 
This system equips \method with retrieval-augmented generation (RAG) for grounding answers in current, citable medical literature, and a suite of perception expert tools that enable it to invoke specialized AI models for high-precision analysis like organ segmentation or disease classification. 
While the core technical contributions of this work lie in the vision architecture and evaluation framework described above, this agentic extension demonstrates how \method can be practically deployed as an active clinical assistant with dynamic knowledge access and verifiable, tool-augmented reasoning.

In summary, our main contributions are:
\begin{itemize}
    \item \textbf{A compositional and cascaded vision architecture} that natively unifies heterogeneous 2D and 3D medical imaging through CaSL Fusion, a cascaded and 2D-anchored interplay mechanism, and progressively enriches a foundational and well-aligned visual representation.
    

    \item \textbf{A new vision-grounded evaluation framework}, featuring the MedIF-Bench for assessing instruction-following in medical scenarios and an RoI-grounded methodology for more clinically aligned and precise report generation evaluation.

    \item \textbf{Leading performance validated by both automatic benchmarks and expert radiologists}. 
    \method achieves state-of-the-art results across 2D and 3D multimodal medical benchmarks alongside textual medical tasks, outperforming both general-purpose and medical MLLMs, with multimodal performance competitive with leading proprietary models such as GPT-5.2 and Gemini-3-Flash.
    A blinded evaluation by six board-certified radiologists on 300 clinical cases further confirms that \method produces the highest-ranked reports and that our RoI-grounded metric best reflects expert clinical judgment.
    We additionally demonstrate that equipping \method with an agentic tool use system---integrating dynamic knowledge retrieval and external specialist models---yields consistent improvements in both text-only and multimodal clinical scenarios.
\end{itemize}


%% file: content/02_Results.tex
\section{Results}

This section presents the results of our study, beginning with an overview of the proposed framework, \textbf{\method}, and the evaluation settings, followed by a detailed analysis of the main findings.

\subsection{Overview}

\subsubsection{System Overview}

\method is a vision-centric multimodal large language model designed for holistic medical understanding. 
As illustrated in~\cref{fig:architecture}, the system is built upon two core contributions:
\text{1)} a \textbf{compositional and cascaded vision encoder} featuring the CaSL Fusion operator, which unifies the processing of diverse 2D and native 3D medical data by progressively enriching a foundational Qwen ViT representation with features from specialist encoders;
and \text{2)} a \textbf{vision-grounded evaluation framework}, including the MedIF-Bench for assessing instruction-following, and an RoI-grounded methodology for report generation evaluation.
To further enhance its practical deployment, \method is complemented by an \textbf{agentic tool use extension} that equips it with retrieval-augmented generation and the ability to invoke external specialized models as tools, demonstrating consistent improvements in both text-only and multimodal clinical scenarios.
We present results for several variants of our proposed architecture.
Our primary models, \method-8B and \method-32B, are built upon the Qwen3-VL-8B and Qwen3-VL-32B foundations, respectively.


\subsubsection{Evaluation Settings}


\noindent\textbf{Models for comparison.} 
To comprehensively evaluate the performance of \method, we conduct extensive comparisons against three categories of state-of-the-art models:
\begin{itemize}
    \item \textbf{Leading generalist MLLMs}: We include leading open-source MLLMs to establish a general-domain baseline. This includes Qwen2.5-VL series~\cite{bai2025qwen2.5-vl}, Qwen3-VL series~\cite{qwen3-vl} and InternVL3 series~\cite{zhu2025internvl3}.

    \item \textbf{Leading medical MLLMs}: To situate \method within the current medical AI landscape, we compare it against a wide array of specialized medical MLLMs. 
    This includes recent models such as Lingshu~\cite{xu2025lingshu}, Hulu-Med~\cite{jiang2025hulu-med}, MedGemma~\cite{sellergren2025medgemma}, HuatuoGPT-V~\cite{chen2024huatuogpt-vision}, LLaVA-Med~\cite{li2023llava-med}, BioMediX2~\cite{mullappilly2024bimedix2}, HealthGPT~\cite{lin2025healthgpt}, MedDr~\cite{he2024meddr}, BiomedGPT~\cite{zhang2024biomedgpt}, Med-R1~\cite{lai2025med-r1} and MedVLM-R1~\cite{pan2025medvlm-r1}.

    \item \textbf{Leading proprietary models}: We also include results from leading proprietary models, including GPT-5.2, Gemini-3-Flash, Claude-Sonnet-4.5, to benchmark against the current frontier of multimodal AI. 
\end{itemize}

To ensure a fair and reproducible comparison, we utilize a unified evaluation kit for leading open-source models. 
For models where official checkpoints are available, we re-evaluate their performance on our full benchmark suite using this standardized protocol.
For 3D volumetric inputs, we by default sample 32 slices per volume, each resized to $256\times256$, unless a model's official inference code prescribes a different setting, in which case we follow its native configuration (\textit{e.g.}, Hulu-Med~\cite{jiang2025hulu-med}, which processes all slices).
For 2D benchmarks, all available images in each sample are provided to the model without subsampling.


\noindent\textbf{Benchmarking suite.} 
We assess all models across a diverse suite of benchmarks (detailed in~\cref{sec:evaluation_system} and \exttabref{ext-tab:benchmark_summary}) categorized as follows:
\begin{itemize}
    \item \textbf{2D multimodal medical benchmarks:} Standard VQA and report generation tasks on 2D medical images.
    \item \textbf{3D multimodal medical benchmarks:} Tasks requiring understanding of volumetric data like CT/MRI scans.
    \item \textbf{Textual medical benchmarks:} Standard medical question-answering datasets to evaluate the underlying language and knowledge capabilities of the LLM component.
    \item \textbf{Medical instruction following benchmark (our MedIF-Bench):} Our proposed benchmark to specifically measure the model's ability to follow complex, clinically-relevant instructions.
\end{itemize}

\noindent\textbf{Evaluation metrics.} 
For VQA benchmarks (both 2D and 3D), we report accuracy for multiple-choice questions (MCQ) and open-ended questions. 
For report generation benchmarks, we employ our proposed RoI-grounded evaluation methodology (\cref{sec:roi_grounded_report_generation}), reporting Precision, Recall, and F1 scores computed via an LLM-as-a-Judge that assesses clinical correctness within specified anatomical regions. 
For textual medical benchmarks, we report accuracy. 
For MedIF-Bench, we report an Overall Instruction-Following (Overall-IF) score that measures the percentage of responses conforming to the prescribed output format via regex-based verification, isolating format compliance as a prerequisite for reliable clinical deployment.

\subsection{Benchmark Evaluation}

\subsubsection{Evaluation on 2D Multimodal Medical Benchmarks}

\noindent\textbf{Leading VQA capabilities among medical MLLMs, competitive with proprietary models.}
\method demonstrates state-of-the-art performance across a comprehensive suite of 2D medical benchmarks, substantially outperforming both leading open-source models and strong proprietary APIs in report generation and challenging VQA tasks. 
Our evaluation is divided into two key areas: a diverse set of VQA benchmarks and the complex generative task of clinical report generation. 

On VQA tasks, \method consistently ranks first or second across all eight benchmarks, showcasing its robust and versatile visual reasoning capabilities. 
\cref{fig:2d_mm_bench_2d_report}a compares \method against leading proprietary, general-purpose, and medical MLLMs on eight 2D benchmarks (detailed scores in \supptabref{tab:2d_mm_bench}).
Our \method-8B model achieves top scores among all medical MLLMs on seven of the eight benchmarks, including the challenging multi-modality datasets that require a comprehensive understanding of all visual modalities and the reasoning-focused MedXpertQA. 
When compared to proprietary models, \method-8B and \method-32B outperform the powerful Gemini-3-Flash on six benchmarks (OmniMedVQA, PMC-VQA, MedFrameQA, VQA-RAD, SLAKE, PathVQA), demonstrating that a specialized, well-designed open-source model can surpass even the most advanced generalist APIs on domain-specific tasks. 
For extremely hard reasoning benchmarks like MedXpertQA, we observe a margin between our model and the proprietary model. 
We anticipate that this gap can be closed through scaling the model since we observe a clear performance boost by scaling \method-8B to \method-32B.

\noindent\textbf{Leading 2D clinical report generation capabilities among medical MLLMs and proprietary models.}
In clinical report generation, \method also sets a new state of the art. 
As shown in~\cref{fig:2d_mm_bench_2d_report}b (detailed scores in~\supptabref{tab:2d_report_gen_bench}), our \method-8B achieves an F1 score of 37.8 on CheXpert-Plus and a remarkable 57.3 on IU-XRAY. 
This represents a substantial improvement over the previous small open-source medical MLLM, Hulu-Med-7B (31.9 and 46.5, respectively), showcasing our model's superior ability to generate clinically accurate and coherent text using the fundamentally enhanced vision capabilities.
When compared with the proprietary models, all versions of our model are on the same level or better than these proprietary models, especially on the CheXpert-Plus dataset, which \method clearly outperforms them.

A qualitative example in~\extfigref{ext-fig:2d-3d-report-case}a further illustrates \method's 2D diagnostic capability.
Despite the clear presence of cardiomegaly and pulmonary vascular prominence in the chest X-ray,multiple strong baselines---including Hulu-Med, Lingshu, and Gemini-3-Flash---erroneously generate ``normal'' reports, likely due to over-reliance on global features.
In contrast, \method accurately identifies both pathologies, correctly describing the ``cardiac silhouette'' as ``mildly enlarged'' and noting the ``mild prominence of the pulmonary vasculature''---a fine-grained perception enabled by its compositional vision encoder architecture.


\subsubsection{Evaluation on 3D Multimodal Medical Benchmarks}

\noindent\textbf{Strong 3D multimodal VQA capabilities among medical MLLMs and proprietary models.}
\method achieves the best or near-best results in native 3D medical understanding across volumetric VQA benchmarks, as shown in~\cref{fig:3d_mm_bench_3d_report_medifbench}a (detailed scores in~\supptabref{tab:3d_mm_bench}).
This performance stems from our vision-centric architecture, featuring a dedicated 3D encoder and the 2D-anchored CaSL Fusion mechanism for volumetric reasoning.

Our \method-8B model, despite its modest size, substantially outperforms much larger models.
On AMOS-MCQ, it achieves 80.2, surpassing the previous best 7B model, Hulu-Med-7B (65.7), by 14.5 points, and even exceeding the 32B Hulu-Med (73.9).
When compared to proprietary APIs, \method-8B outperforms Gemini-3-Flash by 16 points on AMOS-MCQ.
Our \method-32B model further extends this advantage, achieving 81.7 on AMOS-MCQ and 89.0 on CT-Rate MCQ, while remaining competitive with Hulu-Med-32B on 3D-RAD.

\noindent\textbf{Strong 3D report generation capabilities among medical MLLMs and proprietary models.}
\cref{fig:3d_mm_bench_3d_report_medifbench}b reports Precision, Recall and F1 on 3D report generation under the RoI-Grounded protocol (detailed scores in~\supptabref{tab:3d_mm_bench}).
On CT-Rate Report, \method-32B achieves the best F1 (23.9) across all evaluated models, surpassing the strongest medical baseline Hulu-Med-32B (23.4) and all proprietary APIs (Gemini-3-Flash 20.2; GPT-5.2 14.1).
On AMOS Report, \method-32B obtains the best F1 (16.1) among open-source medical MLLMs, while Gemini-3-Flash (23.4) leads overall, indicating headroom on CT reporting that we leave to future work.
At the smaller scale, \method-8B (F1: 12.0 / 21.6) consistently outperforms its 7B medical counterpart Hulu-Med-7B (10.7 / 18.3) on both datasets.

Beyond quantitative metrics, a qualitative analysis on the native CT case in~\extfigref{ext-fig:2d-3d-report-case}b further illustrates how \method excels in both diagnostic accuracy and clinical workflow adherence.
For \textit{diagnostic accuracy}, while the proprietary Gemini-3-Flash generates a generic report incorrectly stating the liver is ``normal in size and attenuation'' and missing the key diagnosis---a common failure of generalist MLLMs on subtle, diffuse density changes in volumetric data---\method correctly identifies the decreased liver density and reaches the final ``Impression: Mild fatty liver''.
For \textit{clinical workflow adherence}, \method's report is meticulously structured according to the requested ``Areas of Focus'', systematically evaluating each organ system.
It further provides specific pertinent negatives, such as ``The intrahepatic and extrahepatic bile ducts are not dilated'' and ``No abnormalities are observed in the gallbladder'', rather than the ground-truth (GT) report's general statement that ``upper abdominal organs are normal''.
This combination of pinpointing pathology and systematically confirming normalcy yields reports that are accurate, structured, and clinically coherent.

\subsubsection{Evaluation on MedIF-Bench}

\noindent\textbf{Strong instruction-following preserved through medical adaptation.}
To address the critical gap between benchmark performance and practical utility, we evaluated models on our newly proposed MedIF-Bench, which is designed to test complex clinical instructions.
As shown in~\cref{fig:3d_mm_bench_3d_report_medifbench}c (detailed scores in~\supptabref{tab:medif_bench}), \method achieves Overall-IF scores of 98.1 (8B) and 98.9 (32B), the best results among medical MLLMs and surpassing all proprietary systems including GPT-5.2 (96.0), Gemini-3-Flash (96.6), and Claude-Sonnet-4.5 (86.5).
More importantly, while general-purpose MLLMs such as Qwen2.5-VL and Qwen3-VL exhibit strong instruction-following, existing medical adaptations of these backbones (e.g., Lingshu-7B: 80.4; Lingshu-32B: 82.6) suffer substantial degradation.
\method is a medical-adapted model that retains near-backbone instruction-following capability,with consistently high scores across diverse tasks (detailed scores in~\supptabref{tab:medif_bench}) including challenging ones such as ``Seer'' (prognosis based on specific conditions) and structured report generation where most other medical models fail.

\noindent\textbf{Instruction-following ability is a prerequisite for reliable evaluation.}
MedIF-Bench also serves as a proxy for model ``evaluability'': a model's inability to adhere to instructional formats leads to inaccurate assessments of its actual knowledge.
The case of MedGemma-1.5-4B-IT illustrates this---it achieves an Overall-IF of only 27.4, and manual inspection confirms that it consistently fails to produce answers in prescribed formats, rendering rule-based and LLM-as-a-judge evaluation unreliable.
This deficiency directly manifests as artificially low scores on other benchmarks (e.g., PMC-VQA: 18.7, \supptabref{tab:2d_mm_bench}; SuperGPQA: 2.25, \supptabref{tab:text_benchmarks}), demonstrating that instruction-following is a prerequisite for evaluating the latent knowledge of any MLLM.
Notably, MedGemma-1.5-4B-IT is trained on diverse medical data yet lacks IF capability, suggesting that data curation alone is insufficient if the training methodology does not explicitly instill robust instruction-following.

\subsubsection{Evaluation on Textual Medical Benchmarks}

\noindent\textbf{Vision-centric architecture need not sacrifice textual competence.}
Despite being a vision-centric architecture, \method's textual abilities match or surpass the best medical MLLMs under unified evaluation, as shown in~\extfigref{ext-fig:text_benchmarks} (detailed scores in~\supptabref{tab:text_benchmarks}).
Our \method-32B achieves the best scores among all medical MLLMs on seven of eight benchmarks, with clear improvements over Hulu-Med-32B on MedXpertQA (26.7 vs.\ 19.8), Medbullets (74.8 vs.\ 67.5), and MedQA-MCMLE (93.8 vs.\ 87.0).
At the smaller scale, \method-8B is competitive with Hulu-Med-7B, achieving the best scores on PubMedQA (77.6), MedQA-MCMLE (89.1), and MMLU-Med (81.3) while trailing on MedMCQA and MedQA-USMLE.
When compared to general-purpose MLLMs, \method consistently outperforms its corresponding backbones, demonstrating that medical adaptation enhances rather than compromises textual knowledge.
We note a remaining gap with the strongest proprietary models, particularly on complex reasoning benchmarks such as MedXpertQA and SuperGPQA, though this gap narrows substantially when scaling from 8B to 32B (e.g., MedXpertQA: 20.0 → 26.7; SuperGPQA: 32.6 → 41.8), suggesting that further scaling can close this gap.

\subsection{Clinical Validation with Expert Radiologists} 

To complement the automatic evaluation presented above, and following methodological precedents for expert validation of AI-generated reports~\cite{tanno2025flamingo_cxr}, we conducted a blinded expert evaluation with six board-certified radiologists (each with $\geq$6 years of clinical experience or longer). 
We randomly sampled 300 cases from our report generation benchmarks (100 Chest X-ray from~\cite{chambon2024CheXpertPlus}, 100 Chest CT from 3D-RAD~\cite{gai2025_3D-RAD} and 100 Abdominal CT from AMOS-MM~\cite{ji2022AMOS}).
Then we collected reports from four systems corresponding to these cases: \method with agentic tools, \method (standalone), Gemini-3-Flash, and Hulu-Med.
For each case, the evaluators examined the original medical image alongside the four anonymized model outputs, then ranked all four reports along three clinically grounded dimensions: 
\emph{Factual Accuracy} (how few hallucinated findings), 
\emph{Completeness} (how few missed abnormalities), and 
\emph{Clinical Utility} (overall value for downstream decision-making).

\noindent\textbf{Experts validate \method's clinical advantage and the benefit of agentic augmentation.}
\cref{fig:anno_expert_eval}a summarizes the aggregated Borda scores across all annotators.
\method with agentic tools ranks first in every dimension---Overall, Accuracy, Completeness, and Operability---with a statistically significant margin over Hulu-Med ($p < 0.001$) and Gemini-3-Flash ($p < 0.001$).
Notably, the standalone \method already outperforms both Gemini-3-Flash and Hulu-Med, and adding the agentic tools yields a further consistent uplift, corroborating the quantitative gains reported in~\supptabref{tab:2d_report_gen_bench}, \supptabref{tab:text_benchmarks}, and \supptabref{tab:3d_mm_bench}.

\noindent\textbf{The RoI-grounded report generation metric achieves the highest correlation with expert judgment.}
We assessed how faithfully each automatic metric reflects the clinical quality perceived by radiologists, as shown in~\cref{fig:anno_expert_eval}b.
Among eleven metrics---including METEOR~\cite{banerjee2005METEOR}, BLEU-4~\cite{papineni2002bleu}, ROUGE-L~\cite{lin2004rouge}, RadGraph-F1~\cite{delbrouck-etal-2024-radgraph}, BERTScore~\cite{Zhang2020BERTScore}, CIDEr~\cite{vedantam2015cider}, and GREEN~\cite{ostmeier2024GREEN}---our proposed metric achieves the highest rank correlation with expert consensus on all three agreement measures (Kendall's $\tau = 0.511$, Spearman's $\rho = 0.572$, Top-1 Accuracy $= 55.5\%$).
At the individual-case level, as shown in~\cref{fig:anno_expert_eval}c, the per-case Kendall's $\tau$ distribution of our metric exhibits a higher mean and lower variance than all alternatives, indicating that its advantage is systematic rather than driven by outliers.

\noindent\textbf{Inter-annotator agreement confirms the reliability of the evaluation.}
To verify the scientific rigor of the annotation study, we assessed inter-annotator agreement using Kendall's coefficient of concordance ($W$). 
As shown in~\cref{fig:anno_expert_eval}d, the overall $W$ across all dimensions is $0.665 \pm 0.275$ ($n = 300$), indicating substantial agreement among the radiologists.
As shown in~\cref{fig:anno_expert_eval}e, agreement is consistent across modalities (CT: $0.678 \pm 0.260$; X-ray: $0.641 \pm 0.300$) and across individual dimensions, with Accuracy reaching $W = 0.687 \pm 0.272$.
These values confirm that the expert rankings are reproducible and that the observed model differences reflect genuine quality gaps rather than annotator noise.

\subsection{Design Analysis}
This section validates our two core design contributions: the RoI-grounded evaluation methodology and the compositional vision architecture.

\subsubsection{Analysis of RoI-Grounded Report Generation}
\label{sec:roi_rg_case}

To validate the design choices behind our RoI-Grounded Report Generation evaluation---which achieves the highest correlation with expert judgment among eleven metrics (\cref{fig:anno_expert_eval}b)---we present a detailed analysis comparing our methodology against RadGraph-F1~\cite{delbrouck-etal-2024-radgraph}, a widely adopted clinical metric for report generation, as employed in recent work~\cite{sellergren2025medgemma}.
We identify two key limitations of this conventional approach and show how our protocol addresses them.


\noindent\textbf{Clinical context ensures meaningful report generation comparisons.}
In a typical prior evaluation setup, such as the one employed by MedGemma~\cite{sellergren2025medgemma}, the model is prompted to generate a report from an image without any clinical context.
To demonstrate why this is problematic, we ablate within our RoI-Grounded pipeline (\cref{sec:roi_grounded_report_generation}) on whether the extracted clinical context (``Clinical Indication'' and ``Area of Focus'') is provided during generation.
We compare two models (\method-8B and Lingshu-7B~\cite{xu2025lingshu}) across report generation benchmarks.

As shown in~\cref{fig:roi_report_ablation_architecture_ablation}a (detailed scores in~\supptabref{tab:context_ablation}), the context-free setting reveals two issues: 
\textbf{1)} metrics drop substantially for both models, and 
\textbf{2)} the performance gap between the two models is severely compressed---under the context-aware setting \method-8B consistently outperforms Lingshu-7B, yet under the context-free setting both produce nearly identical scores.
The root cause is that a single medical image contains many anatomical structures, but a radiologist's report sometimes focuses only on areas relevant to a specific clinical indication.
Without this context, the model may describe regions absent from the GT report, and there is no way to determine whether such claims are hallucinations or legitimate findings that the original radiologist chose not to document---yet all are penalized as false positives, leading to low report generation scores and narrow performance gaps.
By introducing the extracted clinical context, our RoI-Grounded protocol constrains the model to the anatomical areas covered by the GT, ensuring that every claim can be categorized as a True Positive, False Negative, or False Positive.

A representative case is shown in~\extfigref{ext-fig:case_without_context}: 
With clinical context (left), \method's output stays within the anatomical scope of the GT (pleural effusion, pericardial effusion, lung consolidations consistent with Covid-19 pneumonia, and liver findings), so each claim is verifiable.
Without context (middle), the same model produces additional findings on regions the GT never addresses---\textit{e.g.}, mediastinal shift, cardiomegaly, splenomegaly, anasarca, osseous lesions, and a diagnosis of decompensated congestive heart failure---which cannot be distinguished from hallucinations and are uniformly penalized as false positives, deflating the score.

\noindent\textbf{LLM-as-a-Judge evaluation overcomes the synonymy insensitivity and length bias.}
As an alternative to text-matching metrics such as RadGraph-F1~\cite{delbrouck-etal-2022-improving,sellergren2025medgemma}, our RoI-Grounded protocol adopts an LLM-as-a-Judge for report evaluation.
RadGraph-F1 first parses both the predicted and GT reports through RadGraph-XL~\cite{delbrouck-etal-2024-radgraph}---a transformer trained to extract clinical entities (\textit{e.g.}, anatomical structures, observations) and their relations (\textit{e.g.}, \textit{suggestive of}, \textit{located at}) from radiology text---and then computes an F1 score over the matched entity and relation sets.
Critically, RadGraph-XL operates in a \emph{closed-form} fashion: it maps input text to a fixed,pre-defined schema, unlike LLM-based evaluators that accept and produce open-form natural language;furthermore, as a relatively small extractor, it has limited accuracy on out-of-distribution clinical phrasings.
This rigidity enables an LLM-free, computationally cheap pipeline, but introduces two failure modes.
\textbf{1) Synonymy insensitivity:} Semantically equivalent but differently worded findings receive no credit---for example, \textit{``the liver appears enlarged''} fails to match \textit{``hepatomegaly is present''} with a score of zero, despite being clinically identical, as they map to different schema entries with no token overlap.
\textbf{2) Length bias:} Because RadGraph-XL extracts entities from the \emph{entire} input text, longer outputs have a greater statistical chance of overlapping with the GT entities, so even purely verbose language that casually mentions anatomical terms can inflate the score without reflecting clinical accuracy.

\cref{fig:roi_report_ablation_architecture_ablation}b (detailed scores in~\supptabref{tab:radgraph_failure}) provides quantitative evidence of both failure modes.
First, RadGraph-F1 scores cluster within a narrow range across the first three models (11.8--21.0 on CheXpert-Plus; 16.3--36.6 on IU-XRAY), making it difficult to distinguish model quality, whereas our LLM-as-a-Judge spans a wider and more discriminative range (16.8--37.8 on CheXpert-Plus; 36.7--57.3 on IU-XRAY).
Second, the Long Output variant substantially inflates RadGraph-F1 (11.8$\rightarrow$18.8 on CheXpert-Plus; 16.3$\rightarrow$33.0 on IU-XRAY) despite producing lower-quality outputs, while our LLM-as-a-Judge correctly penalizes verbosity (16.8$\rightarrow$15.1 on CheXpert-Plus; 36.7$\rightarrow$33.6 on IU-XRAY).

\subsubsection{Ablation Studies on Architectural Design}

All ablation studies on architectural designs (\textit{e.g.}, the contribution of different encoders or fusion strategies) are conducted on the 2D and 3D multimodal medical benchmarks, as these tasks are most sensitive to changes in visual perception.
To enable rapid iteration, we train on a subset of the full training data and, given the high computational cost of 3D evaluation, also sample a subset of the 3D evaluation data.  
We ensure that every comparison within the same table or figure is performed under identical settings.



\noindent\textbf{Local cross-attention is a better fusion mechanism than other alternatives.}
We compare our proposed CaSL Fusion operator against three alternatives:
\textbf{1)} channel-wise concatenation, which concatenates feature maps along the channel dimension;
\textbf{2)} global cross-attention, where each query token attends to the entirety of the key/value feature map, in contrast to our localized approach; and
\textbf{3)} a mixture-of-vision-experts approach, which computes a token-wise weighted combination of features via a learned gating mechanism~\cite{fedus2022Switch_transformers}.
The model is configured with Qwen ViT as the foundational encoder and DINOv2 as the additional vision encoder.
To facilitate efficient validation, the model is trained for the first three training stages on a subset of the full corpus.
As shown in~\cref{fig:roi_report_ablation_architecture_ablation}c (detailed scores in~\supptabref{tab:ablation_fusion_mechanism}), all evaluated fusion strategies improve over the single-encoder baseline (Qwen ViT only), supporting our hypothesis that a well-aligned foundational encoder can be enhanced by complementary features from a specialist model.
Among the four mechanisms, local cross-attention attains the highest average performance, and we adopt it as the default in all subsequent ablation studies.

\noindent\textbf{DINOv2 augments the Qwen ViT the most.}
We next identify the most effective single specialist encoder to augment the foundational Qwen ViT, comparing DINOv2, ConvNext, and Medsiglip.
As shown in~\cref{fig:roi_report_ablation_architecture_ablation}d (detailed scores in~\supptabref{tab:ablation_one_additional_encoder}), no single encoder dominates both tasks: Medsiglip leads on 2D VQA (64.8 vs.\ 64.5 / 64.1) while ConvNext slightly leads on report generation (27.9 vs.\ 27.8 / 26.6).
However, DINOv2 is the only encoder that ranks within $0.3$ of the best on both tasks, while ConvNext underperforms on VQA and Medsiglip lags on report generation.
We hypothesize that DINOv2's self-supervised pre-training yields more general-purpose visual features that transfer well across both task types, rather than specializing in either.
We therefore adopt DINOv2 as the specialist encoder in all subsequent multi-encoder experiments.


\noindent\textbf{Additional vision encoders complement Qwen ViT, with diminishing returns beyond one additional encoder.}
Building upon our finding that DINOv2 serves as the most effective single specialist encoder in addition to Qwen ViT, we further explore the optimal combinations of two additional encoders, evaluating ensembles of DINOv2 paired with either ConvNeXt or Medsiglip under different cascade orders.
As shown in~\cref{fig:roi_report_ablation_architecture_ablation}e (detailed scores in~\supptabref{tab:ablation_multi_additional_encoder}), the DINOv2$\rightarrow$Medsiglip cascade attains the highest average VQA score (65.2), while the DINOv2$\rightarrow$ConvNeXt cascade attains the highest average score on report generation (28.8).
Compared with the best single-encoder configuration (\cref{fig:roi_report_ablation_architecture_ablation}d), the second specialist encoder adds at most $0.4$ on VQA and $0.9$ on report generation, indicating diminishing returns beyond one additional specialist encoder.
Since report generation places higher demands on visual perception---requiring both global understanding and localized findings---we adopt the Qwen ViT + DINOv2 + ConvNeXt combination as the specialist ensemble in our final \method architecture.

\noindent\textbf{Cascade fusion outperforms parallel fusion on report generation.}
We additionally compare two ways of combining the same DINOv2 + ConvNeXt encoder pair: our Cascade fusion (DINOv2$\rightarrow$ConvNeXt) and a Parallel fusion in which each specialist encoder is independently fused into the Qwen ViT query.
As shown in~\cref{fig:roi_report_ablation_architecture_ablation}e (detailed scores
in~\supptabref{tab:ablation_multi_additional_encoder}), the two strategies perform similarly on 2D VQA (64.4 vs.\ 64.6), but Cascade fusion clearly outperforms Parallel fusion on report generation (28.8 vs.\ 27.3).
We hypothesize that the cascaded design allows the representation to be progressively enriched across specialist encoders, which benefits the open-ended generation setting more than the closed-form VQA setting.
We therefore use Cascade fusion as the default in our final architecture.



\noindent\textbf{The native 3D encoder is essential for 3D volumetric understanding.}
We quantify the contribution of our dedicated 3D perception module by deactivating the 3D encoder at inference time, forcing the model to rely solely on sparsely sampled 2D slices.
As shown in~\cref{fig:roi_report_ablation_architecture_ablation}f (detailed scores in~\supptabref{tab:ablation_3d_encoder}), removing the 3D encoder degrades performance consistently across all evaluated tasks: the average 3D VQA score drops by 3.0 points (65.8 $\rightarrow$ 62.8), and the average 3D report-generation F1 drops by 4.0 points (15.7 $\rightarrow$ 11.7), with the largest single-task degradations on CT-Rate Open VQA (63.6 $\rightarrow$ 56.4) and CT-Rate Report F1 (20.3 $\rightarrow$ 15.5).
These results indicate that volumetric features extracted by the native 3D encoder cannot be substituted by sparse 2D slices alone, supporting the inclusion of a dedicated 3D perception module in \method.


\subsection{Practical Extension with Agentic Tool Use}
We extend \method with an agentic tool-use framework that combines a RAG tool and perception expert tools within a plan--act workflow.
We evaluate this extension on text-only medical benchmarks and 2D/3D multimodal benchmarks at both the 8B and 32B scales, and illustrate the workflow on representative text-only and multimodal cases.

\noindent\textbf{Agentic tools improve 2D and 3D multimodal evaluations, with the largest gains on descriptive tasks.}
On multimodal benchmarks (\cref{fig:agent_improvement}, left column; detailed scores in~\supptabref{tab:2d_report_gen_bench} and~\supptabref{tab:3d_mm_bench}), agentic tools improve \method across both 2D report generation and 3D VQA / report tasks at both scales.                                   
For \method-8B, the largest gains appear on tasks requiring descriptive output, including CheXpert-Plus 2D report (+12.4), 3D-RAD Open VQA (+8.4), AMOS Report (+4.6), and CT-Rate Report (+3.7), while gains on closed-form 3D MCQ tasks are smaller (AMOS MCQ +5.5, 3D-RAD MCQ +2.2, CT-Rate MCQ +2.8).                     
\method-32B follows a similar pattern, with the largest gains again concentrated on descriptive-output tasks (3D-RAD Open VQA +9.4, CheXpert-Plus +6.8, CT-Rate Report +3.7, AMOS Report +3.0) rather than on closed-form MCQ tasks tasks that selects among predefined options.

\noindent\textbf{Agentic tools improve text-only evaluations, with substantial gains on knowledge-intensive benchmarks.}
On text-only medical benchmarks (\cref{fig:agent_improvement}, right column; detailed scores in~\supptabref{tab:text_benchmarks}), agentic tools improve \method consistently at both scales.
For \method-8B, the largest gains appear on the most knowledge-intensive benchmarks, including SuperGPQA (+12.8), MedQA-U (+11.1), MedXpertQA (+7.7), and MedMCQA (+7.3); the only exception is MedQA-M, where the score drops by 4.8 points.
We attribute this to the limited Chinese-language coverage in our retrieval corpus, which restricts the utility of RAG on a Chinese-language exam benchmark.
For \method-32B, gains follow a similar pattern (SuperGPQA +9.1, MedXpertQA +4.5, MedQA-U +4.2), and the MedQA-M deficit no longer appears (+0.7), consistent with the stronger multilingual capability of the larger backbone, which makes better use of the predominantly English retrieved evidence.

\noindent\textbf{The agentic framework makes model decisions verifiable through tool-produced evidence.}
We illustrate the agentic workflow (shown in~\extfigref{ext-fig:agent}c, detailed in~\cref{sec:agentic_tool_use}) on two representative cases (\extfigref{ext-fig:agent}a,b).
In the provided text-only case (\extfigref{ext-fig:agent}a), \method decomposes the clinical question into complementary retrieval queries, invokes the hybrid retrieval engine to collect provenance-aware evidence from heterogeneous sources (\textit{e.g.}, PubMed, textbooks, StatPearls), summarizes the evidence into a traceable intermediate conclusion before generating the final answer.
In the provided multimodal case (\extfigref{ext-fig:agent}b), \method identifies the input as a 3D chest CT volume, invokes a Multi-Organ Lesion Segmenter that returns structured findings (\textit{e.g.}, lesion count and size), and integrates these verified outputs into a radiology-style report with \textit{Findings} and \textit{Impression} sections.
In both scenarios, the intermediate tool outputs remain exposed and auditable, making the reasoning chain verifiable rather than relying solely on implicit parametric generation.

%% file: content/03_Discussion.tex
\section{Discussion}

The integration of MLLMs into clinical practice holds significant promise, yet this transition has been impeded by fundamental obstacles: 
the challenge of processing heterogeneous medical data, the gap between academic evaluation and real-world clinical needs, and the static nature of model knowledge. 
In this work, we introduced \method, a vision-centric MLLM that provides a systematic and holistic framework to address these critical limitations. 
Our work presents a systematic framework that addresses these limitations and demonstrates strong benchmark performance, suggesting a viable path toward deploying MLLMs in clinical environments.

At its core, \method's advancements are built on two principal contributions. 
Architecturally, we moved beyond monolithic vision encoders by proposing a compositional and cascaded architecture featuring the CaSL Fusion operator. 
This design unifies diverse 2D and native 3D data processing through a cascaded and 2D-anchored interplay mechanism, creating a rich and synergistic visual representation that consistently benefits nuanced diagnostic tasks, as confirmed by our ablation studies.
To bridge the evaluation-practice gap, we introduced a new vision-grounded evaluation framework. 
This includes the MedIF-Bench to enforce reliable instruction-following, and a novel RoI-grounded report generation methodology for clinically aligned and factualness-driven report generation evaluation.
To further enhance practical deployment, we developed an agentic tool use extension that equips \method with retrieval-augmented generation and the ability to use external specialist models as tools, demonstrating consistent improvements in both text-only and multimodal clinical scenarios.

Our comprehensive experiments demonstrate that these innovations translate into state-of-the-art performance. 
\method not only outperforms leading open-source models (\textit{e.g.}, Hulu-Med and Lingshu) across a wide array of 2D, 3D, and textual benchmarks but also surpasses powerful proprietary APIs on multimodal tasks. 
The superior performance, particularly in complex multimodal understanding, supports our architectural hypothesis that holistic medical perception benefits from a compositional vision system capable of natively processing heterogeneous 2D and 3D data, rather than relying on a single monolithic encoder.
Independently, the strong correlation between our RoI-Grounded evaluation and expert radiologist judgment confirms that vision-grounded, factualness-driven metrics better reflect real clinical needs than conventional text-matching approaches.
The consistent improvements observed when augmenting \method with agentic tool use further suggest that dynamic knowledge retrieval and specialist tool integration can complement the base model's capabilities.
Furthermore, the strong results on MedIF-Bench underscore that strong instruction-following capabilities are preserved through medical adaptation---a critical factor for clinical usability and a property often degraded in other medical MLLMs.

The broader implication of our work lies in the paradigm shift it advocates for. 
Our results suggest that advancing medical AI requires equal attention to both perception architecture and evaluation methodology---building modular, verifiable systems while ensuring that evaluation faithfully reflects clinical needs.
\method illustrates one instantiation of this modular design philosophy.
Its compositional architecture allows for flexible integration of new specialist encoders as imaging technology evolves. 
Its agentic tool use extension demonstrates how continuous knowledge updates and delegation to auditable, high-precision tools can further enhance a strong foundation model for practical deployment. 
Concurrently, our proposed evaluation paradigm provides the rigorous assessment required to ensure these complex systems are not only capable but also reliable and aligned with clinical workflows.


Despite its strong performance, we acknowledge several limitations that pave the way for future research. 
First, while our agentic system demonstrates the power of tool use, the current toolset is limited. 
Future work should expand this ecosystem to encompass a broader range of modalities (\textit{e.g.}, ultrasound, MRI) and functionalities (\textit{e.g.}, treatment planning simulators). 
Second, despite competitive performance among open-source medical MLLMs, a notable gap remains between \method and the strong proprietary models (\textit{e.g.}, Gemini-3-Flash) on knowledge-intensive text benchmarks. 
We attribute this primarily to the capacity and training data scale differences between our \method and the substantially larger proprietary systems, and note that our agentic tool use extension partially mitigates this gap through knowledge retrieval.
Finally, although \method excels in offline benchmarks, its translation into clinical practice necessitates rigorous human-in-the-loop evaluation to validate its impact on diagnostic accuracy, efficiency, and patient outcomes, while navigating the complex ethical and regulatory landscape.


In conclusion, \method lays the groundwork for a perceptually powerful, rigorously evaluated and verifiable medical AI system.
We hope this work motivates further research to build more clinically-aligned and actionable medical AI systems.

%% file: content/04_Methodology.tex
\clearpage
\section{Methods}


\subsection{Architecture}\label{sec:architecture}
\method is built upon the Qwen-VL series~\cite{qwen3-vl,bai2025qwen2.5-vl}, which is comprised of a Large Language Model (LLM), a redesigned Vision Transformer (ViT) architecture, which we name Qwen ViT in the paper, and a vision-language merger.
The core of the perception capabilities stems from the Qwen ViT, pre-trained on vast general-domain data to develop a strong understanding of general visual concepts.
However, a single architecture, even one as capable as Qwen ViT, is challenging for capturing the full spectrum of visual information present in medical imaging, which ranges from high-frequency textural details (crucial for pathology) to global structural semantics (for organ anatomy). 
We drew inspiration from recent works demonstrating that a composition of models can learn distributions more data-efficiently~\cite{du2024compositional} and that a strategic composition of encoders can induce significantly stronger capabilities~\cite{tong2024cambrian-1}. 
Building on this philosophy, \method extends the Qwen-VL architecture by introducing a compositional ensemble of specialized vision encoders that synergistically fuse diverse native 2D and 3D representations, enabling a more holistic and robust perception of medical imagery, as shown in~\cref{fig:architecture}a and \cref{fig:architecture}b.

\subsubsection{Perception with Compositional Vision Encoders}

\textbf{Enhancing 2D perception with additional 2D vision encoders.}
To construct a comprehensive 2D perception system, we create a compositional ensemble of vision encoders, where each model is selected to complement Qwen ViT's capability, addressing the inherent limitations of the generalist Qwen ViT.
Our two adopted specialist encoders are as follows:
\textbf{1)} \textit{A specialist for local features}: We incorporate ConvNeXt~\cite{liu2022convnet}, a modern convolutional network. 
Its strong inductive bias for spatial locality makes it exceptionally skilled at capturing high-frequency textural details and local patterns, which are critical for interpreting pathological slides or subtle radiological findings.
\textbf{2)} \textit{A specialist for robust semantics}: We include DINOv2~\cite{oquab2023dinov2}, a powerful self-supervised model. 
By learning from visual data without explicit language supervision, it develops robust, high-level semantic representations that are less prone to dataset biases, enhancing its ability to generalize across diverse medical imaging modalities.
This choice is further motivated by recent evidence that self-supervised encoders can scale effectively for medical image understanding even without text supervision~\cite{perez2025RAD-DINO}.
We additionally evaluated MedSigLIP~\cite{sellergren2025medgemma} as a candidate domain-knowledge specialist, as its pre-training on large-scale medical image--text pairs (similar in spirit to BiomedCLIP~\cite{zhang2025BiomedCLIP}) can inject clinical domain knowledge directly. 
However, our ablations (\supptabref{tab:ablation_multi_additional_encoder}) show it does not offer a consistent advantage, and we ultimately adopt DINOv2 for its stronger overall balance across VQA and report generation. 
We note that clinical domain knowledge is instead injected into the model through progressive staged training (\cref{sec:progressive_training}).

\noindent\textbf{Compressing representations with CaSL Fusion $\bm{\bowtie}$}.
To effectively fuse features from the compositional encoder ensemble, we design our fusion strategy around two core principles:
\textbf{1) Principle 1: Information density over token length}: We prioritize increasing the information density of the visual tokens fed into the LLM rather than increasing the number of tokens. 
This is motivated by findings that standard visual representations in MLLMs often contain significant redundancy, and compressing richer information into a fixed-length sequence is more efficient~\cite{yang2025visionzip,zhang2024sparsevlm}.
\textbf{2) Principle 2: Guided alignment via the foundation encoder}: We should leverage the strong, pre-existing alignment between the Qwen ViT and the LLM decoder. 
Therefore, the representations from newly introduced encoders should be fused in a way that guides them towards this established vision-language space, rather than disrupting it.

Let the set of encoders be denoted by $\{\mathcal{E}_i\}_{i=1}^M$, where $\mathcal{E}_1$ is always the foundational Qwen ViT. For an input image $\bm{I}$, each encoder $\mathcal{E}_i$ extracts a feature map $\bm{X}_i$. To harmonize the features, which originate from different embedding spaces and pre-processing pipelines, we first project and resize them:
\begin{equation}\label{eqn:resize_and_proj}
    \bm{X}_i = \bm{\mathcal{E}}_i (\bm{I}); \quad \bar{\bm{X}}_i = \text{Resize}(\bm{W}_{\textrm{align}} \bm{X}_i) \in \mathbb{R}^{N \times D}
\end{equation}
Here, $\bm{W}_{\textrm{align}}$ is a learnable projection matrix that maps the features from encoder $\mathcal{E}_i$ to the hidden dimension $D$ of the Qwen ViT. The Resize operation aligns each feature map to Qwen ViT's spatial shape of $H \times W$, where $N = H \times W$. 
One exception is ConvNeXt, whose feature map is instead resized to a larger spatial resolution to better preserve high-frequency detail; we defer this encoder-specific treatment to \cref{sec:implementation_details}.

To achieve this fusion, we propose the Cascade Spatial-Aware Locality Fusion (CaSL Fusion), denoted by the operator $\bm{\bowtie}$. 
The core idea of CaSL is to incrementally enrich the foundational Qwen ViT representation with features from the specialist encoders in a spatially-aware and computationally efficient manner. 
This is realized through a cascaded application of a local cross-attention mechanism, as shown in~\cref{fig:architecture}c.

We exemplify the process with the fusion of two feature maps, $\widetilde{\bm{X}}_1 = \bar{\bm{X}}_1 \bm{\bowtie} \bar{\bm{X}}_2$. 
We first define a spatial local field extractor, $\mathcal{N}_{\textrm{2D}}$, which extracts a $k \times k$ patch of neighboring tokens (the window size $k$ is detailed in \cref{sec:implementation_details}). 
For the $j$-th feature token $\bar{\bm{X}}_1[j]$, we treat it as a query $\bm{q}_j = \bar{\bm{X}}_1[j] \in \mathbb{R}^{1 \times D}$. 
The corresponding keys and values $\bm{K}_j$ and $\bm{V}_j$ are extracted from the local neighborhood centered at the same spatial location $j$ within the second feature map $\bar{\bm{X}}_2$:
\begin{equation}
    \bm{K}_j, \bm{V}_j = \mathcal{N}_{\textrm{2D}}(\bar{\bm{X}}_2)_{j} = \left\{ \bar{\bm{X}}_2[p] \mid p \in \mathcal{B}_{\textrm{2D}}(j) \right\} \in \mathbb{R}^{k^2 \times D}, \quad \forall j \in \{1, \ldots, N\}
\end{equation}
where $\mathcal{B}_{\textrm{2D}}(j)$ contains the index set around the $j$-th feature in the spatial dimension.
This local extraction enforces a strong spatial inductive bias and dramatically reduces the computational cost from quadratic (over $N$) to linear. 
The fusion is then performed via a standard cross-attention operation~\cite{AttentionAlluNeed}:
\begin{equation} \label{eqn:local_cross_attn}
    \bm{A}_j = \text{softmax}\left( \frac{\bm{q}_j \bm{K}_j^\top}{\sqrt{D}} \right) \in \mathbb{R}^{1 \times k^2}; \quad
    \bm{o}_j = \bm{A}_j \bm{V}_j \in \mathbb{R}^{1 \times D}
\end{equation}
Note that we omit the multi-head formulation and the projections for queries, keys, values and outputs for clarity.
The aggregated feature $\bm{o}_j$ is then integrated back into the query representation using a residual connection and a Feed-Forward Network (FFN), forming a complete fusion block:
\begin{equation}\label{eqn:res_connection}
    \bm{\widetilde{X}}'_1 = \bar{\bm{X}}_1 + \text{2DLocalCrossAttn}(\bar{\bm{X}}_1, \bar{\bm{X}}_2; k); \quad
    \widetilde{\bm{X}}_1 = \widetilde{\bm{X}}'_1 + \text{FFN}(\widetilde{\bm{X}}'_1)
\end{equation}
where $\text{2DLocalCrossAttn}(\cdot)$ denotes the process in~\cref{eqn:local_cross_attn} for all tokens.

For an ensemble of $M$ encoders, the fusion is applied in a cascaded manner, sequentially enriching the base representation:
\begin{equation}\label{eqn:full_CaSL_Fusion}
    \bm{Z} = \left(\left(\bar{\bm{X}_1} \bm{\bowtie} \bar{\bm{X}_2}\right)  \ldots  \bm{\bowtie} \bar{\bm{X}}_M \right)
\end{equation}
Crucially, the fusion operator $\bm{\bowtie}$ is \textit{asymmetric and left-associative}, where the left operand always serves as the query. 
This ensures that the well-aligned Qwen ViT representation remains the central backbone of the fusion process, which is progressively enhanced by the specialist encoders.
The fused feature $\bm{Z}$ replaces the original $\bar{\bm{X}_1}$ and is fed into the LLM decoder.

\noindent\textbf{CaSL Fusion with stochastic residual regularization.}
A potential challenge in our cascaded fusion design is the risk of optimization shortcuts. 
Since the Qwen ViT ($\bar{\bm{X}}_1$) is already well-aligned with the LLM, the model might learn to predominantly rely on this path during fine-tuning, effectively ignoring the contributions from less-aligned specialist encoders. 
To mitigate this, we introduce a stochastic regularization technique during training. 
Specifically, we modify the residual connection in the fusion block (\cref{eqn:res_connection}) by stochastically dropping the original query representation. 
The update rule becomes:
\begin{equation}
    \tilde{\bm{X}}'_1 = \lambda \cdot \bar{\bm{X}}_1 + \text{2DLocalCrossAttn}(\bar{\bm{X}}_1, \bar{\bm{X}}_2; k); \quad \lambda \sim \text{Bernoulli}(1 - p_{\text{drop}})
\end{equation}
During training, this forces the model to learn meaningful representations from the specialist encoders' features. 
During inference, this stochasticity is disabled by setting $\lambda=1$ and $p_{\text{drop}}=0$, ensuring stable and full utilization of all features. 
$p_{\text{drop}}$ is set to 0.1 by default.


\noindent\textbf{CaSL Fusion's compatibility with DeepStack.}
Qwen3-VL~\cite{qwen3-vl} introduced a DeepStack mechanism~\cite{meng2024deepstack}, which injects visual features into multiple, deeper layers of the LLM to avoid the dilution of visual information, enabling a more profound and persistent vision-language alignment.
CaSL Fusion is fully compatible with and designed to leverage this deep fusion paradigm. 
We denote the feature map from the foundational Qwen ViT ($\mathcal{E}_1$) at the $l$-th DeepStack layer as $\bar{\bm{X}}_1^{(l)}$, where $l \in \{1, \ldots, L\}$.
At each layer $l$, the layer-specific Qwen ViT feature, $\bar{\bm{X}}_1^{(l)}$, serves as the query for a new cascaded fusion. 
The specialist features are then sequentially fused with this query. 
This process yields a set of $L$ enriched DeepStack features $\{\bm{Z}^{(l)}\}_{l=1}^L$. 
The computation can be formulated as:
\begin{equation}
    \bm{Z}^{(l)} = \left( \left( \bar{\bm{X}}_{1}^{(l)} \bm{\bowtie} \bar{\bm{X}}_{2} \right) \ldots  \bm{\bowtie} \bar{\bm{X}}_M \right) 
    \quad \forall l \in \{1, \ldots, L\}
\end{equation}
$\bm{Z}^{(l)}$ then replaces the original DeepStack feature $\bar{\bm{X}}_1^{(l)}$ and is injected into the LLM. 
By enabling CaSL Fusion with DeepStack, we ensure that the LLM's reasoning remains continuously grounded in the visual evidence.



\subsubsection{2D-Anchored Native 3D Volumetric Understanding}
A limitation of existing medical MLLMs is their inability to natively process 3D volumetric data such as CT and MRI scans. 
Some methods~\cite{xu2025lingshu,jiang2025hulu-med} treat 3D volumes as a sequence of independent 2D slices, leading to the loss of critical inter-slice spatial context and anatomical continuity.
\method addresses this by integrating a dedicated 3D encoder and proposing a 2D-anchored 3D CaSL Fusion to enrich 2D anchor features with 3D volumetric features.
Our key motivation behind this strategy is twofold: 
\textbf{1)} \textbf{Accelerated alignment}: By leveraging the strong language alignment of the 2D features as anchors, the 3D features can rapidly converge to the vision-language space without requiring extensive 3D-specific alignment data;
\textbf{2)} \textbf{Knowledge transfer}: Even when processing 3D data, the model benefits from the rich pre-trained knowledge embedded in the 2D encoders, enabling robust generalization across diverse volumetric imaging modalities.

\noindent\textbf{Dual-path processing with 2D anchors.}
For a given 3D volume, we employ a dual-path processing strategy.

\textit{Path 1: Native 3D encoding.}
We incorporate a 3D vision encoder $\mathcal{E}_{\textrm{3D}}(\cdot)$~\cite{bolya2025perception_encoder} and we pre-align it with language through contrastive learning on 3D medical volumes and their corresponding radiology reports, following a paradigm that has proven effective for both 2D radiographs~\cite{zhou2022generalized_radiograph} and 3D oncology imaging~\cite{pai2024foundation_cancer}. 
This pre-alignment endows the encoder with an intrinsic understanding of 3D anatomical structures while establishing a basic connection to the language space.
Specifically, we process the entire 3D volume using 3D encoder and obtain a 3D feature map $\bm{P} = \mathcal{E}_{\textrm{3D}} (\bm{I}_{\textrm{3D}})$, where $\bm{I}_{\textrm{3D}}$ denotes the 3D volume used.


\textit{Path 2: 2D anchor encoding.} 
Simultaneously, a set of $S$ representative 2D slices $\mathcal{S}$ are randomly sampled from the input volume $\bm{I}_{\text{3D}}$ (we set $S=4$, used identically at training and inference). 
The anchor slices are thus denoted as $\{\bm{I}_{\text{3D}}[s] \mid s \in \mathcal{S}\}$.
The 2D features (after projection and resize) for these sampled frames can be denoted as:
$\{\bar{{\bm{X}}}_1[s], \bar{{\bm{X}}}_2[s], \ldots, \bar{{\bm{X}}}_M[s] \mid s \in \mathcal{S} \}$.
Utilizing the CaSL Fusion operator in~\cref{eqn:full_CaSL_Fusion}, we can obtain their fused 2D features, $\bm{Z}[s] \in \mathbb{R}^{N \times D}$, for all slices in $S$:
\begin{equation}
    \bm{Z}[s] = \left(\left(\bar{\bm{X}_1}[s] \bm{\bowtie} \bar{\bm{X}_2}[s]\right)  \ldots  \bm{\bowtie} \bar{\bm{X}}_M[s] \right), \quad \forall s \in \mathcal{S}
\end{equation}
This process generates a set of anchor feature maps, each rich with 2D-aligned semantics, ready to guide the fusion with the native 3D representation.

\noindent\textbf{2D-anchored depth-aware CaSL Fusion $\underset{\textrm{3D}}{\bm{\bowtie}}$.}
To effectively fuse the 3D features with the 2D anchor representations, we extend CaSL Fusion to a depth-aware variant.
The key insight is that each 2D anchor slice should attend to the 3D features not only in its spatial neighborhood but also across the depth dimension, thereby capturing the volumetric context that surrounds each slice.

Following the projection and resize operation mentioned in~\cref{eqn:resize_and_proj}, we also perform this operation for the 3D feature $\bm{P}$, obtaining $\bar{\bm{P}} \in \mathbb{R}^{D_{\textrm{vol}} \times N \times D}$, where its spatial shape aligns with the spatial size of $\bm{Z}[s] \in \mathbb{R}^{N \times D}$ and $D_{\textrm{vol}}$ represents 3D feature volume's depth ($D_{\textrm{vol}}=4$, obtained from a 32-frame input with a downsampling factor of 8 along the depth dimension).
We then exemplify the process with the fusion of 2D anchor features with 3D features, $\bm{\widetilde{\bm{Z}}[s]} = \bm{\bm{Z}[s]} \underset{\textrm{3D}}{\bm{\bowtie}} \bar{\bm{P}}$. 
We define a depth-aware spatial local field extractor, $\mathcal{N}_{\textrm{3D}}$, that extracts a patch spanning a $k \times k$ spatial area across the full depth $D_{\textrm{vol}}$.
For the $j$-th feature token in $\bm{Z}[s]$, we treat it as a query $\bm{q}_j = \bm{Z}[s,j] \in \mathbb{R}^{1 \times D}$. 
The corresponding keys and values $\bm{K}_j$ and $\bm{V}_j$ are extracted using depth-aware spatial local field extractor:
\begin{equation}
    \bm{K}_j, \bm{V}_j = \mathcal{N}_{\textrm{3D}}(\bar{\bm{P}})_{j} = \left\{ \bar{\bm{P}}[p] \mid p \in \mathcal{B}_{\textrm{3D}}(j) \right\} \in \mathbb{R}^{D_{\textrm{vol}} \times k^2 \times D}, \quad \forall j \in \{1, \ldots, N\}
\end{equation}
where $\mathcal{B}_{\textrm{3D}}(j)$ contains the index set around the $j$-th feature in the spatial and depth dimension.
After the definition of query, keys and values, we can perform the attention operation defined in~\cref{eqn:local_cross_attn} and residual connections defined in~\cref{eqn:res_connection}:
\begin{equation}\label{eqn:res_connection}
    \bm{\widetilde{\bm{Z}}[s]}' = \bm{Z}[s] + \text{3DLocalCrossAttn}(\bm{Z}[s], \bm{\bar{P}}; k, D_{\textrm{vol}}); \quad
    \bm{\widetilde{\bm{Z}}[s]} = \bm{\widetilde{\bm{Z}}[s]}' + \text{FFN}(\bm{\widetilde{\bm{Z}}[s]}')
\end{equation}
The obtained features $\bm{\widetilde{\bm{Z}}[s]}, \forall s \in \mathcal{S}$ are concatenated to compose a full sequence of length $S \times N$ and fed into the LLM decoder to perform inference.
As with 2D CaSL Fusion $\bm{\bowtie}$, $\underset{\textrm{3D}}{\bm{\bowtie}}$ is also compatible with DeepStack and stochastic residual regularization to enable robust feature learning.

\subsection{Vision-Grounded Evaluation System}\label{sec:evaluation_system}

To evaluate whether a medical multimodal LLM can function as a useful clinical tool, we introduce a comprehensive evaluation system that mirrors the integrated skills of a radiologist. 
As detailed in \exttabref{ext-tab:benchmark_summary}, this system builds on prior frameworks like MedGemma~\cite{sellergren2025medgemma}, Lingshu~\cite{xu2025lingshu}, HuLu-Med~\cite{jiang2025hulu-med}, and HuatuoGPT~\cite{zhang2023huatuogpt} to assess five critical dimensions: 
foundational 2D and 3D visual understanding (evaluated by 2D and 3D VQA tasks), text-based medical knowledge (evaluated by QA tasks), clinical report generation, and medical instruction-following. 
It is worth noting that we extend prior frameworks by incorporating medical instruction-following evaluation, as this capability is overlooked in all existing benchmarks yet critical for determining model usability.
In the following subsections, we describe the evaluation protocol for each benchmark category.


\subsubsection{Evaluation Prompt Templates for VQA and QA}

For 2D and 3D medical VQA, as well as text-only QA problems in the benchmark, we adopt a set of carefully designed prompts and evaluation procedures. The complete prompt templates are shown in~\extfigref{ext-fig:eval-prompt-ifbench}a.

\noindent\textbf{Multiple Choice Questions (MCQ).} As shown in the left panel of~\extfigref{ext-fig:eval-prompt-ifbench}a, the MCQ prompt instructs the model to answer a multiple-choice question based on the provided image(s). 
Note that for text-only QA tasks, the phrase ``based on the provided image(s)'' is removed from the prompt.
The prompt explicitly states that the model must select exactly one option and output only the single uppercase letter of the correct option enclosed in \texttt{<answer></answer>} tags. To reduce ambiguity, the prompt includes both a correct format example (\textit{e.g.}, \texttt{<answer>A</answer>}) and several incorrect format examples that would fail automated grading, such as including extra text, missing tags, outputting option text instead of the letter, or selecting multiple letters. 
We designed this prompt to be strict for three reasons: 
\textbf{1)}~we observed that some open-source models have poor instruction-following ability, and simpler prompts often fail to elicit the correct output format, which is essential for reliable rule-based evaluation via regular expression matching; 
\textbf{2)}~suppressing unrelated output reduces decoding time and thus speeds up evaluation; and 
\textbf{3)}~the prompt differs from the training prompts of most models, so it also serves as a test of generalization to unseen prompts, as recommended by~\cite{mizrahi2024state}. 
The extracted letter is then directly compared against the GT answer for scoring.

\noindent\textbf{Open-ended VQA.} As shown in the upper-right panel of~\extfigref{ext-fig:eval-prompt-ifbench}a, the open-ended prompt asks the model to analyze the provided image(s) and give a concise, direct answer to the question. 
Similarly, for text-only QA tasks, the phrase ``based on the provided image(s)'' is removed from the prompt.
The prompt is kept minimal to avoid biasing the response while still discouraging unnecessary verbosity.
Since open-ended answers cannot be evaluated by exact string matching, we employ an LLM-as-a-Judge\footnote{Unless otherwise specified, all LLM-as-a-Judge evaluations in this work use OpenAI \texttt{gpt-4.1-2025-04-14} as the judge model.} approach for scoring. 
The judge prompt, shown in the lower-right panel of~\extfigref{ext-fig:eval-prompt-ifbench}a, provides the original question, the GT answer, and the model's answer to a language model judge. 
The judge is instructed to determine whether the model's answer is correct by checking for semantic equivalence with the GT---if the model's answer expresses a similar meaning or a valid alternative interpretation of the standard answer, it is considered correct. 
The judge outputs a structured verdict (\texttt{correct} or \texttt{incorrect}) under a \texttt{[JUDGEMENT]} tag, which is then parsed for scoring.

\subsubsection{MedIF-Bench}

We observe that medical domain fine-tuning often hurts a model's instruction-following ability. In our own training experiments, as well as in evaluations of several open-source medical MLLMs, we find that general-purpose MLLM models fine-tuned on medical SFT data frequently fail to comply with output format requirements---even when the underlying medical knowledge is correct. This is problematic because real clinical applications often require outputs in specific structured formats (\textit{e.g.}, JSON for integration with electronic health records, reports with standardized section headers, or reasoning traces with confidence levels). A model that gives a correct diagnosis but ignores the requested format has limited practical utility. To track this issue, we developed MedIF-Bench, a benchmark that evaluates whether medical MLLMs can follow diverse, clinically relevant formatting instructions.

\noindent\textbf{Benchmark construction.}
MedIF-Bench contains 900 samples in total, built around a pool of manually curated formatting instructions organized by task category. 
For each sample, the model receives both a medical task and a specific formatting instruction that dictates how the output must be structured. 
Importantly, \textbf{this benchmark evaluates format compliance only, not medical accuracy}---the goal is to measure whether the model can produce outputs that conform to the requested structure, regardless of whether the medical content is correct. 
Each response is checked against a predefined regular expression pattern: a response scores 1.0 if and only if it fully matches the required format, and 0.0 otherwise.

The instruction pool covers seven task categories, each with its own pool of manually curated instructions and corresponding regex patterns. 
The first group of categories---\textbf{MCQ}, \textbf{MCQ with context}, and \textbf{Open}---are primarily constructed from the benchmark datasets listed in~\exttabref{ext-tab:benchmark_summary}. 
For these categories, we curate prompts that cover a range of common formatting requirements with special markers and structures, where the model must present its selected option or answer in a prescribed format. 
The second group of categories---\textbf{Prognosis}, \textbf{Report generation}, \textbf{SEER}, and \textbf{Treatment}---focus on extracting structured clinical variables from imaging or patient data for specific clinical tasks such as survival prognosis, report generation, and treatment recommendation. 
For these categories, some samples are drawn from the benchmark datasets in~\exttabref{ext-tab:benchmark_summary}, while others are constructed using real patient records from the SEER cancer registry~\cite{seer_program}, grounding the evaluation in realistic oncological data. 
Three examples from MedIF-Bench are shown in~\extfigref{ext-fig:eval-prompt-ifbench}b.

\subsubsection{RoI-Grounded Report Generation}
\label{sec:roi_grounded_report_generation}

Existing evaluation pipelines for medical report generation suffer from two primary flaws. 
First, they typically prompt the model to generate a full report based solely on the image, entirely devoid of any clinical background information. 
Second, existing evaluation metrics, whether surface-level lexical or semantic matching (such as BLEU~\cite{papineni2002bleu}, ROUGE~\cite{lin2004rouge}, BERTScore~\cite{Zhang2020BERTScore}), entity-level extraction that relies on small-scale models with limited accuracy (such as RadGraph-F1~\cite{delbrouck-etal-2024-radgraph}), or holistic LLM-based scoring that conflates accuracy and completeness (such as GREEN~\cite{ostmeier2024GREEN}), fail to provide a fine-grained, factuality-driven decomposition of clinical findings.

To address these limitations, we propose a novel evaluation paradigm: RoI-Grounded Report Generation, which is built upon two key innovations: 
\textbf{1)} context-aware report generation and 
\textbf{2)} factualness-driven LLM-as-a-Judge evaluation.
The complete workflow (\extfigref{ext-fig:roi-pipeline}) proceeds in three stages.
First, we extract a clinical context from the GT report, consisting of potential background information and regions of interest. 
Next, we use this context to guide the model in generating a focused, RoI-grounded medical report. 
Finally, an LLM judge compares the claims in the generated report against the GT, categorizing the model claims to calculate exact clinical precision, recall, and F1 scores.

\noindent\textbf{Context-aware report generation.} 
Instead of naively asking the model to describe an entire image from scratch, we first extract a high-level clinical context for more clinically aligned report generation.
Specifically, the clinical indication and the anatomical areas of focus—from the GT report using LLMs. 
This extracted context is then injected into the prompt for the model-under-test. 
A strict security constraint is applied during the extraction phase to ensure absolutely no specific findings or abnormalities are leaked.
This design choice is driven by two fundamental justifications:
\textbf{1)} \textbf{Mimicking real-world workflows:} Real-world radiologists do not read scans blindly; they consistently review patient background information and clinical indications before writing a report. Our pipeline faithfully simulates this standard clinical practice.
\textbf{2)} \textbf{Ensuring verifiability:} A model's output can only be rigorously verified if its claims fall within the scope of the reference report. 
If a model is evaluated without context and describes a region not mentioned in the GT, it is impossible to objectively determine whether the model hallucinated or if the original human radiologist simply omitted a normal finding. 
By restricting the model to specific Regions of Interest (RoIs) defined by the GT, we ensure that the generated statements can be assessed against the reference report, substantially improving the verifiability of the evaluation.

\extfigref{ext-fig:roi-pipeline}a presents the prompt design for context extraction alongside an actual example using the GPT 4.1 API. 
\extfigref{ext-fig:roi-pipeline}b illustrates how this context is utilized as a condition for report generation on the same actual sample using our MLLM.

\noindent\textbf{Factualness-driven LLM-as-a-Judge evaluation.} 
To move beyond coarse semantic similarity, we leverage the advanced text-processing capabilities of modern LLMs to evaluate exact clinical factualness. Instead of generating a single holistic score, the LLM judge is tasked with comparing the GT report against the model's generated report to categorize the diagnostic claims into three explicit lists of abnormalities: 
\texttt{[MATCHED\_ABNORMALITY]} (True Positives, $TP$), meaning the GT mentions an abnormality and the model correctly identifies it; 
\texttt{[MISSED\_ABNORMALITY]} (False Negatives, $FN$), meaning the GT mentions an abnormality but the model either omits it or falsely claims the area is normal; and 
\texttt{[HALLUCINATED\_ABNORMALITY]} (False Positives, $FP$), meaning the model fabricates an abnormality that is either explicitly stated as normal or completely unmentioned in the GT. 

Based on these exact counts, we compute rigorous clinical metrics: Precision $\frac{TP}{TP+FP}$, Recall $\frac{TP}{TP+FN}$, and the F1-score $2 \times \frac{\text{Precision} \times \text{Recall}}{\text{Precision} + \text{Recall}}$, which provide transparent, granular indicators of diagnostic correctness. 
\extfigref{ext-fig:roi-pipeline}c provides a worked example of this scoring process, showing how TP, FN, and FP counts are extracted and converted to Precision, Recall, and F1.

In the Results section (\cref{sec:roi_rg_case}), we further validate this evaluation design by comparing against the conventional context-free + RadGraph-F1 pipeline~\cite{delbrouck-etal-2024-radgraph,sellergren2025medgemma}, demonstrating that 
\textbf{1)} context-free generation renders metrics unreliable by penalizing unverifiable claims as false positives, and 
\textbf{2)} RadGraph-F1 suffers from synonymy insensitivity and length bias that our LLM-as-a-Judge avoids.


\subsection{Data Curation}\label{sec:data_curation}

Training a clinically useful medical MLLM requires data that simultaneously supports
\textbf{1)} robust multimodal perception across heterogeneous medical imagery,
\textbf{2)} medical reasoning and evidence-based answering, and
\textbf{3)} realistic clinical workflows---especially 3D volumetric CT interpretation.
To this end, we collect and construct a large-scale training corpus designed for our multi-stage
training curriculum (\cref{sec:progressive_training}), as shown in~\extfigref{ext-fig:data_curation}.
The overall corpus contains 22.2 million samples: 12.6M medical multimodal, 4.7M medical text, 3.4M general multimodal, and 1.5M general text (\exttabref{ext-tab:data_sources_v260}).

\noindent\textbf{Data sources.}
Our data originates from three complementary sources:
\textbf{1)} open-source medical multimodal datasets spanning radiology and pathology;
\textbf{2)} medical evidence corpora including clinical guidelines, journal articles, and case reports; and
\textbf{3)} CT--report aligned data, combining open-source CT studies with high-quality reports and hospital-collected CT scans with synthesized reports.

\noindent\textbf{Data synthesis pipelines.}
To address specific training bottlenecks, we develop four data synthesis pipelines.
Detailed procedures and illustrations (\suppfigref{fig:data_pipelines}) for each pipeline are provided in Supplementary Methods.
\begin{itemize}
    \item \textbf{Instruction Warping} (\suppnoteref{sec:instruction_warping}).
    Medical adaptation tends to erode general instruction-following ability, which in turn constrains how the underlying medical knowledge can be expressed in realistic clinical settings (\textit{e.g.}, structured outputs, scope restrictions, format compliance).
    To preserve this ability during fine-tuning, we manually author 120 seed instructions covering diverse clinical roles, scenarios, and formatting requirements, and use LLM-based expansion to produce 21k diverse instruction templates that are applied to existing samples via schema-based recomposition.

    \item \textbf{Dense Caption Generation} (\suppnoteref{sec:dense_caption_mining}).
    Paper-figure captions provide only sparse supervision and composite figures introduce noise.
    We perform sub-figure segmentation and multi-stage filtering to isolate high-quality medical images, then generate dense visually-grounded descriptions using a dual-VLM consensus pipeline, yielding 610k dense-caption samples from 100k journal articles and 785k case reports.

    \item \textbf{Interactive Multimodal CoT Annotation} (\suppnoteref{sec:multi_agent_reasoning}).
    To train long-horizon multimodal reasoning, we mine 100k reasoning-hard instances (high generation length yet low correctness) and solve them with a multi-agent pipeline~\cite{jian2025look} coupling a text-only orchestrator with a multimodal visual grounder through iterative evidence-seeking interaction, retaining 67k trajectories whose conclusions match reference answers.

    \item \textbf{Large-Scale CT Caption with Expert Toolset} (\suppnoteref{sec:ct_report_alignment}).
    To enable native 3D understanding at scale, we curate 0.2M open-source CT studies with high-quality reports~\cite{bai2024m3d,hamamci2024CT-Rate,blankemeier2024merlin,huang2023inspect} and develop a segmentation-assisted annotation pipeline based on lesion~\cite{he2025vista3d} and organ~\cite{wasserthal2023totalsegmentator} segmentation models to generate pseudo-report supervision for 1.1M hospital-collected CT scans, substantially expanding CT--report aligned data.
    The CT corpus supports both Stage \ding{175} (rapid 3D encoder alignment) and Stage \ding{176} (holistic 2D/3D instruction tuning).
\end{itemize}

\subsection{Agentic Tool Use} \label{sec:agentic_tool_use}

Beyond the core architectural and evaluation contributions, we develop an agentic tool use extension to enhance \method's practical clinical deployment. Although MLLMs excel at language-based reasoning and flexible instruction following, their reliability in high-stakes settings can be further improved by addressing 
\textbf{1)} time-sensitive medical knowledge that cannot be fully memorized in parameters, 
\textbf{2)} strict traceability requirements for clinical accountability, and 
\textbf{3)} perception bottlenecks in certain imaging sub-tasks where specialist vision models remain stronger than generalist MLLMs. 

To address these deployment considerations, we equip \method with a tool ecosystem and an explicit \emph{plan--act} workflow, illustrated in~\extfigref{ext-fig:agent}c.
Given a multimodal clinical input (a user query together with one or more medical images), \method first enters a planning stage, in which it analyzes the query, optionally performs multi-query rewriting for text inputs~\cite{li2024dmqr-rag}, identifies the evidence required to answer it, and produces an ordered tool-invocation plan.
In the subsequent stage of tool use and execution, \method sequentially calls the planned tools---which fall into two complementary categories, \emph{knowledge tools} for evidence-grounded retrieval and \emph{perception expert tools} for high-precision medical image analysis (described below)---and collects their structured outputs as auditable intermediate evidence.
Finally, the model synthesizes the query, the original visual input, and the aggregated tool outputs into a single clinical response, ensuring that the final answer is supported by traceable evidence rather than implicit parametric recall alone.
This design follows a principle of \textit{capability compositionality}: 
\method serves as the central reasoning and orchestration engine, while tools provide verifiable evidence and specialized perception signals that can be audited and integrated into final outputs.

\subsubsection{Knowledge Tools for Evidence-Grounded Retrieval}
\label{sec:rag_tool}

In clinical settings, answers must be supported by up-to-date, authoritative evidence and remain traceable.
Pure parametric recall is inadequate because medical knowledge evolves quickly and unverifiable statements are unacceptable.
We therefore equip \method with a RAG knowledge tool built on a hybrid retrieval engine that combines vector-based passage retrieval with knowledge-graph (KG)–based concept recall.
To handle the pervasive ambiguity in medical text (synonyms, abbreviations, brand--generic drug names), the retrieval engine performs concept normalization using UMLS~\cite{UMLS2024AA} before querying, mapping medical entities in the query to standardized concepts.
The vector channel then retrieves semantically matched passages from large-scale text corpora, while the KG channel leverages UMLS and PrimeKG~\cite{Chandak2022PrimeKG} to expand normalized concepts via clinically meaningful relations (\textit{e.g.}, drug--disease, symptom--disease, contraindication), reducing misses caused by surface-form mismatch.
Retrieved passages from both channels are then aggregated by clustering content that states the same clinical claim, and \method summarizes each cluster with attached citations to produce a compact, evidence-grounded context.

We organize our evidence resources into three complementary blocks:
\textbf{1)} large-scale unstructured text corpora for broad coverage, including PubMed~\cite{macleod2002pubmed}, Wikipedia, StatPearls~\cite{statpearls}, and medical textbooks~\cite{jin2021medqa_usmle}, following MedRAG~\cite{xiong2024medrag};
\textbf{2)} open-source medical knowledge graphs (UMLS and PrimeKG, described above); and
\textbf{3)} curated in-house repositories comprising a guideline knowledge base of 18K clinical guidelines and expert consensuses released since 2006 by official authorities in China and the United States, and a journal knowledge base of 100K research articles from leading medical journals published since 2015.
Detailed retrieval and aggregation procedures are provided in Supplementary Methods.

\subsubsection{Perception Tools for Specialized Visual Analysis}
\label{sec:perception_tools}

MLLMs offer flexible multimodal reasoning, but can be unreliable on perception-critical imaging tasks that demand voxel-level precision and stable quantification (\textit{e.g.}, exact lesion boundaries, small-structure detection, and subtle CT signs).
In clinical settings, these perception outputs often define the downstream conclusion---measurements, staging, response assessment, and follow-up comparability.
We therefore equip \method with perception expert tools and treat them as external senses: when a question hinges on fine-grained visual evidence, \method actively delegates perception to specialized models and then reasons over their structured results.
We currently integrate three CT-focused experts and two chest X-ray-focused experts:
\begin{itemize}
    \item \textbf{Multi-organ lesion segmentation expert}~\cite{he2025vista3d} segments 29 common lesion types across eight organs (breast, colon, esophagus, stomach, kidney, lung, liver, pancreas) and produces structured lesion inventories with counts, locations, and sizes.
    
    \item \textbf{Fatty liver assessment expert}~\cite{gao2026multi} estimates fatty liver severity from CT and localizes the liver parenchyma region used for assessment, producing a severity grade with the supporting parenchyma region.

    \item \textbf{Abdominal lymph node segmentation expert}~\cite{yu2025effective} segments abdominal lymph nodes and outputs node enumeration with size measurements, supporting the detection of lymphadenopathy.
    
    \item \textbf{Chest X-ray finding classification expert}~\cite{cohen2020limits,yang2025chest} predicts structured probabilities for thoracic findings from chest radiographs, integrating TorchXRayVision (18 common findings) and CheXFound (40 findings).

    \item \textbf{Chest X-ray finding generation expert}~\cite{chexagent-2024} generates clinically coherent chest X-ray finding descriptions, used for sentence-level imaging evidence when classification scores alone are insufficient.
\end{itemize}

Each tool produces structured outputs that are normalized into compact evidence objects and verbalized into natural-language evidence statements before being fed back to \method.
Full tool descriptions and routing details are provided in Supplementary Methods.

\subsection{Progressive Staged Training}\label{sec:progressive_training}

Training a complex, multi-component architecture like \method from scratch is challenging and computationally prohibitive. 
To ensure stability and effective knowledge transfer from pre-trained components, we devise a progressive training strategy. 
This curriculum-like approach systematically builds the model's capabilities by gradually increasing task complexity and data modality, moving from 2D alignment to holistic 3D instruction following. 
The entire process is summarized in~\exttabref{ext-tab:training_stages} and detailed below.

\begin{itemize}
    \item \textbf{Stage 1: Shallow multi-encoder alignment (2D)}. 
    The initial stage focuses on rapidly aligning the feature spaces of the newly introduced specialist 2D encoders with the foundational Qwen ViT. 
    Using a large corpus of 2D medical image-text pairs, we exclusively train the projection layers, the CaSL Fusion modules and all the encoders. 
    The LLM and all base vision encoders remain frozen, allowing for an efficient and targeted alignment of the perception components, specifically in the medical domain.

    \item \textbf{Stage 2: Deep vision-language alignment (2D)}. 
    Once the visual features are harmonized, this stage aims to establish a deep semantic alignment between the fused 2D visual representation and the entire LLM. 
    We unfreeze the LLM and continue training on the same 2D image-text alignment data. 
    This step fine-tunes the full model to understand and describe the rich, multi-encoder visual information.

    \item \textbf{Stage 3: 2D instruction tuning}. 
    With the model now capable of grounded 2D perception, we teach it to follow complex instructions and perform diverse medical tasks. 
    We fine-tune the entire model on a high-quality, multi-task 2D visual instruction dataset. 
    This stage imbues the model with advanced reasoning and task-execution capabilities for 2D imagery.
    
    \item \textbf{Stage 4: Rapid 3D encoder alignment}.
    This stage introduces the 3D modality. 
    Leveraging our 2D-anchored strategy, the goal is to efficiently align the new 3D encoder with the rest of the model and to retain the task-execution capabilities for 2D imagery. 
    Using 3D volume-text alignment data, we freeze the LLM and all 2D perception components (encoders and fusion modules) and train only the 3D encoder and the 2D-anchored depth-aware fusion modules. 
    This isolates the training to the new 3D components, enabling rapid and stable convergence.
    
    \item \textbf{Stage 5: Holistic 2D/3D instruction tuning}. 
    In the final stage, we unlock the model's full potential by fine-tuning all components on a comprehensive visual instruction tuning dataset combining 2D and 3D medical imagery. 
    This step unifies the model's capabilities, enabling it to handle complex instructions that require a deep understanding of both 2D and 3D medical data.

\end{itemize}

This staged training ensures that each new capability is built upon a stable foundation, culminating in a robust and powerful multimodal medical model.

\subsection{Implementation Details}\label{sec:implementation_details}

\noindent\textbf{Architectural details.}
The final \method builds on Qwen3-VL, whose native ViT processes images at dynamic resolution (patch size $16$ with $2\times2$ token merging). 
We augment it with two specialist 2D encoders (DINOv2-Large and ConvNeXt-Large-d-320) and a native 3D encoder (PE-3D). 
Each encoder operates at the input resolution of its pre-trained checkpoint so as to preserve its original representational quality; these resolutions therefore differ across encoders: $224\times224$ (center crop) for DINOv2-Large and $320\times320$ for ConvNeXt-Large-d-320, while PE-3D processes volumes at $32\times256\times256$ (depth$\times$height$\times$width) with a patch size of $8^3$. 
MedSigLIP, assessed only as an ablation candidate at $448\times448$, is not part of the final configuration. 
For 3D inputs, we sample $S=4$ anchor slices, and the resulting 3D feature depth is $D_{\textrm{vol}}=4$ (a 32-frame input downsampled by a factor of 8 along the depth dimension). 
During training, the stochastic residual regularization uses a drop probability of $p_{\textrm{drop}}=0.1$, which is disabled at inference by setting $\lambda=1$. 
The local cross-attention window size $k$ in CaSL Fusion (\cref{eqn:local_cross_attn}) is set per encoder rather than shared: $k=3$ for DINOv2 and the native 3D encoder, and $k=5$ for ConvNeXt. 
The larger window for ConvNeXt is tied to a dedicated resizing scheme: the ConvNeXt feature map is directly upsampled to $5H \times 5W$ to retain finer spatial detail (a larger upsampling factor was not adopted due to memory constraints during training). 
Consequently, each Qwen ViT token corresponds to a $5 \times 5$ high-frequency sub-region of the ConvNeXt feature map, which preserves the high-frequency textural detail that ConvNeXt is specialized to capture.

\noindent\textbf{Training details.}
All stages share the same optimization setup. 
We use the AdamW optimizer ($\beta_1=0.9$, $\beta_2=0.999$, $\epsilon=10^{-8}$) with weight decay $0$ and gradient clipping at a maximum norm of $1.0$. 
The learning rate is set to $1\times10^{-5}$ for all trainable components (the LLM, the projection layers, and the vision encoders), following a cosine schedule with $100$ warmup steps, and all training is performed in BF16 precision. 
Training is implemented with DeepSpeed ZeRO-3, FlashAttention-2, and gradient checkpointing, using a sequence cutoff length of $4096$, and is conducted on NVIDIA H20 GPUs. 
The per-stage configurations differ mainly in the number of GPUs, effective batch size, and number of epochs. 
Stages~1--3 use an effective batch size of $128$ on $16$ GPUs, trained for $1$, $1$, and $3$ epochs, respectively. 
Stage~4 uses an effective batch size of $256$ on $16$ GPUs for $1$ epoch. 
Stage~5 uses an effective batch size of $256$ on $32$ GPUs for $3$ epochs.

%% file: content/07_Figures.tex


\begin{figure}[!t]
\includegraphics[width=1.0\textwidth]{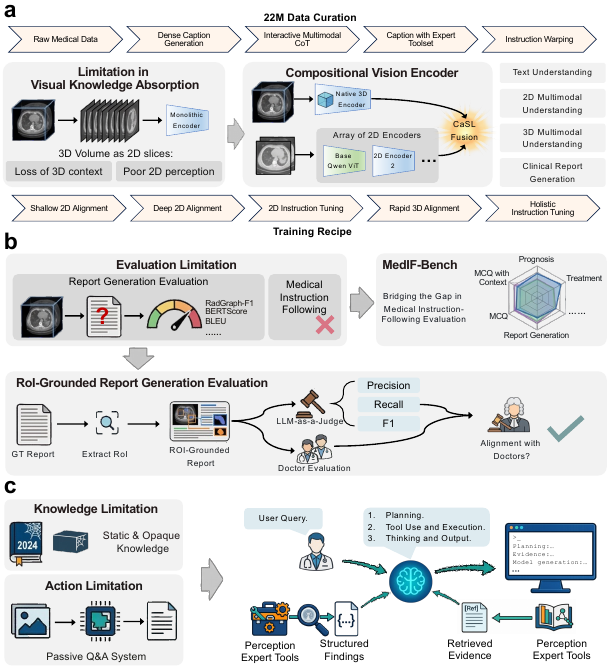}
\caption{
\textbf{Overview of the \method framework.}
\textbf{a}, Compositional vision encoder. 
We address the limitation of monolithic encoders by proposing a compositional architecture with a native 3D encoder and an array of 2D encoders unified via CaSL Fusion, supported by a 22M-sample data curation pipeline and a progressive 5-stage training recipe.
\textbf{b}, Vision-grounded evaluation.
We identify key evaluation limitations---lack of instruction-following assessment and reliance on global-level text-matching metrics---and introduce MedIF-Bench for medical instruction following and an RoI-Grounded Report Generation Evaluation scheme that assesses diagnostic accuracy at the region level using an LLM-as-a-Judge.
\textbf{c}, Agentic tool use extension.
To enhance deployment, we equip \method with perception expert tools and a retrieval-augmented generation pipeline within a plan--act workflow, demonstrating consistent improvements in both text-only and multimodal clinical scenarios.
}
\label{fig:contributions_overview}
\end{figure}


\clearpage
\begin{center}
\includegraphics[width=1.0\linewidth]{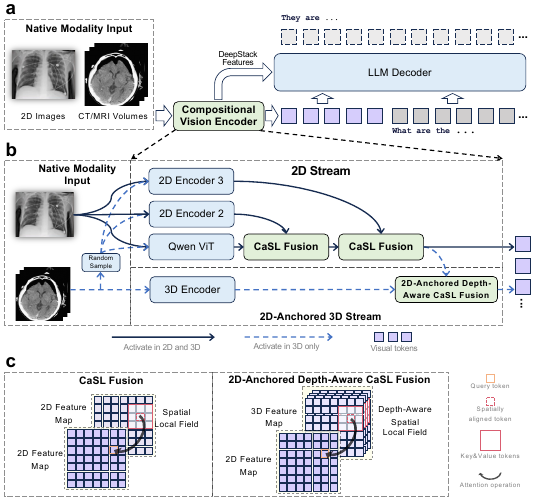}
\captionof{figure}{
\textbf{The architecture details of \method, a multimodal large language model for holistic medical understanding}.
\textbf{a}, The overall framework of \method. 
The model accepts native 2D and 3D medical images as input. 
A Compositional Vision Encoder module processes these inputs to generate a sequence of information-dense visual tokens.
These tokens are then fed into an LLM decoder to enable complex reasoning and generation. 
The architecture is compatible with deep fusion mechanisms like DeepStack, where visual features are injected at multiple layers of the LLM.
\textbf{b}, A detailed view of the Compositional Vision Encoder module, which comprises two primary processing streams. 
The 2D Stream handles 2D images using a compositional ensemble of encoders: a foundational Qwen ViT and several specialist encoders (\textit{e.g.}, Encoder 2, Encoder 3). Features are progressively refined through a series of CaSL Fusion blocks in a cascaded manner. 
The 2D-Anchored 3D Stream is activated for 3D volumetric input. 
It processes the full 3D volume with a native 3D encoder while simultaneously processing randomly sampled 2D slices (anchors) through the 2D stream. 
A final CaSL Fusion block then integrates the 3D features, guided by the well-aligned 2D anchor representations.
\textbf{c}, A schematic of the core CaSL Fusion operator. 
Left: The standard CaSL Fusion operator used in the 2D stream. 
A query token from one feature map attends to a small, spatially local field of key/value tokens from another feature map, enforcing a spatial inductive bias. 
Right: The 2D-Anchored Depth-Aware CaSL Fusion operator. 
A 2D query token attends to a 3D local field that spans both a spatial neighborhood and the depth dimension of the 3D feature map, enabling the 2D representation to be enriched with volumetric context.
}
\label{fig:architecture}
\end{center}

\begin{figure}[!t]
\centering
\includegraphics[width=1.0\linewidth]{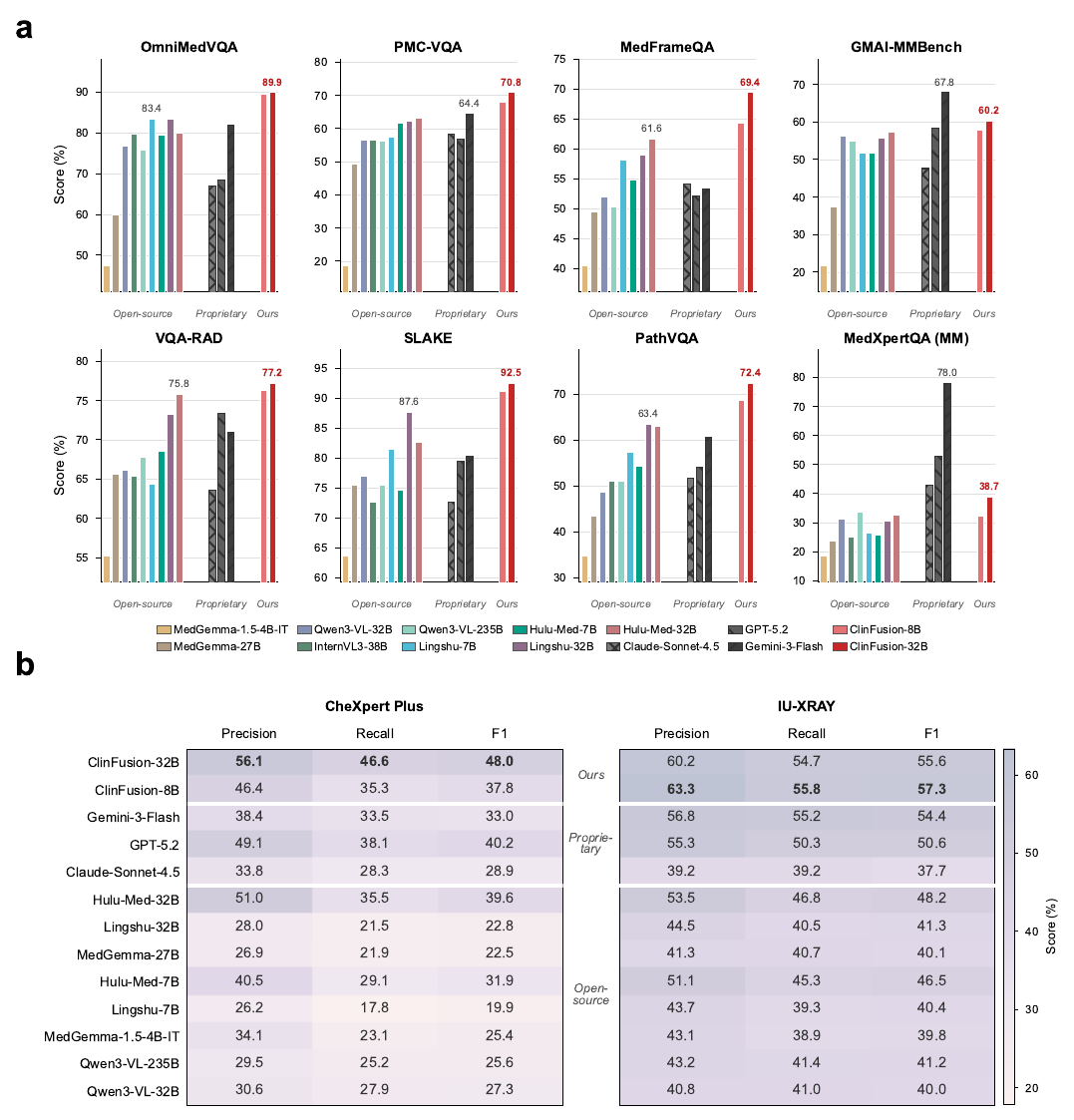}
\caption{
\textbf{Performance of \method on 2D multimodal medical benchmarks.}
\textbf{a},  Results on 2D multimodal medical VQA benchmarks, grouped into multi-modality VQA (OmniMedVQA, PMC-VQA, MedFrameQA, GMAI-MMBench), specific-modality VQA (VQA-RAD, SLAKE, PathVQA), and a reasoning-oriented benchmark (MedXpertQA-MM).
We compare \method against leading proprietary models, general-purpose MLLMs, and medical MLLMs at various parameter scales.
Results are all re-benchmarked by us in a fair protocol.
\textbf{b},  Results on 2D report generation benchmarks (CheXpert-Plus and IU-Xray), evaluated using the proposed RoI-Grounded Report Generation protocol and reported as Precision, Recall and F1.
}
\label{fig:2d_mm_bench_2d_report}
\end{figure}

\begin{figure}[!t]
\centering
\includegraphics[width=1.0\linewidth]{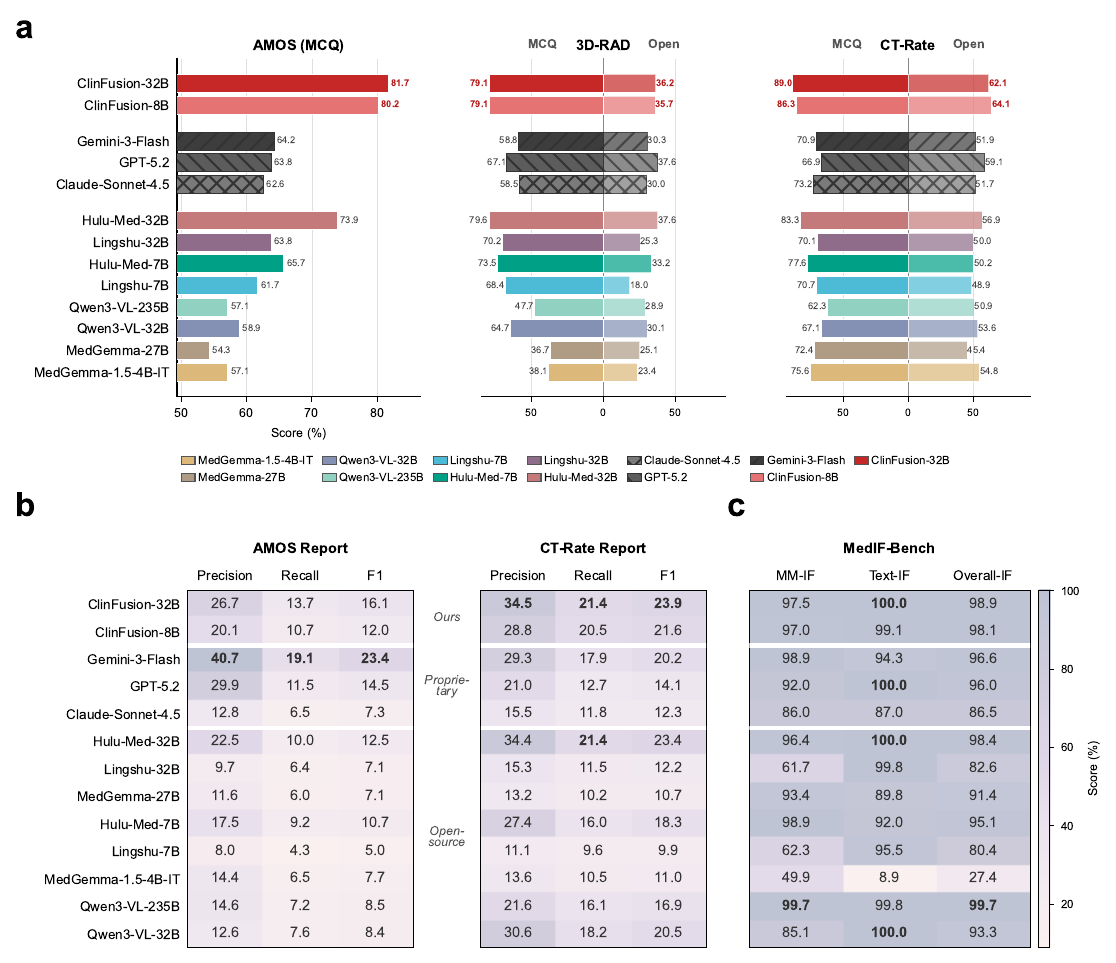}
\caption{
\textbf{Performance of \method on 3D multimodal medical benchmarks and MedIF-Bench.}
\textbf{a},  3D multimodal medical benchmark results on AMOS-MM (MCQ), 3D-RAD (MCQ and Open-ended), and CT-Rate-VQA (MCQ and Open-ended).
We compare our model with proprietary and open-source models, re-benchmarked by us in a fair protocol.
\textbf{b},  3D report generation results on AMOS-MM Report and CT-Rate Report, evaluated using the RoI-Grounded protocol and reported as Precision, Recall and F1. 
\textbf{c},  Results on our proposed MedIF-Bench, reporting composite instruction-following scores for multimodal inputs (MM-IF), text-only inputs (Text-IF), and the overall average (Overall-IF). 
}
\label{fig:3d_mm_bench_3d_report_medifbench}
\end{figure}

\clearpage
\begin{center}
\includegraphics[width=1.0\linewidth]{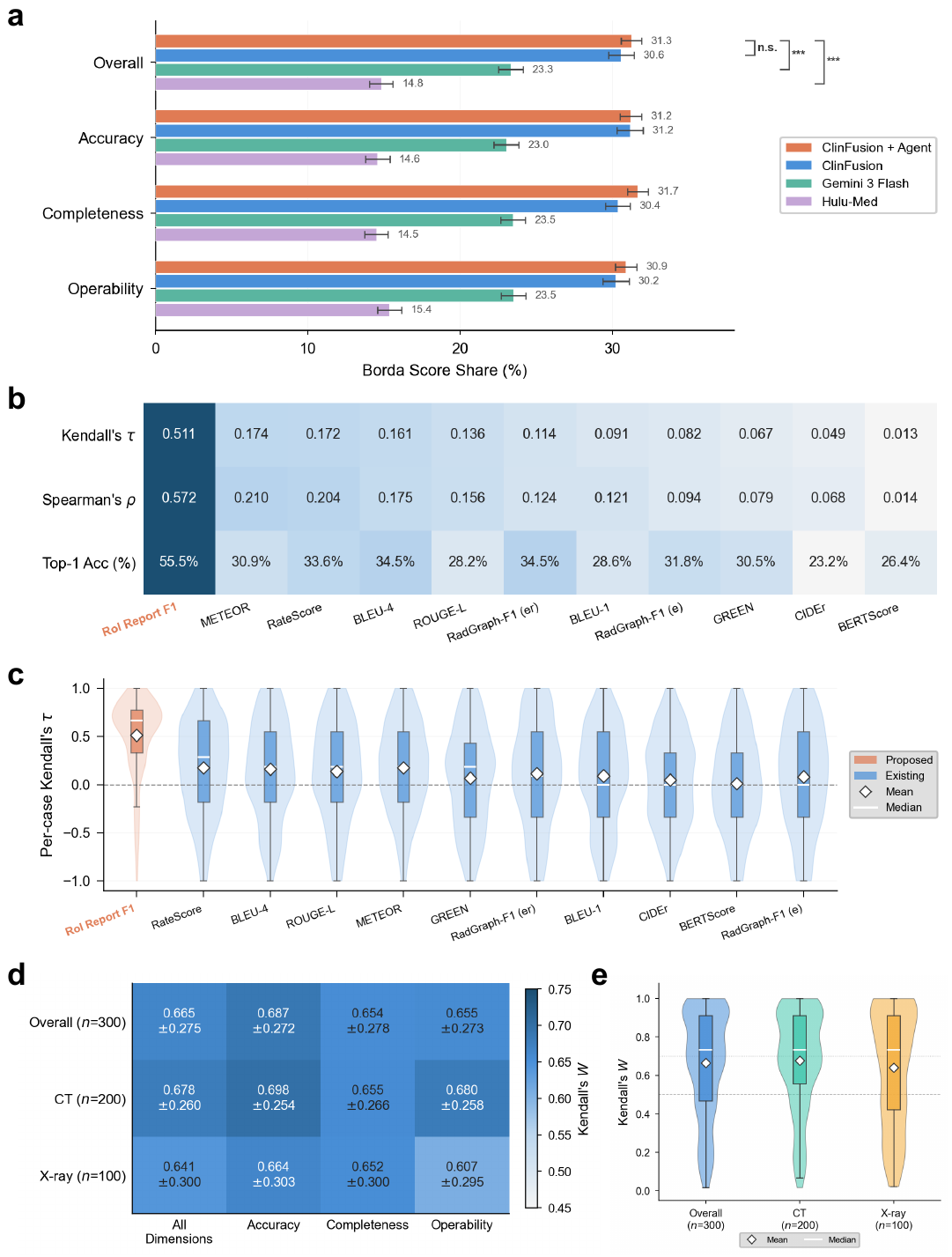}
\captionof{figure}{\textbf{Expert radiologist evaluation.}
\textbf{a},  Expert radiologist model ranking.
Aggregated Borda score shares from radiologists across three original evaluation dimensions (Accuracy, Completeness, Operability) and one aggregated dimension (Overall). 
Scores were aggregated using the Borda count method (score = 4 - rank).
Error bars denote standard error of the mean over per-case Borda shares.
Brackets indicate Wilcoxon signed-rank test (two-sided, paired by case) comparing ClinFusion + Agent with each competing model on the Overall dimension (n.s.: not significant; ***: $p < 0.001$).
\textbf{b},  Correlation between automated metrics and expert annotations.
Kendall's $\tau$, Spearman's $\rho$, and Top-1 Accuracy (the fraction of cases where the metric correctly identifies the doctor-preferred model) for each automatic metric computed against the radiologist consensus ranking. 
Metrics are sorted by Kendall's $\tau$ in descending order.
\textbf{c},  Per-case distribution of ranking agreement between automated metrics and doctor annotations.
Violin plots showing the distribution of per-case Kendall's $\tau$ between each automated metric and expert-annotated rankings. 
White diamonds and horizontal white lines indicate the mean and median, respectively.
\textbf{d},  Inter-annotator agreement: heatmap of Kendall's coefficient of concordance ($W$) stratified by imaging modality (CT, X-ray) and evaluation dimension. 
Values represent mean $W \pm s.d.$.
\textbf{e},  Inter-annotator agreement: violin plots showing the distribution of $W$ across individual cases. 
Box plots indicate the median (white horizontal line), interquartile range (box) and 1.5 x IQR (whiskers). 
White diamonds indicate the mean. 
Dashed lines indicate thresholds for moderate (W = 0.5) and strong (W = 0.7) agreement.
}
\label{fig:anno_expert_eval}
\end{center}
\clearpage

\clearpage
\begin{center}
\includegraphics[width=1.0\linewidth]{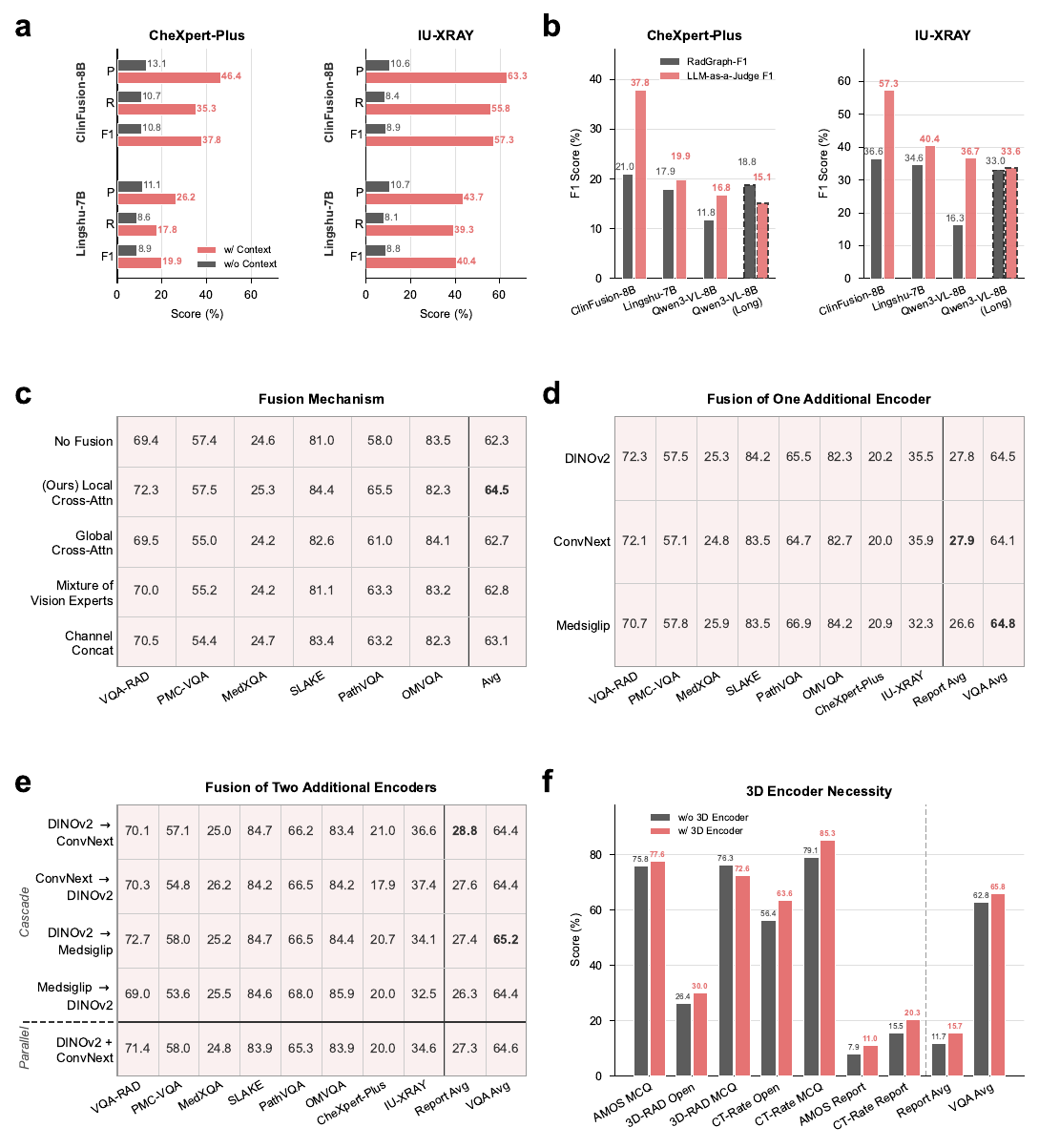}
\captionof{figure}{
\textbf{Ablation studies on the RoI-Grounded report generation evaluation and on the architectural design of \method.}
\textbf{a},  Ablation on the necessity of clinical context for report generation.
\method-8B and Lingshu-7B~\cite{xu2025lingshu} are evaluated under two settings: with the extracted clinical context (``w/ Context'') and without it (``w/o Context''), where the Clinical Indication and Area of Focus are removed from the generation prompt.
Precision, Recall, and F1 are reported on CheXpert-Plus and IU-XRAY.
\textbf{b},  Comparison between RadGraph-F1 and our LLM-as-a-Judge metric on CheXpert-Plus and IU-XRAY across four model variants, including a Qwen3-VL-8B (Long Output) variant prompted to encourage verbose generation.
RadGraph-F1 reports entity-relation F1, while the LLM-as-a-Judge reports F1 for clarity.
\textbf{c},  Ablation on different fusion mechanisms for the compositional vision encoder.
A single-encoder baseline (Qwen ViT) is compared against a dual-encoder setup (Qwen ViT + DINOv2) using channel concatenation, global cross-attention, mixture of vision experts, and our local cross-attention (CaSL Fusion).
Accuracy (\%) is reported on six 2D medical VQA benchmarks.
\textbf{d},  Ablation on the choice of the one additional specialist 2D encoder.
The base Qwen ViT is augmented with one additional encoder (DINOv2, ConvNext, or Medsiglip) and evaluated on 2D VQA (accuracy, \%) and report generation.
\textbf{e},  Ablation on the choice of multiple additional 2D encoders, the multi-encoder fusion strategy, and the fusion order.
The base Qwen ViT is augmented with two additional encoders combined either through Cascade fusion (with different orderings) or Parallel fusion, and evaluated on 2D VQA and report generation.
\textbf{f},  Ablation on the necessity of the native 3D encoder during inference.
The full model is compared against an ablated version where the 3D encoder is deactivated, forcing the model to rely solely on sampled 2D slices.
Accuracy (\%) is reported on 3D VQA benchmarks (AMOS, 3D-RAD, CT-Rate) and Precision, Recall, and F1 are reported on 3D report generation (AMOS Report, CT-Rate Report).
}
\label{fig:roi_report_ablation_architecture_ablation}
\end{center}
\clearpage

\begin{figure}[!t]
\centering
\includegraphics[width=1.0\linewidth]{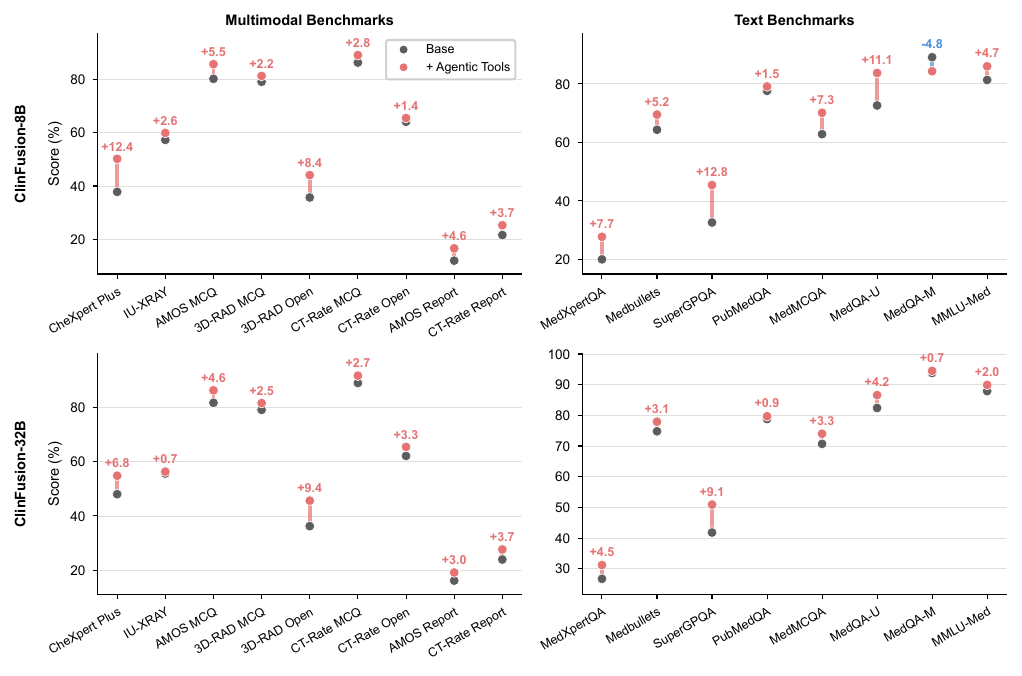}
\caption{
\textbf{Performance gains of \method when augmented with agentic tool use.}
For each benchmark, grey and red dots denote the Base and \mbox{+Agentic Tools} scores respectively, connected by a vertical segment, with the absolute difference annotated above (red for positive, blue for negative).
Rows correspond to \method-8B (top) and \method-32B (bottom); columns correspond to multimodal benchmarks (left) and text-only medical benchmarks (right).
For VQA and exam benchmarks scores are accuracy (\%); for report generation scores are F1 (\%) under the RoI-Grounded protocol.
}
\label{fig:agent_improvement}
\end{figure}

%% file: content/09_ExtendedData.tex
\beginextendeddata

\section*{Extended Data}


\begin{figure}[!htbp]
\centering
\includegraphics[width=1.0\linewidth]{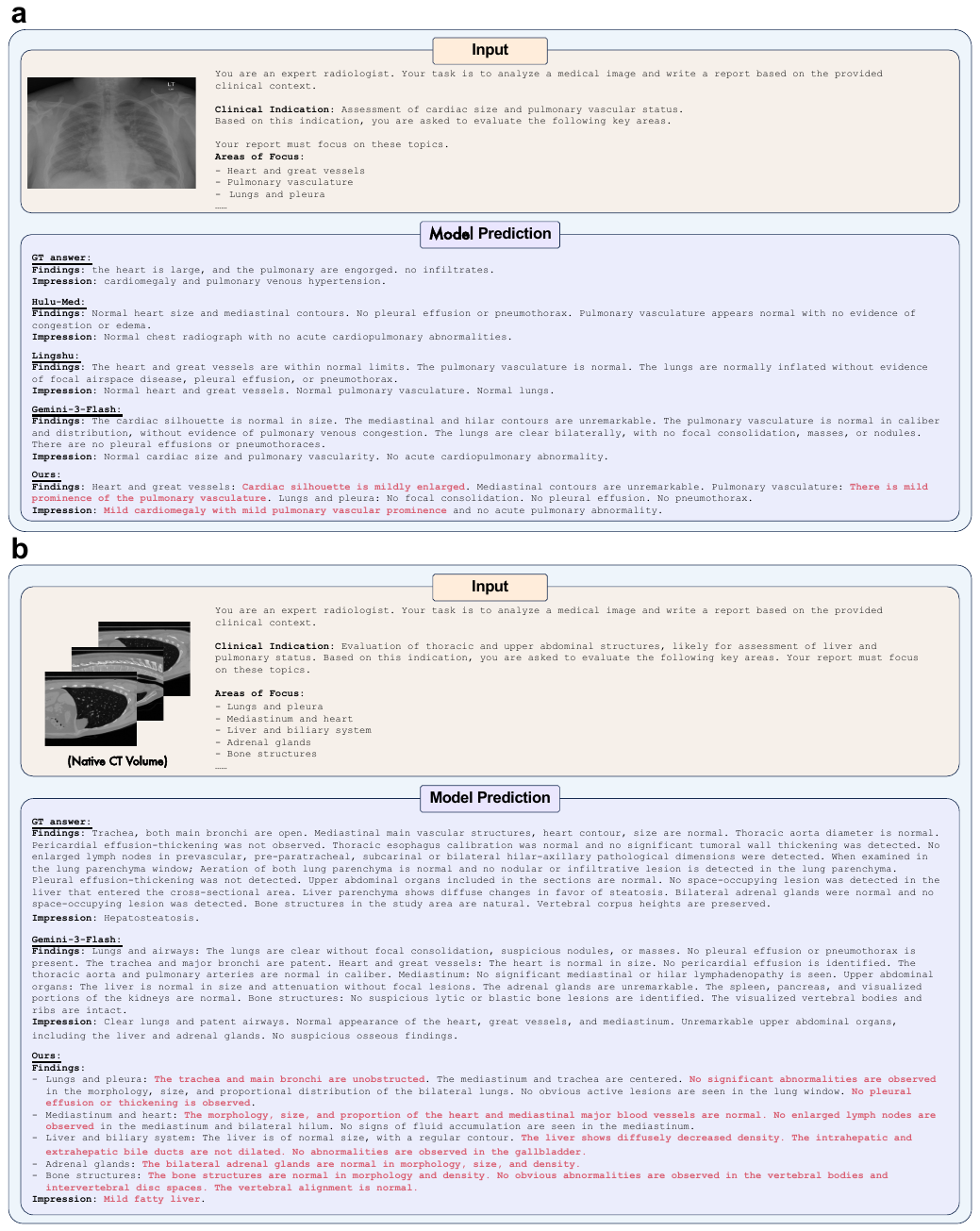}
\caption{
\textbf{Case visualization of 2D and 3D report generation} comparing \method with other methods.
\textbf{a}, 2D report generation case. 
\textbf{b}, 3D report generation case.
The input prompt directs the MLLMs to focus on specific ROIs.
}
\label{ext-fig:2d-3d-report-case}
\end{figure}

\begin{figure}[!t]
\centering
\includegraphics[width=1.0\linewidth]{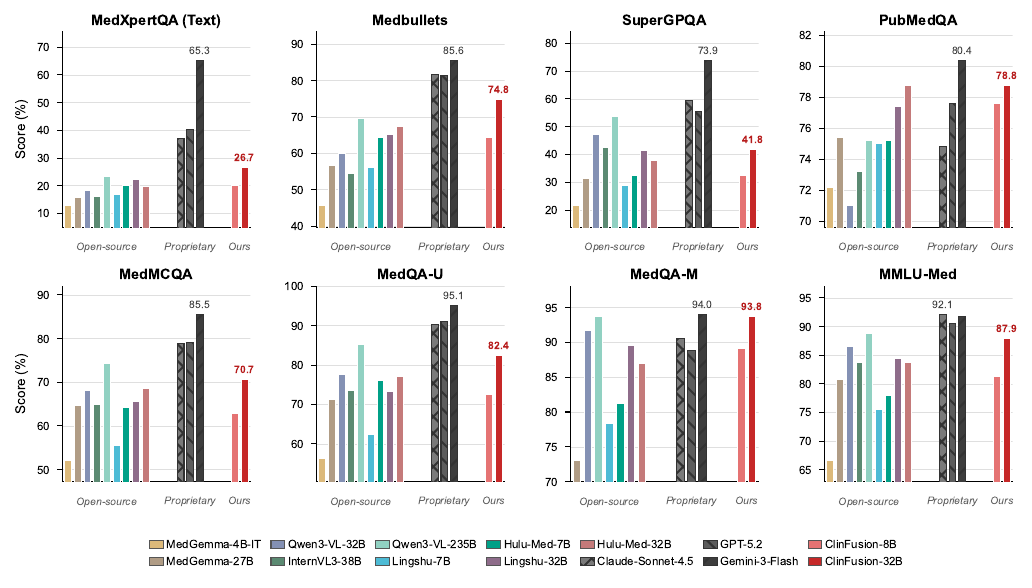}
\caption{
\textbf{Performance of \method on textual medical benchmarks.}
Results on eight text-only benchmarks spanning complex reasoning (MedXpertQA-Text, Medbullets, SuperGPQA), text understanding (PubMedQA), and medical examinations (MedMCQA, MedQA-USMLE, MedQA-MCMLE, MMLU-Medical). 
Models are compared across open-source (general-purpose and medical MLLMs), proprietary, and our models in a fair protocol re-benchmarked by us. 
}
\label{ext-fig:text_benchmarks}
\end{figure}

\begin{figure}[!t]
\centering
\includegraphics[width=1.0\linewidth]{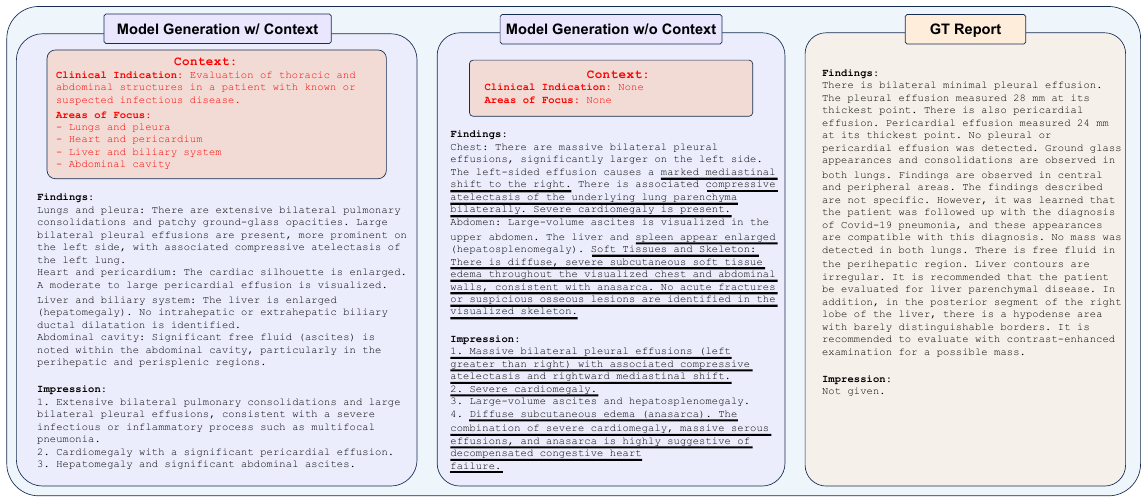}
\caption{
\textbf{Comparison of report generation with and without clinical context on the same CT scan.} 
We reuse the example from~\cref{sec:roi_grounded_report_generation}. 
Right: The GT report.
Left: The output of \method when provided with the extracted clinical context (Clinical Indication and Area of Focus).
Middle: the output of the same model without any clinical context.
\underline{Underlined text} indicates claims that are \emph{not verifiable} against the GT.
}
\label{ext-fig:case_without_context}
\end{figure}

\begin{figure}[!t]
\centering
\includegraphics[width=1.0\linewidth]{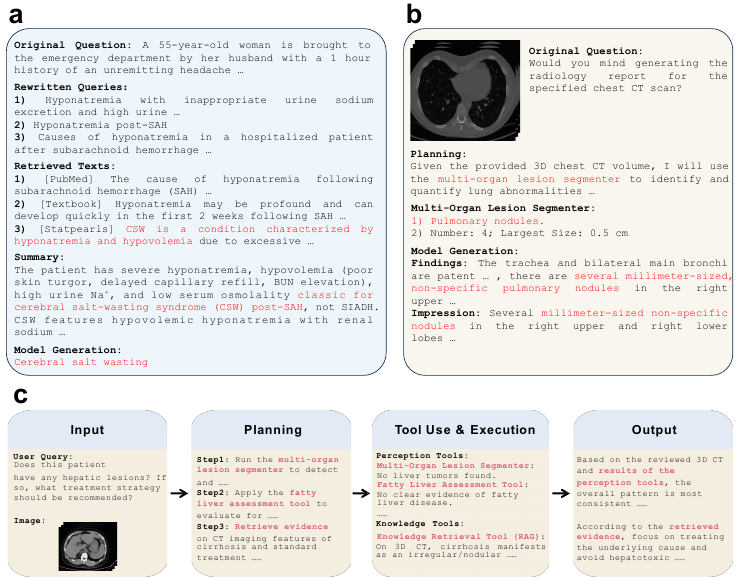}
\caption{\textbf{Agentic tool use: case studies and pipeline overview.}
\textbf{a}, Text-only task: The agent rewrites the original clinical question into focused retrieval queries, invokes the hybrid retrieval tool to collect provenance-aware evidence (\textit{e.g.}, PubMed/textbooks/StatPearls), and then synthesizes an evidence-grounded diagnosis in the final generation. 
\textbf{b}, Multimodal task: Radiology report generation from a 3D chest CT volume. The agent performs modality-aware routing, selects a Multi-Organ Lesion Segmenter to produce structured, quantitative findings (\textit{e.g.}, pulmonary nodule count and size), and then integrates the verified tool outputs into a radiology-style report.
\textbf{c}, Pipeline: Given multimodal clinical inputs (text and medical images), \method performs task-aware planning to select and invoke appropriate tools (including knowledge tools and perception expert tools), aggregates the returned tool outputs, and generates the final clinical response. 
In the illustrated liver CT case, \method executes a three-step workflow by calling a Multi-Organ Lesion Segmenter, a Fatty Liver Assessment Tool, and a Knowledge Retrieval Tool (RAG), and then integrates their results to produce a coherent case analysis and evidence-grounded management recommendations.
}
\label{ext-fig:agent}
\end{figure}

\begin{figure}[!t]
\centering
\includegraphics[width=1.0\linewidth]{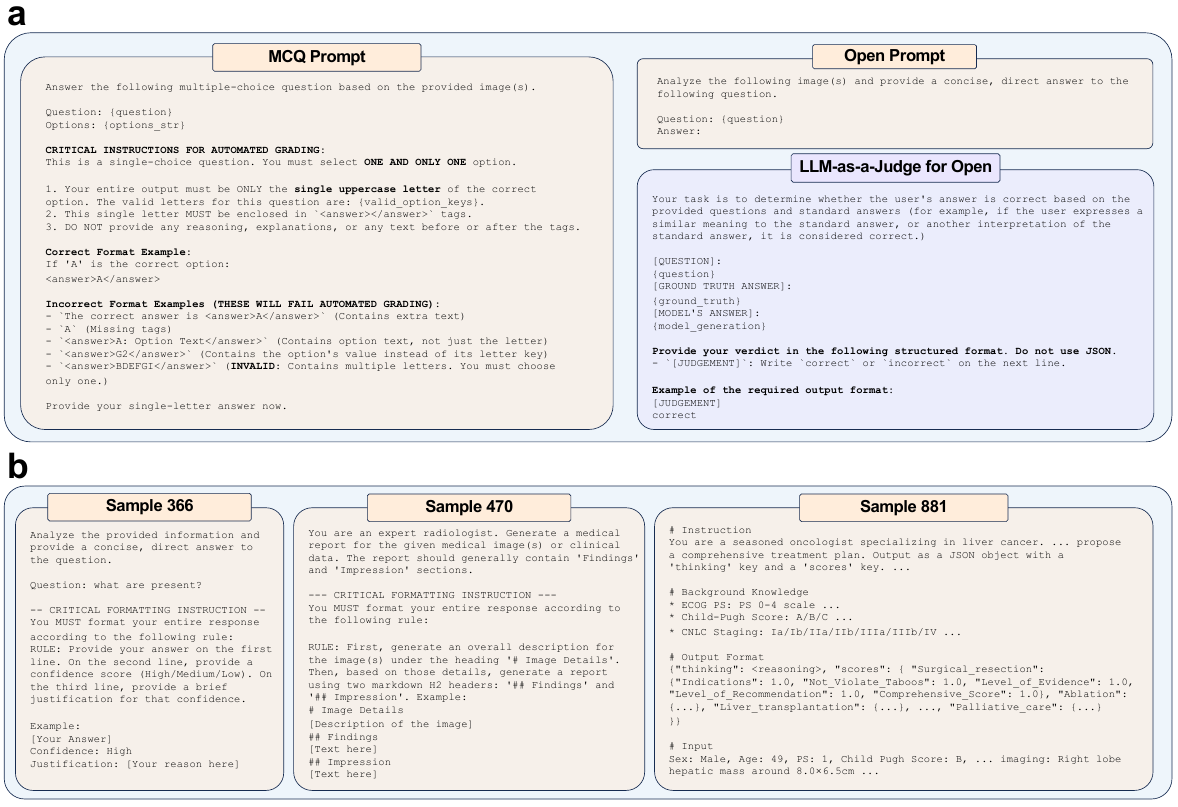}
\caption{\textbf{Evaluation prompt templates and MedIF-Bench examples.}
\textbf{a}, Prompt templates used for VQA evaluation. 
Left: The MCQ prompt enforces a strict single-letter output format enclosed in \texttt{<answer></answer>} tags, with explicit correct and incorrect format examples to ensure reliable rule-based extraction. 
Upper right: The open-ended prompt requests a concise, direct answer. 
Lower right: The LLM-as-a-Judge prompt used to evaluate open-ended responses by comparing the model's answer against the GT answer for semantic correctness.
\textbf{b}, Three examples from MedIF-Bench (prompts are abbreviated for display; full prompts are substantially longer). 
Left (Sample 366): An open-ended VQA task where the model must structure its answer on the first line, followed by a confidence score and justification on subsequent lines, representing clinical workflows that require both a concise answer and an explicit confidence assessment. 
Middle (Sample 470): A report generation task where the model must produce a radiology report with a \texttt{\# Image Details} section followed by markdown \texttt{\#\#} headers for ``Findings'' and ``Impression,'' mirroring standardized institutional reporting formats. 
Right (Sample 881): A SEER-based treatment planning task where the model must output a complex nested JSON object containing a reasoning trace and structured scores for multiple treatment modalities, representing integration with clinical decision-support systems.
}
\label{ext-fig:eval-prompt-ifbench}
\end{figure}

\begin{figure}[!t]
\centering
\includegraphics[width=1.0\linewidth]{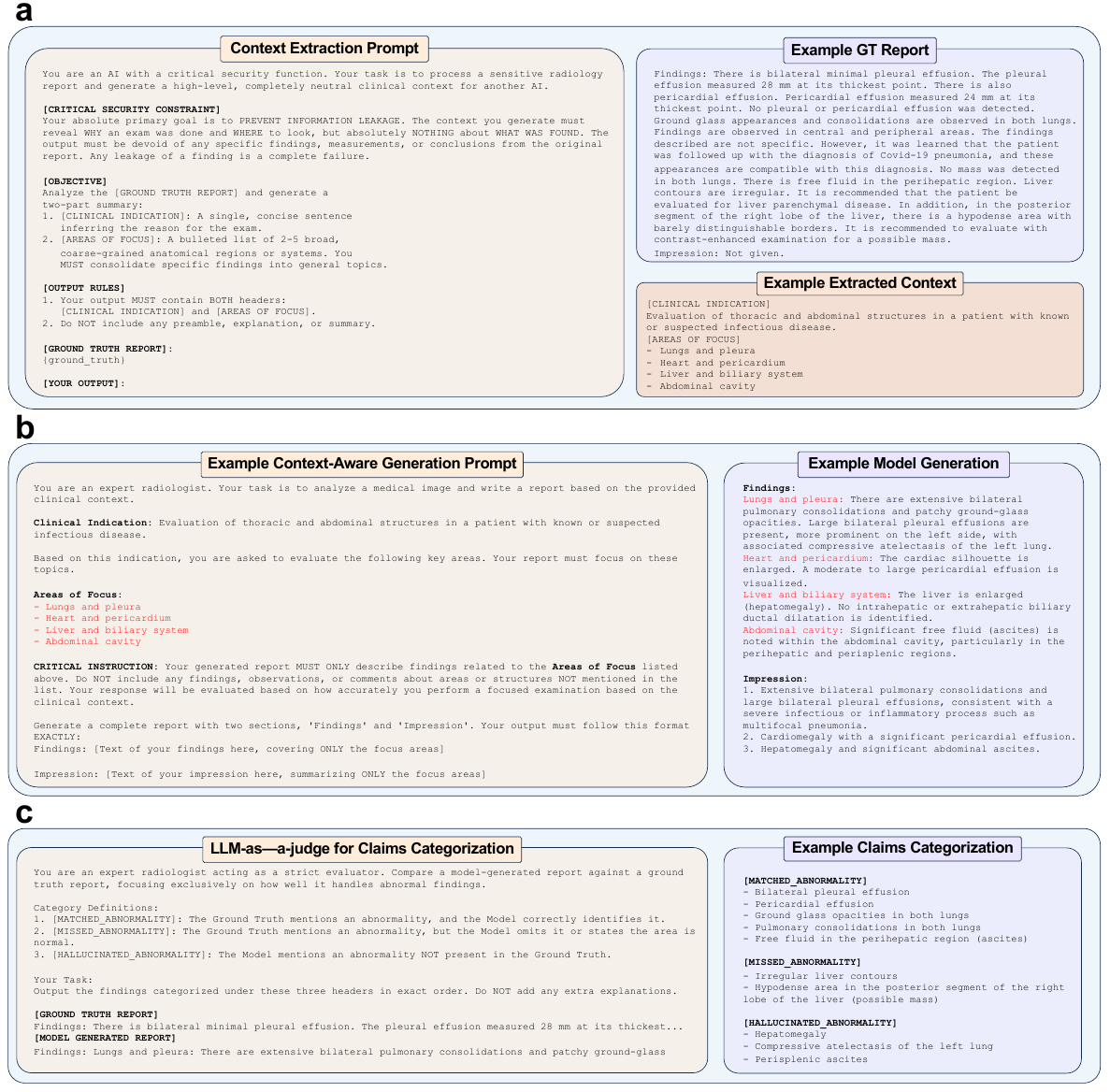}
\caption{\textbf{ROI-Grounded report generation evaluation pipeline.}
\textbf{a}, Context extraction procedure via GPT 4.1 API. 
The left panel displays the prompt for context extraction, while the right panels present an actual example of an input GT report (top right) and its corresponding extracted context (bottom right). 
This procedure is designed to minimize leakage of actual diagnostic answers and reveals only the regions of interest.
\textbf{b}, Context-conditioned report generation. 
The context successfully extracted in panel (a) is utilized here to condition the generation process. 
This extracted context is combined with the instruction prompt and presented to our MLLM, strictly constraining the model to generate findings exclusively for the provided Regions of Interest.
\textbf{c}, LLM-as-a-Judge evaluation. 
The left panel shows the evaluation prompt presented to the LLM judge (abbreviated for presentation), which strictly defines the categorization rules. 
The right panel displays the actual categorized output, evaluating the MLLM-generated report from panel (b) against the original GT report from panel (a). 
In this specific case, the True Positives ($TP$, matched) are 5, False Negatives ($FN$, missed) are 2, and False Positives ($FP$, hallucinated) are 3. 
Consequently, the metrics are computed as: Precision $= 5/(5+3) = 0.6250$, Recall $= 5/(5+2) = 0.7143$, and F1-score $= 2 \times (0.6250 \times 0.7143) / (0.6250 + 0.7143) = 0.6667$.
}
\label{ext-fig:roi-pipeline}
\end{figure}

\begin{figure}[!t]
\centering
\includegraphics[width=1.0\linewidth]{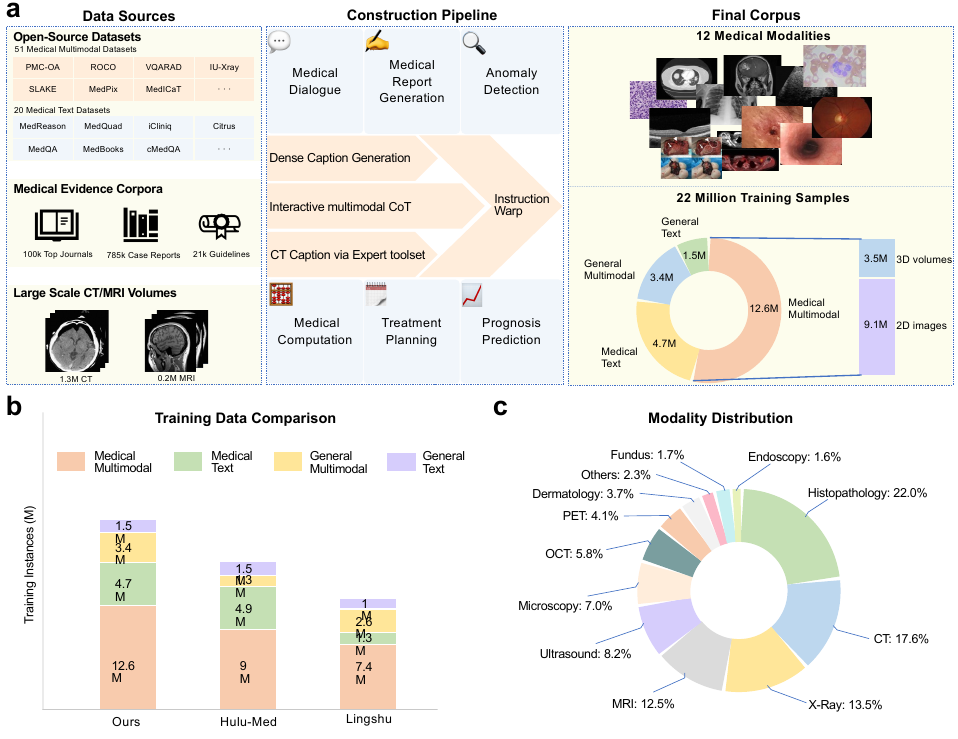}
\caption{
\textbf{Overview of data curation for \method}. 
\textbf{a}, Data sources and construction pipeline. 
We collect open-source datasets, medical evidence corpora, and large-scale CT/MRI volumes. These data are then processed through the Dense Caption Generation, Medical Multimodal CoT, and Large-Scale CT Caption pipelines, yielding 22.2M training samples spanning 12 medical imaging modalities and supporting diverse clinical tasks, including medical dialogue, medical report generation, anomaly detection, medical computation, treatment planning and prognosis prediction.
\textbf{b}, Training data scale comparison among \method, Hulu-Med~\cite{jiang2025hulu-med}, and Lingshu~\cite{xu2025lingshu}, decomposed into medical multimodal, medical text, general multimodal, and general text instances. 
\textbf{c}, Modality distribution of medical images in the final corpus.
}
\label{ext-fig:data_curation}
\end{figure}


\input{tables/data_sources}

\input{tables/benchmark_summary}

\input{tables/training_stages}

\input{tables/data_access}

%% file: tables/data_sources.tex

\begin{table*}[!htbp]
\centering
\scriptsize
\caption{\textbf{Data sources used for \method training}. We report an approximate number of training items for each data type in \textbf{Scale}. * denotes datasets constructed by our data synthetic pipeline.}
\label{ext-tab:data_sources_v260}
\setlength{\tabcolsep}{5pt}
\renewcommand{\arraystretch}{1.18}

\newcolumntype{Y}{>{\RaggedRight\arraybackslash}X}

\begin{tabularx}{\textwidth}{c c Y r}
\toprule
\textbf{Stage} & \textbf{Modality} & \multicolumn{1}{c}{\textbf{Datasets}} & \textbf{Scale} \\
\midrule

\multirow[t]{2}{*}{\textbf{S1}} &
\textbf{Medical multimodal} &
\parbox[t]{\linewidth}{\RaggedRight\fontsize{6.8}{8}\selectfont
PMC-OA~\cite{lin2023pmc};
ROCO~\cite{pelka2018roco};
ROCOv2~\cite{ruckert2024rocov2};
PubMedVision~\cite{chen2024pubmedvision};
MIMIC-CXR~\cite{johnson2019mimiccxr};
IU-Xray~\cite{demnerfushman2016openi};
open-i~\cite{demnerfushman2016openi};
MedICaT~\cite{subramanian2020medicat};
MedPix-2.0~\cite{siragusa2025medpix20};
BIOMEDICA Clinical Subset~\cite{lozano2025biomedicaopenbiomedicalimagecaption};
BIOMEDICA Dermatology Subset~\cite{lozano2025biomedicaopenbiomedicalimagecaption};
BIOMEDICA Histopathology Subset~\cite{lozano2025biomedicaopenbiomedicalimagecaption};
BIOMEDICA Microscopy Subset~\cite{lozano2025biomedicaopenbiomedicalimagecaption};
BIOMEDICA Surgery Subset~\cite{lozano2025biomedicaopenbiomedicalimagecaption};
Synthetic Case Report Caption$^{*}$;
Synthetic Journal Caption$^{*}$.
} &
$\sim$3.9M \\
& \textbf{General multimodal} &
\parbox[t]{\linewidth}{\RaggedRight\fontsize{6.8}{8}\selectfont
Llava-v1.5-Caption~\cite{liu2023visualinstructiontuning}; 
PixMo~\cite{deitke2024molmo}.
} &
$\sim$2.3M \\
\addlinespace
\midrule

\multirow[t]{3}{*}{\textbf{S2}} &
\textbf{Medical multimodal} &
\parbox[t]{\linewidth}{\RaggedRight\fontsize{6.8}{8}\selectfont
\textit{Same as S1 Medical Multimodal datasets.}
} &
$\sim$3.8M \\
& \textbf{General multimodal} &
\parbox[t]{\linewidth}{\RaggedRight\fontsize{6.8}{8}\selectfont
\textit{Same as S1 General multimodal datasets.}
} &
$\sim$2.3M \\
\addlinespace
\midrule

\multirow[t]{4}{*}{\textbf{S3}} &
\textbf{Medical multimodal} &
\parbox[t]{\linewidth}{\RaggedRight\fontsize{6.8}{8}\selectfont
LLaVA-Med~\cite{li2023llavamed};
Quilt-LLaVA~\cite{Seyfioglu_2024_CVPR};
FairVLMed;
CheXpert Plus~\cite{chambon2024CheXpertPlus};
Kvasir-VQA~\cite{gautam2024kvasirvqa}; \\
LLaVA-1.5-Instruct~\cite{liu2023visualinstructiontuning};
MIMIC-Ext-MIMICCXR-VQA~\cite{bae2024mimicext}; \\
PathVQA~\cite{he2020pathvqa};
PMC-VQA~\cite{zhang2023pmcvqa};
SLAKE~\cite{liu2021slake};
VQA-Med-2019~\cite{benabacha2019vqamed};
VQA-RAD~\cite{lau2018vqarad}; \\
PMC-VQA-2-recaptions;
SLAKE-recaptions;
VQA-Med-2019-recaptions;
VQA-RAD-recaptions;
Path-VQA-recaptions; \\
GMAI-Reasoning10K~\cite{gmai_reasoning10k_arxiv};
Synthetic Medical OCR$^{*}$;
Private CT$^{*}$;
Synthetic Case Report VQA$^{*}$;
Synthetic Journal VQA$^{*}$.
} &
$\sim$5.3M \\
& \textbf{General multimodal} &
\parbox[t]{\linewidth}{\RaggedRight\fontsize{6.8}{8}\selectfont
Llava-v1.5-MM~\cite{liu2023visualinstructiontuning};
ALLaVA~\cite{chen2024allava}.
} &
$\sim$1.1M \\
& \textbf{Medical text} &
\parbox[t]{\linewidth}{\RaggedRight\fontsize{6.8}{8}\selectfont
MedQuAD~\cite{benabacha2019medquad};
medical-o1-verifiable-problem~\cite{chen2024huatuogpto1};
medical-o1-reasoning-SFT;
ApolloCorpus~\cite{wang2024apollo};
Medical-R1-Distill-Data; \\
AlpaCare-MedInstruct-52k~\cite{zhang2025alpacare};
HealthCareMagic-100k~\cite{hf_healthcaremagic_100k};
PMC-LLaMA~\cite{wu2024pmc};
HuatuoGPT2-GPT4-SFT-140K~\cite{chen2023huatuogptii}; \\
MedReason~\cite{wu2025medreason};
MedThoughts-8K~\cite{medthoughts2025};
MedQA~\cite{jin2021medqa_usmle};
open-book-CMB-exam~\cite{hf_openbook_cmb_exam};
cMedQA~\cite{hf_cmedqa_v20};
HealthCareMagic~\cite{hf_healthcaremagic_100k};
iCliniq-10K~\cite{hf_healthcaremagic_100k};
Citrus\_S3~\cite{wang2025citrus}; \\
MedBooks-18-CoT~\cite{kim2024medbooks18cot};
Synthetic Case Report Long CoT$^{*}$.
} &
$\sim$4.7M \\
& \textbf{General text} &
\parbox[t]{\linewidth}{\RaggedRight\fontsize{6.8}{8}\selectfont
Llava-v1.5-text-subset~\cite{liu2023visualinstructiontuning};
OpenHermes-2.5~\cite{openhermes25}.
} &
$\sim$1.5M \\
\addlinespace
\midrule

\multirow[t]{1}{*}{\textbf{S4}} &
\textbf{Medical multimodal} &
\parbox[t]{\linewidth}{\RaggedRight\fontsize{6.8}{8}\selectfont
CT-Rate-report~\cite{hamamci2024CT-Rate};
Merlin-report~\cite{blankemeier2024merlin};
RadGenome-ChestCT~\cite{zhang2024radgenomechestct};
M3D-CAP~\cite{bai2024m3d};
OLD-INSPECT~\cite{huang2023inspect};
AMOS-MM-report~\cite{amosmm2024_zenodo}; \\
AbdomenCT$^{*}$; ChronicCT$^{*}$; CardiovascularCT$^{*}$; PulmonaryNodulesCT$^{*}$.
} &
$\sim$1.3M \\
\addlinespace
\midrule

\multirow[t]{4}{*}{\textbf{S5}} &
\textbf{Medical multimodal} &
\parbox[t]{\linewidth}{\RaggedRight\fontsize{6.8}{8}\selectfont
\textit{All S3 medical multimodal datasets} \;+\;
AMOS-MM-VQA~\cite{amosmm2024_zenodo};
CT-Rate-VQA~\cite{hamamci2024CT-Rate};
3D-RAD~\cite{gai2025_3drad};
Synthetic-M3d-VQA$^{*}$;
Merlin-MCQ~\cite{blankemeier2024merlin};
Merlin-VQA~\cite{blankemeier2024merlin}.
} &
$\sim$6.4M \\
& \textbf{General multimodal} &
\parbox[t]{\linewidth}{\RaggedRight\fontsize{6.8}{8}\selectfont
\textit{Same as S3 General multimodal datasets.}
} &
$\sim$1.1M \\
& \textbf{Medical text} &
\parbox[t]{\linewidth}{\RaggedRight\fontsize{6.8}{8}\selectfont
\textit{Same as S3 Medical text datasets.}
} &
$\sim$4.7M \\
& \textbf{General text} &
\parbox[t]{\linewidth}{\RaggedRight\fontsize{6.8}{8}\selectfont
\textit{Same as S3 General text datasets.}
} &
$\sim$1.5M \\
\bottomrule
\end{tabularx}
\end{table*}

%% file: tables/benchmark_summary.tex
\begin{table*}[!htbp]
    \centering
\caption{\textbf{Summary of evaluation benchmarks.} We categorize our evaluation framework into five key dimensions: 2D multimodal medical VQA, text-only medical knowledge, 3D volumetric medical VQA, clinical report generation, and instruction-following (\# indicates sample counts). Under the Modality column, \textit{Single} and \textit{Multi} distinguish between single-image and multi-image inputs; notably, for 3D, report generation, and instruction-following tasks, the model typically processes a single multimodal file (\textit{e.g.}, one 3D volume or one 2D image) per instance. Task types are classified as Multiple Choice (MCQ), Open-ended (Open), Clinical Report Generation (Report), or Instruction-Following (IF). $\dagger$ denotes a manually curated subset of the original dataset, filtered to ensure data quality and evaluation fidelity.}
    \label{ext-tab:benchmark_summary}
    \footnotesize 
    \setlength{\tabcolsep}{0pt} 
    \begin{tabular*}{\linewidth}{@{\extracolsep{\fill}}llclll}
        \toprule
        \textbf{Benchmark} & \textbf{Modality/Genre} & \textbf{Scale (\#)} & \textbf{Type} & \textbf{Lang.} & \textbf{Source} \\
        \midrule
        \multicolumn{6}{c}{\textit{2D Multimodal Medical VQA (Total: 141,587)}} \\
        \midrule
        VQA-RAD & Radiology (Single) & 451 & MCQ \& Open & EN & \cite{lau2018vqarad} \\
        SLAKE & Radiology (Single) & 2,094 & MCQ \& Open & EN \& CN & \cite{liu2021slake} \\
        PathVQA & Pathology (Single) & 6,719 & MCQ \& Open & EN & \cite{he2020pathvqa} \\
        PMC-VQA & Diverse (Single) & 33,430 & MCQ & EN & \cite{zhang2023pmcvqa} \\
        OmniMedVQA & Diverse (Single) & 88,996 & MCQ & EN & \cite{hu2024omnimedvqa} \\
        MedFrameQA & Diverse (Multi) & 2,851 & MCQ & EN & \cite{yu2025MedFrameQA} \\
        GMAI-MMBench & Diverse (Single) & 4,550 & MCQ & EN & \cite{ye2024Gmai-mmbench} \\
        MedXpertQA (MM) & Diverse (Multi) & 2,000 & MCQ & EN & \cite{zuo2025medxpertqa} \\
        \midrule
        \multicolumn{6}{c}{\textit{Text-only Medical Knowledge (Total: 17,074)}} \\
        \midrule
        MedQA (USMLE) & Clinical Exam & 1,273 & MCQ & EN & \cite{jin2021medqa_usmle} \\
        MedQA (MCMLE) & Clinical Exam & 3,426 & MCQ & CN & \cite{jin2021medqa_usmle} \\
        MedMCQA & Entrance Exam & 4,183 & MCQ & EN & \cite{pal2022medmcqa} \\
        MMLU (Medical) & Multi-discipline & 1,871 & MCQ & EN & \cite{hendrycks2021mmlu} \\
        PubMedQA & Research QA & 500 & MCQ & EN & \cite{jin2019pubmedqa} \\
        MedBullets & Clinical Knowledge & 616 & MCQ & EN & \cite{chen2025medbullets} \\
        SuperGPQA & General Practice & 2,755 & MCQ & EN & \cite{pteam2025supergpqa} \\
        MedXpertQA (Text) & Diverse & 2,450 & MCQ & EN & \cite{zuo2025medxpertqa} \\
        \midrule
        \multicolumn{6}{c}{\textit{3D Volumetric Medical VQA (Total: 48,907)}} \\
        \midrule
        CT-Rate-VQA\textsuperscript{$\dagger$} & Non-Contrast Chest CT & 12,210 & MCQ \& Open & EN & \cite{hamamci2024CT-Rate} \\
        3D-RAD & Non-Contrast Chest CT & 33,910 & MCQ \& Open & EN & \cite{gai2025_3D-RAD} \\
        AMOS-MM\textsuperscript{$\dagger$} & Abdominal CT & 2,787 & MCQ & EN & \cite{ji2022AMOS} \\
        \midrule
        \multicolumn{6}{c}{\textit{2D/3D Clinical Report Generation (Total: 3,935)}} \\
        \midrule
        IU-Xray & Chest X-ray & 296 & Report & EN & \cite{demner2016IUXRay} \\
        CheXpert-Plus & Chest X-ray & 200 & Report & EN & \cite{chambon2024CheXpertPlus} \\
                CT-Rate & Non-Contrast Chest CT & 3,039 & Report & EN & \cite{hamamci2024CT-Rate} \\
        AMOS-MM & Abdominal CT & 400 & Report & EN & \cite{ji2022AMOS} \\
        \midrule
        \multicolumn{6}{c}{\textit{Medical Instruction Following}} \\
        \midrule
        {MedIF-Bench} & Mixed & 900 & IF & EN & {Ours} \\
        \bottomrule
    \end{tabular*}
\end{table*}

%% file: tables/training_stages.tex
\begin{table*}[t]
\centering
\caption{\textbf{The progressive training strategy for \method}. 
``Trainable Components'' lists the parts of the model being updated in each stage.}
\label{ext-tab:training_stages}
\resizebox{\textwidth}{!}{%
\begin{tabular}{clll}
\toprule
\textbf{Stage} & \textbf{Objective} & \textbf{Training Data} & \textbf{Trainable Components} \\
\midrule
\multirow{2}{*}{{\Large \ding{172}}} & \textbf{Shallow multi-encoder alignment (2D)} & 2D Image-Text Pairs & All Encoders, 2D Fusion Modules \\
   & \textit{Align specialist 2D encoders.} & & \\
\midrule
\multirow{2}{*}{{\Large \ding{173}}} & \textbf{Deep vision-language alignment (2D)} & 2D Image-Text Pairs & Full Model (All Encoders, 2D Fusion Modules, LLM) \\
   & \textit{Align fused 2D features with LLM.} & & \\
\midrule
\multirow{2}{*}{{\Large \ding{174}}} & \textbf{2D Instruction Tuning} & 2D Instruction Datasets & Full Model (All Encoders, 2D Fusion Modules, LLM) \\
   & \textit{Teach 2D task execution.} & & \\
\midrule
\multirow{2}{*}{{\Large \ding{175}}} & \textbf{Rapid 3D encoder alignment} & 3D Volume-Text Pairs & 3D Encoder, 3D Fusion Modules \\
   & \textit{Efficiently align 3D encoder.} & & \\
\midrule
\multirow{2}{*}{{\Large \ding{176}}} & \textbf{Holistic 2D/3D instruction tuning} & 2D/3D Instruction Datasets & Full Model (All Encoders, All Fusion Modules, LLM) \\
   & \textit{Unlock full 2D/3D capabilities.} & & \\
\bottomrule
\end{tabular}
}
\end{table*}

%% file: tables/data_access.tex
\begin{table*}[t]
\centering
\scriptsize
\setlength{\tabcolsep}{2.5pt}
\renewcommand{\arraystretch}{0.92}
\providecommand{\fiturl}[1]{\mbox{\url{#1}}}
\caption{
\textbf{Dataset availability, download links, and access terms for all data sources used in \method training}. 
Datasets are grouped by training stages described in \cref{sec:progressive_training}.
The Access column reports the verified license, access category, or source-license inheritance where applicable.}
\label{tab:clinfusion_data_access}
\definecolor{StageOne}{HTML}{d7f9f8}
\definecolor{StageTwo}{HTML}{ffffea}
\definecolor{StageThree}{HTML}{fff0d4}
\definecolor{StageFour}{HTML}{eef4ff}
\definecolor{StageFive}{HTML}{f4ecff}
\makebox[\textwidth][c]{%
\adjustbox{max width=1.0\textwidth}{
\begin{tabular}{l l l}
\toprule
\textbf{Dataset Name} & \textbf{Link} & \textbf{Access} \\
\midrule
\rowcolor{StageOne}\multicolumn{3}{l}{\textbf{Application Stage: S1 -- Medical multimodal}} \\
\rowcolor{StageOne} PMC-OA & \fiturl{https://huggingface.co/datasets/axiong/pmc_oa} & Under CC \\
\rowcolor{StageOne} ROCO & \fiturl{https://github.com/razorx89/roco-dataset} & Under CC \\
\rowcolor{StageOne} ROCOv2 & \fiturl{https://huggingface.co/datasets/eltorio/ROCOv2-radiology} & CC BY-NC-SA 4.0 \\
\rowcolor{StageOne} PubMedVision & \fiturl{https://huggingface.co/datasets/FreedomIntelligence/PubMedVision} & Apache 2.0 \\
\rowcolor{StageOne} MIMIC-CXR & \fiturl{https://physionet.org/content/mimic-cxr/} & PhysioNet License \\
\rowcolor{StageOne} IU-Xray & \fiturl{https://openi.nlm.nih.gov/} & Open Access \\
\rowcolor{StageOne} open-i & \fiturl{https://openi.nlm.nih.gov/} & Open Access \\
\rowcolor{StageOne} MedICaT & \fiturl{https://github.com/allenai/medicat} & Under CC \\
\rowcolor{StageOne} MedPix-2.0 & \fiturl{https://github.com/CHILab1/MedPix-2.0} & CC BY-NC-SA 4.0 \\
\rowcolor{StageOne} BIOMEDICA Clinical Subset & \fiturl{https://huggingface.co/datasets/BIOMEDICA/biomedica_webdataset_24M} & Under CC \\
\rowcolor{StageOne} BIOMEDICA Dermatology Subset & \fiturl{https://huggingface.co/datasets/BIOMEDICA/biomedica_webdataset_24M} & Under CC \\
\rowcolor{StageOne} BIOMEDICA Histopathology Subset & \fiturl{https://huggingface.co/datasets/BIOMEDICA/biomedica_webdataset_24M} & Under CC \\
\rowcolor{StageOne} BIOMEDICA Microscopy Subset & \fiturl{https://huggingface.co/datasets/BIOMEDICA/biomedica_webdataset_24M} & Under CC \\
\rowcolor{StageOne} BIOMEDICA Surgery Subset & \fiturl{https://huggingface.co/datasets/BIOMEDICA/biomedica_webdataset_24M} & Under CC \\
\rowcolor{StageOne} Synthetic Case Report Caption & Synthetic Data & Internal Synthetic Data \\
\rowcolor{StageOne} Synthetic Journal Caption & Synthetic Data & Internal Synthetic Data \\
\midrule
\rowcolor{StageOne}\multicolumn{3}{l}{\textbf{Application Stage: S1 -- General multimodal}} \\
\rowcolor{StageOne} LLaVA-v1.5-Caption & \fiturl{https://github.com/haotian-liu/LLaVA} & CC BY-NC 4.0 \\
\rowcolor{StageOne} PixMo & \fiturl{https://huggingface.co/datasets/allenai/pixmo-cap} & ODC-BY v1.0 \\
\midrule
\rowcolor{StageTwo}\multicolumn{3}{l}{\textbf{Application Stage: S2 -- Medical multimodal}} \\
\rowcolor{StageTwo} Same as S1 Medical multimodal datasets &  & Same access terms as S1 Medical multimodal \\
\midrule
\rowcolor{StageTwo}\multicolumn{3}{l}{\textbf{Application Stage: S2 -- General multimodal}} \\
\rowcolor{StageTwo} Same as S1 General multimodal datasets &  & Same access terms as S1 General multimodal \\
\midrule
\rowcolor{StageThree}\multicolumn{3}{l}{\textbf{Application Stage: S3 -- Medical multimodal}} \\
\rowcolor{StageThree} LLaVA-Med & \fiturl{https://github.com/microsoft/LLaVA-Med} & CC BY-NC 4.0 \\
\rowcolor{StageThree} Quilt-LLaVA & \fiturl{https://huggingface.co/datasets/wisdomik/QUILT-LLaVA-Instruct-107K} & CC BY-NC-ND 3.0 \\
\rowcolor{StageThree} FairVLMed & \fiturl{https://zenodo.org/records/13178701} & CC BY-NC-ND 4.0 \\
\rowcolor{StageThree} CheXpert Plus & \fiturl{https://aimi.stanford.edu/datasets/chexpert-plus} & Credentialed Access \\
\rowcolor{StageThree} Kvasir-VQA & \fiturl{https://datasets.simula.no/kvasir-vqa/} & CC BY-NC 4.0 \\
\rowcolor{StageThree} LLaVA-1.5-Instruct & \fiturl{https://github.com/haotian-liu/LLaVA} & CC BY-NC 4.0 \\
\rowcolor{StageThree} MIMIC-Ext-MIMICCXR-VQA & \fiturl{https://physionet.org/content/mimic-ext-mimic-cxr-vqa/1.0.0/} & PhysioNet License \\
\rowcolor{StageThree} PathVQA & \fiturl{https://huggingface.co/datasets/flaviagiammarino/path-vqa} & MIT \\
\rowcolor{StageThree} PMC-VQA & \fiturl{https://huggingface.co/datasets/RadGenome/PMC-VQA} & MIT \\
\rowcolor{StageThree} SLAKE & \fiturl{https://huggingface.co/datasets/BoKelvin/SLAKE} & CC BY 4.0 \\
\rowcolor{StageThree} VQA-Med-2019 & \fiturl{https://zenodo.org/records/10499039} & CC BY 4.0 \\
\rowcolor{StageThree} VQA-RAD & \fiturl{https://huggingface.co/datasets/flaviagiammarino/vqa-rad} & CC0-1.0 \\
\rowcolor{StageThree} PMC-VQA-2-recaptions & \fiturl{https://huggingface.co/datasets/RadGenome/PMC-VQA} & Internal Synthetic Data \\
\rowcolor{StageThree} SLAKE-recaptions & \fiturl{https://huggingface.co/datasets/BoKelvin/SLAKE} & Internal Synthetic Data \\
\rowcolor{StageThree} VQA-Med-2019-recaptions & \fiturl{https://zenodo.org/records/10499039} & Internal Synthetic Data \\
\rowcolor{StageThree} VQA-RAD-recaptions & \fiturl{https://huggingface.co/datasets/flaviagiammarino/vqa-rad} & Internal Synthetic Data \\
\rowcolor{StageThree} Path-VQA-recaptions & \fiturl{https://huggingface.co/datasets/flaviagiammarino/path-vqa} & Internal Synthetic Data \\
\rowcolor{StageThree} GMAI-Reasoning10K & \fiturl{https://huggingface.co/datasets/General-Medical-AI/GMAI-Reasoning10K} & CC BY 4.0 \\
\rowcolor{StageThree} Synthetic Medical OCR & Synthetic Data & Internal Synthetic Data \\
\rowcolor{StageThree} Synthetic Case Report VQA & Synthetic Data & Internal Synthetic Data \\
\rowcolor{StageThree} Synthetic Journal VQA & Synthetic Data & Internal Synthetic Data \\
\midrule
\rowcolor{StageThree}\multicolumn{3}{l}{\textbf{Application Stage: S3 -- General multimodal}} \\
\rowcolor{StageThree} LLaVA-v1.5-MM & \fiturl{https://github.com/haotian-liu/LLaVA} & CC BY-NC 4.0 \\
\rowcolor{StageThree} ALLaVA & \fiturl{https://huggingface.co/datasets/FreedomIntelligence/ALLaVA-4V} & Apache 2.0 \\
\midrule
\rowcolor{StageThree}\multicolumn{3}{l}{\textbf{Application Stage: S3 -- Medical text}} \\
\rowcolor{StageThree} MedQuAD & \fiturl{https://huggingface.co/datasets/lavita/MedQuAD} & Open Access \\
\rowcolor{StageThree} medical-o1-verifiable-problem & \fiturl{https://huggingface.co/datasets/FreedomIntelligence/medical-o1-verifiable-problem} & Apache 2.0 \\
\rowcolor{StageThree} medical-o1-reasoning-SFT & \fiturl{https://huggingface.co/datasets/FreedomIntelligence/medical-o1-reasoning-SFT} & Apache 2.0 \\
\rowcolor{StageThree} ApolloCorpus & \fiturl{https://huggingface.co/datasets/FreedomIntelligence/ApolloCorpus} & Apache 2.0 \\
\rowcolor{StageThree} Medical-R1-Distill-Data & \fiturl{https://huggingface.co/datasets/FreedomIntelligence/Medical-R1-Distill-Data} & Apache 2.0 \\
\rowcolor{StageThree} AlpaCare-MedInstruct-52k & \fiturl{https://huggingface.co/datasets/lavita/AlpaCare-MedInstruct-52k} & CC BY 4.0 \\
\rowcolor{StageThree} HealthCareMagic-100k & \fiturl{https://huggingface.co/datasets/lavita/ChatDoctor-HealthCareMagic-100k} & Apache 2.0 \\
\rowcolor{StageThree} icliniq10k & \fiturl{https://huggingface.co/datasets/zhengComing/iCliniq-10K} & Apache 2.0 \\
\rowcolor{StageThree} PMC-LLaMA & \fiturl{https://github.com/chaoyi-wu/PMC-LLaMA} & Apache 2.0 \\
\rowcolor{StageThree} HuatuoGPT2-GPT4-SFT-140K & \fiturl{https://huggingface.co/datasets/FreedomIntelligence/HuatuoGPT2-GPT4-SFT-140K} & Apache 2.0 \\
\rowcolor{StageThree} MedReason & \fiturl{https://huggingface.co/datasets/UCSC-VLAA/MedReason} & CC BY 4.0 \\
\rowcolor{StageThree} MedThoughts-8K & \fiturl{https://huggingface.co/datasets/hw-hwei/MedThoughts-8K} & MIT \\
\rowcolor{StageThree} MedQA & \fiturl{https://github.com/jind11/MedQA} & MIT \\
\rowcolor{StageThree} open-book-CMB-exam & \fiturl{https://huggingface.co/datasets/wangrongsheng/open-book-CMB-exam} & Apache 2.0 \\
\rowcolor{StageThree} cMedQA & \fiturl{https://github.com/zhangsheng93/cMedQA2} & GPL-3.0 \\
\rowcolor{StageThree} HealthCareMagic & \fiturl{https://huggingface.co/datasets/lavita/ChatDoctor-HealthCareMagic-100k} & Apache 2.0 \\
\rowcolor{StageThree} iCliniq-10K & \fiturl{https://huggingface.co/datasets/zhengComing/iCliniq-10K} & Apache 2.0 \\
\rowcolor{StageThree} Citrus\_S3 & \fiturl{https://huggingface.co/datasets/jdh-algo/Citrus_S3} & MIT \\
\rowcolor{StageThree} MedBooks-18-CoT & \fiturl{https://huggingface.co/datasets/dmis-lab/meerkat-instructions} & CC BY-NC 4.0 \\
\rowcolor{StageThree} Synthetic Case Report Long CoT & Synthetic Data & Internal Synthetic Data \\
\midrule
\rowcolor{StageThree}\multicolumn{3}{l}{\textbf{Application Stage: S3 -- General text}} \\
\rowcolor{StageThree} LLaVA-v1.5-text-subset & \fiturl{https://github.com/haotian-liu/LLaVA} & CC BY-NC 4.0 \\
\rowcolor{StageThree} OpenHermes-2.5 & \fiturl{https://huggingface.co/datasets/teknium/OpenHermes-2.5} & Apache 2.0 \\
\midrule
\rowcolor{StageFour}\multicolumn{3}{l}{\textbf{Application Stage: S4 -- Medical multimodal}} \\
\rowcolor{StageFour} CT-Rate-report & \fiturl{https://huggingface.co/datasets/ibrahimhamamci/CT-RATE} & CC BY-NC-SA 4.0 \\
\rowcolor{StageFour} Merlin-report & \fiturl{https://github.com/StanfordMIMI/Merlin} & MIT \\
\rowcolor{StageFour} RadGenome-ChestCT & \fiturl{https://huggingface.co/datasets/RadGenome/RadGenome-ChestCT} & CC BY 4.0 \\
\rowcolor{StageFour} M3D-CAP & \fiturl{https://huggingface.co/datasets/GoodBaiBai88/M3D-Cap} & Apache 2.0 \\
\rowcolor{StageFour} OLD-INSPECT & \href{https://aimi.stanford.edu/datasets/inspect-Multimodal-Dataset-for-Pulmonary-Embolism-Diagnosis-and-Prognosis}{\nolinkurl{https://aimi.stanford.edu/datasets/inspect-Multimodal-Dataset...}} & Credentialed Access \\
\rowcolor{StageFour} AMOS-MM-report & \fiturl{https://zenodo.org/doi/10.5281/zenodo.10992154} & CC BY 4.0 \\
\rowcolor{StageFour} AbdomenCT & Synthetic Data & Internal Synthetic Data \\
\rowcolor{StageFour} ChronicCT & Synthetic Data & Internal Synthetic Data \\
\rowcolor{StageFour} CardiovascularCT & Synthetic Data & Internal Synthetic Data \\
\rowcolor{StageFour} PulmonaryNodulesCT & Synthetic Data & Internal Synthetic Data \\
\midrule
\rowcolor{StageFive}\multicolumn{3}{l}{\textbf{Application Stage: S5 -- Medical multimodal}} \\
\rowcolor{StageFive} All S3 Medical multimodal datasets &  & Same access terms as S3 Medical multimodal \\
\rowcolor{StageFive} AMOS-MM-VQA & \fiturl{https://zenodo.org/doi/10.5281/zenodo.10992154} & CC BY 4.0 \\
\rowcolor{StageFive} CT-Rate-VQA & \fiturl{https://huggingface.co/datasets/ibrahimhamamci/CT-RATE} & CC BY-NC-SA 4.0 \\
\rowcolor{StageFive} 3D-RAD & \fiturl{https://github.com/Tang-xiaoxiao/3D-RAD} & MIT \\
\rowcolor{StageFive} Synthetic-M3d-VQA & Synthetic Data & Internal Synthetic Data \\
\rowcolor{StageFive} Merlin-MCQ & \fiturl{https://github.com/StanfordMIMI/Merlin} & MIT \\
\rowcolor{StageFive} Merlin-VQA & \fiturl{https://github.com/StanfordMIMI/Merlin} & MIT \\
\midrule
\rowcolor{StageFive}\multicolumn{3}{l}{\textbf{Application Stage: S5 -- General multimodal}} \\
\rowcolor{StageFive} Same as S3 General multimodal datasets &  & Same access terms as S3 General multimodal \\
\midrule
\rowcolor{StageFive}\multicolumn{3}{l}{\textbf{Application Stage: S5 -- Medical text}} \\
\rowcolor{StageFive} Same as S3 Medical text datasets &  & Same access terms as S3 Medical text \\
\midrule
\rowcolor{StageFive}\multicolumn{3}{l}{\textbf{Application Stage: S5 -- General text}} \\
\rowcolor{StageFive} Same as S3 General text datasets &  & Same access terms as S3 General text \\
\bottomrule
\end{tabular}}}
\end{table*}

%% file: content/06_Appendix.tex
\renewcommand{\figurename}{Supplementary Fig.}
\setcounter{figure}{0}
\renewcommand{\thefigure}{\arabic{figure}}
\renewcommand{\tablename}{Supplementary Table}
\setcounter{table}{0}
\renewcommand{\thetable}{\arabic{table}}

\setcounter{section}{0}
\renewcommand{\thesection}{S\arabic{section}}

\section*{Supplementary Information}



\section{Supplementary Methods: Data Synthesis Pipelines}

\begin{figure}[!htbp]
\centering
\includegraphics[width=1.0\linewidth]{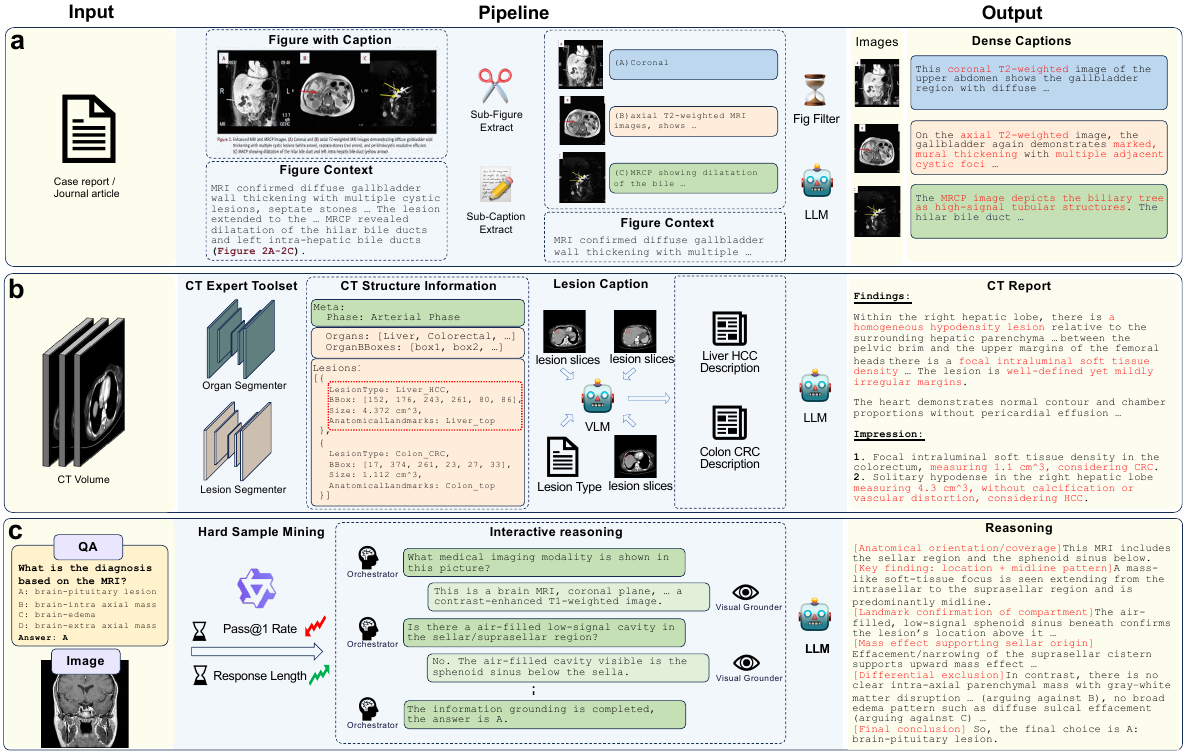}
\caption{
\textbf{Data synthesis pipeline of \method}.
\textbf{a}, Dense caption generation.
We parse case reports and journal articles to obtain figure images together with their original captions and surrounding context, segment composite figures into sub-panels, and extract panel-wise subcaptions aligned to each sub-figure. After multi-stage filtering to remove non-clinical, heavily annotated, or low-quality regions, we regenerate dense, visually grounded descriptions by conditioning on both the sub-image and the original (sub)caption/context.
\textbf{b}, CT annotation with visual expert toolset.
For CT volumes without paired reports, we employ a CT Visual Expert to produce a structured CT representation that includes organ context and lesion-level attributes (\textit{e.g.}, type, coarse 3D location, and extent proxies). Using the lesion entries in this representation, we retrieve representative CT slices that cover each lesion and overlay lesion-region markers on the selected slices. The marked slices, together with the corresponding lesion metadata, are then provided to a VLM to generate lesion-focused captions. Finally, these lesion captions and global structured attributes are consolidated by an LLM into standardized radiology-style Findings/Impression pseudo-reports.
\textbf{c}, Interactive multimodal CoT annotation for reasoning-hard samples.
We first score candidate multimodal QA instances with Qwen3-VL and prioritize samples that elicit long analytical responses yet exhibit low pass@1, indicating failures in long-horizon grounded reasoning rather than superficial knowledge gaps. The selected hard instances are then solved with a carefully designed interactive reasoning protocol that separates a text-only orchestrator from a multimodal visual grounder. Given the question, the orchestrator decomposes it into a sequence of visual fact-check queries, iteratively interrogates the visual grounder to obtain image-grounded evidence, and terminates once the accumulated evidence is sufficient to support a final answer. After completion, an additional LLM pass rewrites the full dialogue trajectory into a coherent reasoning trace for instruction tuning.
}
\label{fig:data_pipelines}
\end{figure}

This section provides the detailed procedures for the four data synthesis pipelines summarized in the Methods (Section~\ref{sec:data_curation}).
These pipelines target distinct training bottlenecks:
\textbf{1)} \emph{instruction warping} diversifies medical instructions to mitigate the loss of instruction-following ability during medical adaptation;
\textbf{2)} \emph{dense caption generation} converts noisy paper-figure caption pairs into grounded, fine-grained vision--language supervision;
\textbf{3)} \emph{interactive multimodal CoT annotation} synthesizes long-horizon, visually grounded reasoning trajectories for hard multimodal QA; and
\textbf{4)} \emph{large-scale CT caption with expert toolset} produces 1.1M radiology-style pseudo-reports for hospital-collected CT volumes that lack reports.
The complete workflows for Dense Caption Generation, CT Caption with Expert Toolset, and Interactive Multimodal CoT Annotation are illustrated in~\suppfigref{fig:data_pipelines}.

\subsection{Instruction Warping}
\label{sec:instruction_warping}

Medical adaptation tends to erode general instruction-following ability, which in turn constrains how the underlying medical knowledge can be expressed in realistic clinical settings, such as structured outputs (JSON), scope restriction (``only describe the RoI''), negative constraints (``ignore lungs''), or domain-specific formatting required by clinical workflows.
To address this, we build an instruction augmentation pipeline via \textbf{instruction warping}, inspired by self-instruction, to diversify instructions along multiple axes while preserving medical semantics.

\noindent\textbf{Seed templates.}
We manually author 120 seed instructions spanning diverse clinical application scenarios and task types, each with explicit formatting requirements (\textit{e.g.}, strict headers, JSON schemas, table columns, and a fixed Final Answer anchor).
The seeds are designed to cover four axes:
\textbf{1)} dialogue roles (\textit{e.g.}, radiologist note, patient-facing explanation, triage assistant);
\textbf{2)} scenarios (\textit{e.g.}, outpatient consult, ED triage, follow-up comparison, multidisciplinary discussion);
\textbf{3)} format-following (JSON/tables/bullets/checklists with length and field constraints); and
\textbf{4)} constraints (scope-only, negative constraints, uncertainty handling, citation requirements).

\noindent\textbf{LLM-based expansion.}
Building on these seeds, we use GPT-5.1 to automatically expand them into 21k instruction templates.
For each data sample, we parse it into structured schema fields and then recompose it using a sampled template, which substantially increases instruction diversity while introducing minimal semantic drift.

\subsection{Dense Caption Generation}
\label{sec:dense_caption_mining}

A non-trivial portion of large-scale medical multimodal data is collected from paper figures paired with captions.
However, these pairs often suffer from low alignment quality: captions typically state only the most critical conclusion and thus provide weak, non-dense supervision; meanwhile, paper figures are frequently composite, mixing medical images with plots, arrows, symbols, and annotations.
Such visual clutter introduces noise and also deviates from real clinical imagery, reducing transferability to practical medical settings (\suppfigref{fig:data_pipelines}a).

\noindent\textbf{Sub-figure segmentation and filtering.}
Following~\cite{lozano2025biomedicaopenbiomedicalimagecaption}, we perform sub-figure segmentation on document screenshots to decompose complex figures into candidate sub-images.
We then apply a multi-stage filtering pipeline to retain only high-quality medical imagery:
\textbf{1)} remove non-image panels (plots/tables);
\textbf{2)} remove heavily annotated or low-resolution regions;
\textbf{3)} validate modality and medicalness; and
\textbf{4)} deduplicate near-duplicates.

\noindent\textbf{Dense caption regeneration.}
For retained sub-images, we generate dense descriptions by conditioning on the original caption and the visual content, moving from ``headline'' captions to grounded, fine-grained supervision that inventories visible findings, anatomical context, and modality-specific patterns, while avoiding non-visual speculation.
We employ a two-pass generation scheme:
\textbf{1) Visual inventory.} We use two complementary VLMs, GPT-5.1 and Gemini 3 Pro, to independently produce a visual inventory that enumerates directly observable elements (\textit{e.g.}, findings, locations, and appearance cues). We then retain only the consistent, cross-model overlapping items as a reliability filter.
\textbf{2) Clinically coherent description.} The consensus inventory is then used to generate a clinically coherent description with appropriate medical phrasing and structure.

We curate our raw corpus from 100k journal articles and 785k case reports, yielding 749k high-quality medical images paired with captions.
Applying the above dense-captioning pipeline produces 610k dense-caption samples, summarized as Synthetic Case Report Caption and Synthetic Journal Caption in~\exttabref{ext-tab:data_sources_v260}.
These dense captions are primarily used in early alignment stages to strengthen vision--language grounding, and also serve as rich supervision for later instruction tuning.

\subsection{Interactive Multimodal CoT Annotation}
\label{sec:multi_agent_reasoning}

Beyond perception, medical MLLMs must sustain multi-step reasoning over multimodal evidence.
We find that many existing medical VQA instances are either too short or can be solved by shallow pattern matching, and thus do not strongly incentivize long-horizon multimodal reasoning.
To explicitly train this capability, we construct a hard multimodal reasoning dataset via a multi-agent annotation pipeline, following the core idea of iterative look-again~\cite{jian2025look} reasoning (\suppfigref{fig:data_pipelines}c).

\noindent\textbf{Difficulty mining.}
We assign each candidate multimodal QA instance a difficulty score using a fixed solver model.
For each instance, we query the solver under a standardized prompt for $n$ stochastic samples, and compute:
\textbf{1)} the empirical correctness rate (pass@1); and
\textbf{2)} the average generation length measured in output tokens.
In practice, we use Qwen3-VL-8B~\cite{qwen3-vl} as the solver model.

Our intuition is that instances with low correctness yet high generation length often induce extensive but unproductive reasoning, which are desirable seeds for constructing reasoning-focused training data.
Accordingly, we prioritize instances by:
\begin{equation}
S^{\mathrm{hard}}(\text{sample}) = \bigl(1 - \mathrm{pass@1}(\text{sample})\bigr) \cdot \log\bigl(1 + \overline{Len}(\text{sample})\bigr),
\end{equation}
where $\mathrm{pass@1}(\text{sample})$ denotes the solver's single-sample accuracy and $\overline{Len}(\text{sample})$ is the average response length.
We label instances with $S^{\mathrm{hard}} > 4$ as reasoning-hard samples and further apply diversity-aware sampling to avoid over-representing redundant patterns, yielding 100k diverse reasoning-hard instances.

\noindent\textbf{Multi-agent reasoning.}
We design a multi-agent pipeline that couples a text-only orchestrator with a multimodal visual grounder through iterative, evidence-seeking interaction.
For each instance, the orchestrator has access only to the textual components (question and candidate options) and cannot directly observe the image/volume.
It first analyzes the question to identify the visual attributes, anatomical cues, or regions that must be verified, then converts them into targeted fact-checking sub-questions.
The visual grounder receives the medical image/volume along with the orchestrator's queries, inspects the visual evidence, and returns grounded observations.
We use GPT-5.1 as the orchestrator and Gemini 3 Pro as the visual grounder.

The orchestrator integrates the returned evidence, updates its hypotheses, and either requests additional, more specific visual checks or finalizes the answer once the accumulated evidence is sufficient.
This closed-loop interaction proceeds for multiple rounds and enforces that each reasoning step is supported by newly retrieved visual evidence, thereby mitigating grounding drift as reasoning depth grows.
We retain only trajectories whose final conclusion matches the reference answer, resulting in 67k high-confidence hard multimodal reasoning instances that are primarily used for instruction tuning to improve long-horizon, visually grounded reasoning fidelity.

\subsection{Large-Scale CT Caption with Expert Toolset}
\label{sec:ct_report_alignment}

CT is a cornerstone modality for diagnosing and differentiating many diseases, and natively understanding 3D CT volumes is central to real clinical deployment.
We therefore construct CT--report aligned supervision at scale through a two-pronged strategy:
\textbf{1)} collecting open-source CT datasets that already include high-quality reports, and
\textbf{2)} expanding coverage with hospital-collected CT scans that lack reports by generating report-style pseudo-labels (\suppfigref{fig:data_pipelines}b).

\noindent\textbf{Open-source CT with high-quality reports.}
We curate 0.2M CT studies from open-source datasets with high-quality radiology reports~\cite{bai2024m3d,hamamci2024CT-Rate,blankemeier2024merlin,huang2023inspect}.
These provide clean paired supervision for 3D alignment and report generation evaluation.

\noindent\textbf{Hospital CT collection and pseudo-report generation.}
To further advance CT-centric modeling, we collaborate with multiple hospitals to curate a large-scale internal dataset comprising 1.5M CT scans.
Due to data privacy agreements, we were only permitted to access the de-identified CT images, without paired radiology reports.
To enable report-level supervision at scale, we developed an annotation-guided CT report annotation pipeline based on a visual-expert toolset, which primarily consists of a lesion segmentation model~\cite{he2025vista3d} and an organ segmentation model~\cite{wasserthal2023totalsegmentator}.
This toolset produces structured pseudo-annotations from the images, which are subsequently transformed into radiology-style captions.

The pipeline proceeds in four steps.
\textbf{1)} The lesion segmentation model produces lesion masks with anatomical labels, which are aggregated into a structured CT representation capturing clinically relevant attributes (\textit{e.g.}, organ presence, lesion location, and lesion patterns).
\textbf{2)} For each lesion entry, we retrieve representative CT slices that cover the lesion and overlay lesion-region markers on the selected slices.
\textbf{3)} The marked slices, together with lesion metadata, are provided to a VLM to generate lesion-focused local captions.
\textbf{4)} We then synthesize the structured attributes and lesion captions into standardized radiology-style Findings/Impression text via medically constrained generation.

Rigorous quality control is applied throughout, including mask sanity checks (size bounds and topology constraints), cross-slice consistency verification, and text-level validation to reduce spurious outputs and avoid unsupported claims.
Using this pipeline, we generate pseudo-report supervision for 1.1M CT cases, substantially expanding the scale and diversity of CT--report aligned data.

The full CT--report aligned corpus (0.2M open-source + 1.1M pseudo-labeled) supports both Stage \ding{175} (rapid 3D encoder alignment) and Stage \ding{176} (holistic 2D/3D instruction tuning).
The large pseudo-labeled portion improves volumetric diversity and robustness, while the smaller open-source portion provides high-fidelity alignment anchors.

\section{Supplementary Methods: Agentic Tool Use}

This section provides the detailed retrieval algorithms for the knowledge tool (Section~\ref{sec:rag_tool}) and the tool routing and interface design for the perception expert tools (Section~\ref{sec:perception_tools}) summarized in the Methods.

\subsection{Knowledge Tool: Hybrid Retrieval and Cross-Source Aggregation}

The knowledge tool combines two complementary retrieval channels followed by cross-source evidence aggregation.

\noindent\textbf{Hybrid retrieval with concept normalization}. 
Medical queries and documents contain pervasive ambiguity (synonyms, abbreviations, brand–generic drug names, variant spellings). 
To reduce such ambiguity at the retrieval stage, we perform concept normalization using UMLS before querying: medical entities mentioned in the query are mapped to standardized medical concepts. 
Retrieval then proceeds through two complementary channels. 
First, a vector search engine retrieves semantically matched evidence passages from large-scale corpora, improving recall under linguistic variation. 
Second, KG-based recall leverages two open-source medical knowledge graphs: UMLS~\cite{UMLS2024AA}, which provides standardized medical concept identifiers for robust normalization across synonyms and abbreviations, and PrimeKG~\cite{Chandak2022PrimeKG}, which offers rich biomedical relations (\textit{e.g.}, drug--disease, symptom--disease, contraindication) that support link-based expansion to clinically relevant upstream/downstream concepts.
This structured recall enables concept-to-evidence routing and reduces misses caused by surface-form mismatch.
The two channels are integrated by normalized concepts and semantic similarity, producing a candidate evidence set grounded in explicit medical entities and relations.

\noindent\textbf{Cross-source evidence aggregation}. 
Medical QA often benefits from combining complementary evidence views: vector retrieval typically returns highly query-aligned passages, while KG-based recall provides richer upstream/downstream context anchored to normalized entities. 
Since these two channels differ in granularity and surface form, we aggregate evidence by clustering retrieved chunks across both channels using semantic similarity and overlap of normalized medical entities, grouping content that states the same clinical claim or offers mutual support from different sources. 
\method then summarizes each cluster to preserve the shared clinical conclusion together with key qualifiers (\textit{e.g.}, indication, contraindication, dose constraints) and attaches citations to the supporting chunks. 
This cluster-then-summarize aggregation jointly leverages the precision of embedding-based retrieval and the relational breadth of KG recall, producing a compact, evidence-grounded context for faithful, traceable clinical answers.

\subsection{Perception Expert Tools: Tool Descriptions and Routing}

\noindent\textbf{Detailed tool descriptions.}
We provide expanded descriptions of the five perception experts integrated into \method, each producing structured findings that the agent can directly use for evidence-based reasoning.
\begin{itemize}
    \item \textbf{Multi-organ lesion segmentation expert}~\cite{he2025vista3d} segments 29 common lesion types across eight organs: breast, colon, esophagus, stomach, kidney, lung, liver, and pancreas.
    It enables \method to ground answers in explicit lesion inventories (how many, where, how large), support tumor burden quantification, and maintain consistency across multi-lesion or multi-organ cases where free-form visual description is error-prone.

    \item \textbf{Fatty liver assessment expert}~\cite{gao2026multi} estimates fatty liver severity from CT and localizes the liver parenchyma region used for assessment.
    It enables \method to provide standardized steatosis assessment with traceable imaging evidence (parenchyma region), improving reproducibility for screening, risk stratification, and longitudinal monitoring.

    \item \textbf{Abdominal lymph node segmentation expert}~\cite{yu2025effective} segments abdominal lymph nodes and is sensitive to node size, supporting the detection of lymphadenopathy.
    It enables \method to perform structured nodal evaluation (node enumeration + size-based criteria), which is central to oncologic staging and response tracking and is difficult to do reliably via generic vision-language perception.

    \item \textbf{Chest X-ray finding classification expert}~\cite{cohen2020limits,yang2025chest} predicts structured probabilities for a broad set of thoracic findings from chest radiographs. 
    In our system, this expert combines TorchXRayVision, which provides standardized preprocessing and strong pre-trained chest X-ray models for 18 common findings, with CheXFound, a chest X-ray foundation model pretrained on up to one million images and supporting classification over 40 findings. 
    This expert enables \method to supplement free-form image understanding with explicit pathology-level evidence, improving robustness and consistency in report generation and medical VQA.

    \item \textbf{Chest X-ray finding generation expert}~\cite{chexagent-2024} generates clinically coherent chest X-ray findings descriptions from radiographs. 
    Built on CheXagent, a vision-language foundation model trained on the large-scale CheXinstruct dataset and evaluated across diverse chest X-ray interpretation tasks, this expert is mainly used to draft findings-style descriptions that summarize the visible thoracic abnormalities and their imaging context. 
    It enables \method to obtain richer sentence-level imaging evidence when classification scores alone are insufficient, improving the completeness and fluency of chest X-ray report drafting and finding-oriented question answering.
\end{itemize}

\noindent\textbf{Tool routing and schema-first interface.}
\method does not follow a rigid workflow; instead it seeks the specific visual evidence required by the user’s question and calls tools accordingly. 
When the query implies that the answer depends on precise localization or measurement—--such as requesting the largest lesion size, asking whether lymph nodes are enlarged, asking for fatty liver grade, or asking for lesion counts in a target organ—\method treats this as an evidence gap and initiates tool calls rather than relying on approximate visual judgment. 
To make tool use stable, controllable, and consumable by \method, we adopt a schema-first interface. 
Each perception expert is registered with a capability card that specifies the supported anatomies and label set, the measurable endpoints such as diameters, volumes, and severity grades, and the types of clinical questions the tool can resolve. 
This explicit specification enables intentional routing. 
Tool outputs are normalized into compact evidence objects and then verbalized into natural-language evidence statements before being fed back to \method. 
Overly granular numeric fields such as voxel-level coordinates are mapped to coarse standardized location descriptors, \textit{e.g.}, \texttt{liver\_top}, which reduces sensitivity to coordinate noise while preserving clinically actionable localization.


\section{Supplementary Ablation Studies}



\noindent\textbf{Stochastic residual regularization shows benefits to 3D volumetric understanding.}
Based on the above experiments, we add the 3D ViT as the fourth vision encoder to enable native volumetric understanding.
We train the model with all five stages with partial 2D and 3D data to ensure fast feedback.
Our prior experiments indicated an effect of diminishing returns when adding a third vision encoder, suggesting that later-stage components may struggle to integrate effectively into an already powerful model. 
We hypothesized that a stronger regularization scheme could mitigate this side effect by encouraging more robust feature learning.
To test this, we evaluated the impact of applying stochastic residual regularization during training.
The results, presented in~\supptabref{tab:ablation_stochastic_3d}, provide evidence for our hypothesis. 
Employing the stochastic approach leads to a clear improvement in both the average 3D VQA score (from 64.8 to 65.8) and the average report generation metrics (from 23.4 / 13.3 / 15.1 to 23.7 / 14.0 / 15.7). 
However, we also observed that these benefits are contingent on sufficient training duration.
In shorter training runs, the performance gap between the stochastic and deterministic models was negligible or even inverted. 
This suggests that while stochastic regularization provides a more robust learning signal, it requires adequate optimization steps to fully manifest its advantages.

\begin{figure}[!htbp]
\centering
\includegraphics[width=1.0\linewidth]{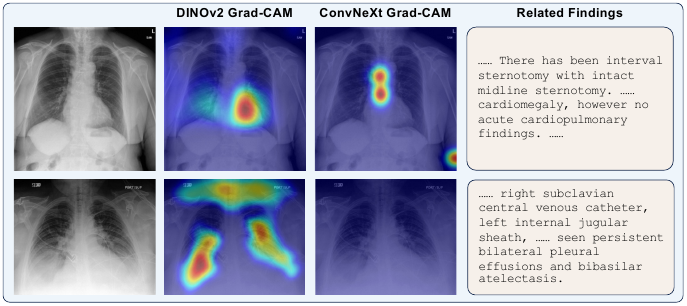}
\caption{
\textbf{Grad-CAM visualizations} revealing the complementary attentional focus of specialist encoders.
For each example, we show the original image, the Grad-CAM heatmaps for DINOv2 and ConvNext, and the corresponding related findings in the report.
}
\label{fig:grad-cam-case}
\end{figure}

\noindent\textbf{Interpreting the functionality of additional encoders.}
To qualitatively interpret how the compositional vision architecture contributes to the final prediction, we employ Grad-CAM~\cite{selvaraju2020grad_cam} to visualize the regions of focus for our specialist encoders, DINOv2 and ConvNext. 
We analyze two chest X-ray report generation examples to illustrate how these encoders exhibit a complementary division of labor, which is then unified by our CaSL Fusion mechanism to produce a comprehensive and accurate report.
As shown in~\suppfigref{fig:grad-cam-case}, the specialist encoders demonstrate distinct yet synergistic attentional patterns. 
In the first example (top row), which describes a post-operative chest, the Grad-CAM visualizations reveal a clear separation of concerns. 
DINOv2, known for its strong semantic feature extraction, directs its focus squarely on the cardiac silhouette. This global, semantic focus is critical for assessing overall organ size and shape, leading to the identification of cardiomegaly.
In comparison, ConvNext, with its powerful inductive bias for local patterns and textures, precisely highlights the midline sternotomy wires. 
This demonstrates its specialization in identifying fine-grained, man-made objects or textural abnormalities. 
The second example (bottom row) further reinforces this observation. 
The case involves multiple findings, including bilateral pleural effusions and atelectasis. 
Here, DINOv2 casts a wide attentional net, focusing on the bilateral pleural effusions and the lung bases, which are key for identifying the reported bibasilar atelectasis. 
Intriguingly, in this specific case, ConvNext's visualization does not show significant activation, suggesting that its role may be more nuanced or context-dependent. 
This might indicate that for cases dominated by large-scale fluid and density changes (like effusions), DINOv2's global semantic understanding dominates, while ConvNext's contribution becomes more critical when fine-grained textural details are diagnostically relevant. 
This dynamic interplay, managed by our cascaded fusion, allows \method to flexibly leverage the most relevant visual evidence for any useful case, leading to a more robust and holistic perception.



\clearpage
\section{Supplementary Figures}

\clearpage
\input{content/08_Tables}

%% file: content/08_Tables.tex
\section{Supplementary Tables}

\input{tables/2d_mm_bench}


\input{tables/2d_report_generation}


\input{tables/language_bench}

\input{tables/3d_mm_bench}

\input{tables/medif_bench}

\input{tables/ablation_report_generation}

\input{tables/ablation_fusion_mechanism}

\input{tables/ablation_one_additonal_encoder}
\input{tables/ablation_multi_additional_encoder}
\input{tables/ablation_3d_encoder}
\input{tables/ablation_stochastic_3d}

%% file: tables/2d_mm_bench.tex
\begin{table*}[!htbp]
    \setlength{\tabcolsep}{5pt}
    \centering
    \caption{\textbf{Results on 2D Multimodal Medical Benchmarks}. 
    We list results of leading proprietary models, leading generalist MLLMs, and other medical MLLMs.
    \textbf{Bold} results denote the best scores in medical MLLMs and \underline{underlined} results denote the second-best scores.
    $\heartsuit$ denotes that the model does not support multi-image input.
    $\dag$ indicates that results were re-benchmarked using open-source model weights or model APIs within our own evaluation framework. 
    }
    \label{tab:2d_mm_bench}
    \resizebox{1.\linewidth}{!}{
    \begin{tabular}{lcccccccc}
        \toprule
        & \multicolumn{4}{c}{\textbf{Multi-modality}} & \multicolumn{3}{c}{\textbf{Specific-modality}} & \textbf{Reasoning} \\
        & \multicolumn{4}{c}{\textbf{Benchmarks}} & \multicolumn{3}{c}{\textbf{Benchmarks}} & \textbf{Benchmark} \\
        \cmidrule(lr){2-5} \cmidrule(lr){6-8} \cmidrule(l){9-9}
        \textbf{Models} & OM.VQA & PMCVQA & MedF & GMAI & VQA-RAD & SLAKE & PathVQA & MedXQA \\
        \midrule
        \multicolumn{9}{c}{\textit{Proprietary Models}} \\
        \midrule
        GPT-5.2\textsuperscript{\dag} & 68.4 & 56.7 & 52.1 & 58.4 & 73.4 & 79.4 & 54.0 & 52.9 \\
        Gemini-3-Flash\textsuperscript{\dag} & 82.0 & 64.4 & 53.3 & 67.8 & 71.0 & 80.3 & 60.6 & 78.0 \\
        Claude-Sonnet-4.5\textsuperscript{\dag} & 67.1 & 58.4 & 54.2 & 47.6 & 63.6 & 72.6 & 51.7 & 42.8 \\
        \midrule
        \multicolumn{9}{c}{\textit{General-purpose MLLMs}} \\
        \midrule
        Qwen2.5-VL-7B\textsuperscript{\dag} & 63.3 & 51.7 & 46.9 & 44.0 & 59.6 & 67.4 & 43.4 & 22.5 \\
        Qwen3-VL-8B\textsuperscript{\dag} & 76.4&54.0 &42.1 &54.0 &61.9 &70.4 &42.8&24.9 \\
        InternVL3-8B & 79.1 & 53.8 & -- & -- & 65.4 & 72.8 & 48.6 & 22.4 \\
        InternVL3-14B & 78.9 & 54.1 & -- & -- & 66.3 & 72.8 & 48.0 & 23.1 \\
        Qwen2.5-VL-32B\textsuperscript{\dag} &67.2 &53.2 & 51.5 &47.4 &72.7 & 77.2&46.3 & 25.4 \\
        Qwen3-VL-32B\textsuperscript{\dag} & 76.8 & 56.5 & 52.0 & 56.3 & 66.1& 76.9& 48.6& 31.2\\
        InternVL3-38B & 79.8 & 56.6 & -- & -- & 65.4 & 72.7 & 51.0 & 25.2 \\
        Qwen3-VL-235B\textsuperscript{\dag} & 75.8 & 56.2 & 50.3 & 54.9  & 67.8 & 75.5 & 51.1 & 33.7 \\
        \midrule
        \multicolumn{9}{c}{\textit{Medical MLLMs}} \\
        \midrule
        \multicolumn{9}{l}{\textit{--- Models < 10B ---}} \\
        BiomedGPT\textsuperscript{$\heartsuit$} & 27.9 & 27.6 & -- & -- & 16.6 & 13.6 & 11.3 & -- \\
        Med-R1-2B & -- & 47.4 & -- & -- & 39.0 & 54.5 & 15.3 & 21.1 \\
        MedVLM-R1-2B & 77.6 & 48.8 & -- & -- & 49.2 & 56.3 & 36.0 & 21.4 \\
        HealthGPT-M3 & 71.5 & 55.4 & -- & -- & 56.8 & 70.8 & 55.4 & 22.4 \\
        BioMediX2-8B & 66.0 & 41.8 & -- & -- & 55.7 & 54.1 & 34.6 & 21.9 \\
        LLaVA-Med-7B & 34.8 & 22.7 & -- & -- & 46.6 & 51.9 & 35.2 & 20.8 \\
        MedGemma-4B-IT & 70.7 & 49.2 & -- & -- & 72.3 & 78.2 & 48.1 & 25.4 \\
        HuatuoGPT-V-7B & 74.3 & 53.1 & -- & -- & 67.6 & 68.1 & 44.8 & 23.2 \\
        Lingshu-7B & 82.9 & 56.3 & -- & -- & 67.9 & 83.1 & 61.9 & 26.7 \\
        Hulu-Med-7B & \underline{84.2} & \underline{66.8} & -- & -- & \textbf{78.0} & \underline{86.8} & \underline{65.6} & \underline{29.0} \\
        MedGemma-1.5-4B-IT\textsuperscript{\dag} & 47.5 & 18.7 & 40.5 & 21.8 & 55.2 & 63.6 & 34.7 & 18.5 \\
        Lingshu-7B\textsuperscript{\dag} & 83.4 & 57.4 & \underline{58.2} & \underline{51.8} & 64.3 & 81.5 & 57.3 & 26.4 \\
        Hulu-Med-7B\textsuperscript{\dag} & 79.4 & 61.7 & 54.8 & \underline{51.8} & 68.5 & 74.7 & 54.2 & 25.6 \\
        \textbf{\method-8B} & \textbf{89.6} & \textbf{67.8} & \textbf{64.3} & \textbf{57.9} & \underline{76.3} & \textbf{91.1} & \textbf{68.7} & \textbf{32.4} \\
        \midrule
        \multicolumn{9}{l}{\textit{--- Models > 10B ---}} \\
        HealthGPT-14B & 75.2 & 56.4 & -- & --  & 65.0 & 66.1 & 56.7 & 24.7 \\
        HuatuoGPT-V-34B & 74.0 & 56.6 & -- & --  & 61.4 & 69.5 & 44.4 & 22.1 \\
        Lingshu-32B & 83.4 & 57.9 & -- & --  & 76.7 & 86.7 & 65.5 & 30.9 \\
        MedDr-40B\textsuperscript{$\heartsuit$} & 64.3 & 13.9 & -- & -- & 65.2 & 66.4 & 53.5 & -- \\
        Hulu-Med-14B & \underline{85.1} & 68.9 & -- & -- & 76.1 & 86.5 & 64.4 & 30.0 \\
        Hulu-Med-32B & 84.6 & \underline{69.4} & -- & -- & \textbf{81.4} & 85.7 & \underline{67.3} & \underline{34.0} \\
        MedGemma-27B-IT\textsuperscript{\dag} & 60.0 & 49.2 & 49.5 & 37.3 & 65.6 & 75.5 & 43.4 & 23.6 \\
        Lingshu-32B\textsuperscript{\dag} & 83.4 & 62.3 & 58.9 & 55.6 & 73.2 & \underline{87.6} & 63.4 & 30.5 \\
        Hulu-Med-32B\textsuperscript{\dag} & 80.0 & 63.2 & \underline{61.6} & \underline{57.3} & 75.8 &  82.6 & 63.1& 32.6 \\
        \textbf{\method-32B} & \textbf{89.9} & \textbf{70.8} & \textbf{69.4} & \textbf{60.2} & \underline{77.2} & \textbf{92.5} & \textbf{72.4} & \textbf{38.7} \\
        \bottomrule
    \end{tabular}}
\end{table*}

%% file: tables/2d_report_generation.tex
\begin{table*}[!t]
    \centering
    \caption{\textbf{Results on 2D Report Generation Benchmarks}. 
    \textbf{Bold} results denote the best scores among medical MLLMs, and \underline{underlined} results denote the second-best scores.
    $\dag$ indicates that results were re-benchmarked using open-source model weights or model APIs within our own evaluation framework. 
    }
    \label{tab:2d_report_gen_bench}
    \resizebox{\linewidth}{!}{ 
    \begin{tabular}{lcccccc}
        \toprule
        & \multicolumn{3}{c}{\textbf{CheXpert-Plus}} & \multicolumn{3}{c}{\textbf{IU-XRAY}} \\
        \cmidrule(lr){2-4} \cmidrule(lr){5-7}
        \textbf{Models} & Precision & Recall & F1 & Precision & Recall & F1 \\
        \midrule
        \multicolumn{7}{c}{\textit{Proprietary Models}} \\
        \midrule
        GPT-5.2\textsuperscript{\dag} & 49.1 & 38.1 & 40.2 & 55.3 & 50.3 & 50.6 \\
        Gemini-3-Flash\textsuperscript{\dag} & 38.4 & 33.5 & 33.0 & 56.8 & 55.2 & 54.4 \\
        Claude-Sonnet-4.5\textsuperscript{\dag} & 33.8 & 28.3 & 28.9 & 39.2 & 39.2 & 37.7 \\
        \midrule
        \multicolumn{7}{c}{\textit{General-purpose MLLMs}} \\
        \midrule
        Qwen2.5-VL-7B\textsuperscript{\dag} & 13.7 & 11.5 & 12.0 & 33.0 & 32.9 & 32.4 \\
        Qwen3-VL-8B\textsuperscript{\dag} & 20.0 & 16.3 & 16.8 & 37.0 & 37.3& 36.7 \\
        Qwen2.5-VL-32B\textsuperscript{\dag} & 22.2 & 16.2 & 17.3 & 33.5 & 31.9 & 31.9 \\
        Qwen3-VL-32B\textsuperscript{\dag} & 30.6 & 27.9 & 27.3 & 40.8 & 41.0 & 40.0 \\
        Qwen3-VL-235B\textsuperscript{\dag} & 29.5 & 25.2  & 25.6 & 43.2  & 41.4  & 41.2  \\
        \midrule
        \multicolumn{7}{c}{\textit{Medical MLLMs}} \\
        \midrule
        \multicolumn{7}{l}{\textit{--- Models < 10B ---}} \\
        MedGemma-1.5-4B-IT\textsuperscript{\dag} & 34.1 & 23.1 & 25.4 & 43.1 & 38.9 & 39.8 \\
        Lingshu-7B\textsuperscript{\dag} & 26.2 & 17.8 & 19.9 & 43.7 & 39.3 & 40.4 \\
        Hulu-Med-7B\textsuperscript{\dag} & 40.5 & 29.1 & 31.9 & 51.1 & 45.3 & 46.5 \\
        \textbf{\method-8B} & \underline{46.4} & \underline{35.3} & \underline{37.8} & \underline{63.3} & \underline{55.8} & \underline{57.3} \\
        \textbf{\method-8B + Agentic tools} & \textbf{57.3} & \textbf{49.8} & \textbf{50.2} & \textbf{64.5} & \textbf{59.4} & \textbf{59.9} \\

        \midrule
        \multicolumn{7}{l}{\textit{--- Models > 10B ---}} \\
        MedGemma-27B-IT\textsuperscript{\dag} & 26.9 & 21.9 & 22.5 & 41.3 & 40.7 & 40.1 \\
        Lingshu-32B\textsuperscript{\dag} & 28.0 & 21.5 & 22.8 & 44.5 & 40.5 & 41.3 \\
        Hulu-Med-32B\textsuperscript{\dag} & 51.0 & 35.5 & 39.6 & 53.5 & 46.8 & 48.2 \\
        \textbf{\method-32B} & \underline{56.1} & \underline{46.6} & \underline{48.0} & \underline{60.2} & \underline{54.7} & \underline{55.6} \\ 
        \textbf{\method-32B + Agentic tools} & \textbf{60.2} & \textbf{55.4} & \textbf{54.8} & \textbf{60.8} & \textbf{56.1} & \textbf{56.3} \\ 
        
        \bottomrule
    \end{tabular}
    } 
\end{table*}

%% file: tables/language_bench.tex
\begin{table*}[!t]
    \setlength{\tabcolsep}{5pt}
    \centering
    \caption{
    \textbf{Results on Textual Medical Benchmarks}.
    We list results of leading proprietary models, leading generalist MLLMs, and other medical MLLMs.
    \textbf{Bold} results denote the best scores in medical MLLMs and \underline{underlined} results denote the second-best scores.
    $\dag$ indicates that results were re-benchmarked using open-source model weights or model APIs within our own evaluation framework. 
    }
    \label{tab:text_benchmarks}
    \resizebox{1.\linewidth}{!}{
    \begin{tabular}{l ccc c cccc}
        \toprule
        & \multicolumn{3}{c}{\textbf{Complex Reasoning Benchmarks}} & \textbf{Text Understanding} & \multicolumn{4}{c}{\textbf{Medical Exam Benchmarks}} \\
        & & & & \textbf{Benchmark} & & & \\
        \cmidrule(lr){2-4} \cmidrule(lr){5-5} \cmidrule(lr){6-9}
        \textbf{Models} & MedXQA & Medbullets & SGPQA & PubMedQA & MedMCQA & MedQA-U & MedQA-M  & MMLU-Med \\
        \midrule
        \multicolumn{9}{c}{\textit{Proprietary Models}} \\
        \midrule
        GPT-5.2 & 40.4 & 81.3 & 55.6 & 77.6 & 79.1 & 91.0 & 88.9 & 90.5 \\
        Gemini-3-Flash & 65.3 & 85.6 & 73.9 & 80.4 & 85.5 & 95.1 & 94.0 & 91.7 \\
        Claude-Sonnet-4.5 & 36.9 & 81.8 & 59.4 & 74.8 & 78.8 & 90.4 & 90.6 & 92.1 \\
        \midrule
        \multicolumn{9}{c}{\textit{General-purpose MLLMs}} \\
        \midrule
        Qwen2.5-VL-7B\textsuperscript{\dag} & 12.0 & 43.0 & 25.8 & 74.6 & 53.2 & 56.4 & 76.9 & 74.1 \\
        Qwen3-VL-8B\textsuperscript{\dag} & 14.6 & 51.9 & 36.9 & 72.2 & 60.7 & 66.2 & 87.2 & 80.0 \\
        InternVL3-8B & 13.1 & 48.5 & 31.2 & 75.4 & 57.7 & 62.1 & -- & 77.5 \\
        InternVL3-14B & 14.1 & 49.5 & 37.9 & 77.2 & 62.0 & 70.1 & -- & 81.7 \\
        Qwen2.5-VL-32B\textsuperscript{\dag} & 15.8 & 54.2 & 39.3 & 71.6 & 62.9 & 70.8 & 90.0 & 83.3 \\
        Qwen3-VL-32B\textsuperscript{\dag} & 18.2 & 60.1 & 47.4 & 71.0 & 68.2 & 77.5 & 91.7 & 86.6 \\
        InternVL3-38B & 16.0 & 54.6 & 42.5 & 73.2 & 64.9 & 73.5 & -- & 83.8 \\
        Qwen3-VL-235B\textsuperscript{\dag} & 23.3 & 69.6 & 53.8 & 75.2 & 74.3 & 85.2 & 93.7 & 88.8 \\
        \midrule
        \multicolumn{9}{c}{\textit{Medical MLLMs}} \\
        \midrule
        \multicolumn{9}{l}{\textit{--- Models < 10B ---}} \\
        MedVLM-R1-2B & 11.8 & 33.8 & 19.1 & 66.4 & 39.7 & 42.3 & -- & 51.8 \\
        BioMediX2-8B & 13.4 & 45.9 & 25.2 & 75.2 & 52.9 & 58.9 & -- & 68.6 \\
        MedGemma-4B-IT & 12.8 & 45.6 & 21.6 & 72.2 & 52.2 & 56.2 & -- & 66.7 \\
        HealthGPT-M3 & 11.5 & 41.4 & 18.9 & 57.8 & 54.2 & 55.0 & -- & 72.5 \\
        LLaVA-Med-7B & 9.90 & 34.4 & 16.1 & 26.4 & 39.4 & 42.0 & -- & 50.6 \\
        HuatuoGPT-V-7B & 10.1 & 40.9 & 21.9 & 72.8 & 51.2 & 52.9 & -- & 69.3 \\
        Lingshu-7B & 16.5 & 56.2 & 26.3 & 76.6 & 55.9 & 63.3 & -- & 74.5 \\
        Hulu-Med-7B & 19.6 & 61.5 & 31.1 & 77.4 & \underline{67.6} & 73.5 & -- & 79.5 \\
        MedGemma-1.5-4B-IT\textsuperscript{\dag} & 9.18 & 33.4 & 2.25 & 6.26 & 6.29 & 35.7 & 4.23 & 10.3 \\
        Lingshu-7B\textsuperscript{\dag} & 17.0 & 56.2 & 28.9 & 75.0 & 55.5 & 62.5 & 78.4 & 75.6 \\
        Hulu-Med-7B\textsuperscript{\dag} & \underline{20.0} & \underline{64.3} & \underline{32.6} & 75.2 & 64.2 & \underline{76.0} & 81.2 & 77.9 \\
        \textbf{\method-8B} & \underline{20.0} & \underline{64.3} & \underline{32.6} & \underline{77.6} & 62.8 & 72.6 & \textbf{89.1} & \underline{81.3} \\
        \textbf{\method-8B + Agentic tools} & \textbf{27.7} & \textbf{69.5} & \textbf{45.4} & \textbf{79.1} & \textbf{70.1} & \textbf{83.7} & \underline{84.3} & \textbf{86.0} \\
        \midrule
        \multicolumn{9}{l}{\textit{--- Models > 10B ---}} \\
        HealthGPT-14B & 11.3 & 39.8 & 25.7 & 68.0 & 63.4 & 66.2 & -- & 80.2 \\
        Lingshu-32B & 22.7 & 65.4 & 41.1 & 77.8 & 66.1 & 74.7 & -- & 84.7 \\
        HuatuoGPT-V-34B & 11.4 & 42.7 & 26.5 & 72.2 & 54.7 & 58.8 & -- & 74.7 \\
        MedDr-40B & 12.0 & 44.3 & 24.0 & 77.4 & 38.4 & 59.2 & -- & 65.2 \\
        Hulu-Med-14B & 23.2 & 68.5 & 37.7 & \underline{79.8} & 70.4 & 78.1 & -- & 83.3 \\
        Hulu-Med-32B & 24.2 & 68.8 & \underline{41.8} & \textbf{80.8} & \underline{72.8} & 80.4 & -- & 85.6 \\
        MedGemma-27B-IT\textsuperscript{\dag} & 15.8 & 56.8 & 31.4 & 75.4 & 64.7 & 71.4 & 73.1 & 80.7 \\
        Lingshu-32B\textsuperscript{\dag} & 22.2 & 65.3 & 41.4 & 77.4 & 65.6 & 73.4 & 89.6 & 84.5 \\
        Hulu-Med-32B\textsuperscript{\dag} & 19.8 & 67.5 & 38.0 & 78.8 & 68.5 & 77.1 & 87.0 & 83.7 \\
        \textbf{\method-32B} & \underline{26.7} & \underline{74.8} & \underline{41.8} & 78.8 & 70.7 & \underline{82.4} & \underline{93.8} & \underline{87.9} \\
        \textbf{\method-32B + Agentic tools} &  \textbf{31.2} & \textbf{77.9} & \textbf{50.9} & 79.7 & \textbf{74.0} & \textbf{86.6} & \textbf{94.5} & \textbf{89.9}\\
        \bottomrule
    \end{tabular}}
\end{table*}


%% file: tables/3d_mm_bench.tex
\begin{table*}[!t]
    \centering
    \caption{\textbf{Results on 3D Multimodal Medical Benchmarks}. 
    \textbf{Bold} results denote the best scores among medical MLLMs, and \underline{underlined} results denote the second-best scores.
    $\dag$ indicates that results were re-benchmarked using open-source model weights or model APIs within our own evaluation framework. 
    }
    \label{tab:3d_mm_bench}
    \resizebox{\linewidth}{!}{
    \begin{tabular}{l c cc cc ccc ccc}
        \toprule
        & \textbf{AMOS} & \multicolumn{2}{c}{\textbf{3D-RAD}} & \multicolumn{2}{c}{\textbf{CT-Rate}} & \multicolumn{3}{c}{\textbf{AMOS Report}} & \multicolumn{3}{c}{\textbf{CT-Rate Report}} \\
        \cmidrule(lr){2-2} \cmidrule(lr){3-4} \cmidrule(lr){5-6} \cmidrule(lr){7-9} \cmidrule(lr){10-12}
        \textbf{Models} & MCQ & MCQ & Open & MCQ & Open & Precision & Recall & F1 & Precision & Recall & F1 \\
        \midrule
        \multicolumn{12}{c}{\textit{Proprietary Models}} \\
        \midrule
        GPT-5.2 & 63.8 & 67.1 & 37.6 & 66.9 & 59.1 & 29.9 & 11.5 & 14.5 & 21.0 & 12.7 & 14.1 \\
        Gemini-3-Flash & 64.2 & 58.8 & 30.3 & 70.9 & 51.9 & 40.7 & 19.1 & 23.4 &  29.3 & 17.9 & 20.2 \\
        Claude-Sonnet-4.5 & 62.6 & 58.5 & 30.0 & 73.2 & 51.7 & 12.8 & 6.49 & 7.35 & 15.5 &  11.8& 12.3 \\
        \midrule
        \multicolumn{12}{c}{\textit{General-purpose MLLMs}} \\
        \midrule
        Qwen2.5-VL-7B\textsuperscript{\dag} & 53.4 & 40.8 & 28.6 & 67.3 & 44.5 & 6.31 & 4.18 & 4.69 & 10.6 & 9.42 & 9.61  \\
        Qwen3-VL-8B\textsuperscript{\dag} & 50.2 & 52.7 & 28.5 & 59.6 & 44.7 & 11.6 & 6.51 & 7.30 & 22.6 & 16.3 & 17.3 \\
        Qwen2.5-VL-32B\textsuperscript{\dag} &  55.8 & 46.4 &  24.0& 64.0 & 53.1 & 6.55 & 4.09 & 4.57 & 16.9 & 13.7 & 14.2 \\
        Qwen3-VL-32B\textsuperscript{\dag} & 58.9 & 64.7 & 30.1 & 67.1 & 53.6 & 12.6 & 7.61 & 8.37 & 30.6 & 18.2 & 20.5 \\
        Qwen3-VL-235B\textsuperscript{\dag} & 57.1 & 47.7 & 28.9 & 62.3 & 50.9 & 14.6 & 7.25 & 8.53 & 21.6 & 16.1 & 16.9 \\
        \midrule
        \multicolumn{12}{c}{\textit{Medical MLLMs}} \\
        \midrule
        \multicolumn{12}{l}{\textit{--- Models < 10B ---}} \\
        MedGemma-1.5-4B-IT\textsuperscript{\dag} & 57.1 & 38.1 & 23.4 & 75.6 & 54.8 & 14.4 & 6.53& 7.74& 13.6 & 10.5 & 11.0 \\
        Lingshu-7B\textsuperscript{\dag} & 61.7 & 68.4 & 18.0 & 70.7 & 48.9 & 7.96 & 4.35 & 5.04 & 11.1 & 9.61 & 9.86 \\
        Hulu-Med-7B\textsuperscript{\dag} & 65.7 & 73.5 & 33.2 & 77.6 & 50.2 & 17.5 & 9.18 &  10.7& 27.4& 16.0&18.3 \\
        \textbf{\method-8B} & \underline{80.2} & \underline{79.1} & \underline{35.7} & \underline{86.3} & \underline{64.1} & \underline{20.1} & \underline{10.7} & \underline{12.0} & \underline{28.8} & \underline{20.5} & \underline{21.6} \\ 
        \textbf{\method-8B + Agentic tools}  & \textbf{85.7} & \textbf{81.3} & \textbf{44.1} & \textbf{89.1} & \textbf{65.5} & \textbf{23.2} & \textbf{17.1} & \textbf{16.6} & \textbf{30.0} & \textbf{23.5} & \textbf{25.3} \\ 
        \midrule
        \multicolumn{12}{l}{\textit{--- Models > 10B ---}} \\
        MedGemma-27B-IT\textsuperscript{\dag} & 54.3 & 36.7 & 25.1 & 72.4 & 45.4 & 11.6 & 6.00 & 7.07 & 13.2 & 10.2 & 10.7 \\
        Lingshu-32B\textsuperscript{\dag} & 63.8 & 70.2 & 25.3 & 70.1 & 50.0 & 9.65 & 6.41 & 7.10 & 15.3 & 11.5 & 12.2 \\
        Hulu-Med-32B\textsuperscript{\dag} & 73.9 & \underline{79.6} & \underline{37.6} & 83.3 & 56.9 & 22.5 & 9.97 & 12.5 & 34.4 & \underline{21.4} & 23.4 \\
        \textbf{\method-32B} & \underline{81.7} & 79.1 & 36.2 & \underline{89.0} & \underline{62.1} & \underline{26.7} & \underline{13.7} & \underline{16.1} & \underline{34.5} & \underline{21.4} & \underline{23.9} \\ 
        \textbf{\method-32B + Agentic tools} &  \textbf{86.3} & \textbf{81.6} & \textbf{45.6} & \textbf{91.7} & \textbf{65.4}  & \textbf{28.4} & \textbf{17.8} & \textbf{19.1} & \textbf{34.9} & \textbf{25.9} & \textbf{27.6}   \\ 
        \bottomrule
    \end{tabular}
    } 
\end{table*}

%% file: tables/medif_bench.tex
\begin{table*}[!t]
    \centering
    \caption{\textbf{Results on our proposed MedIF-Bench}. 
    We report scores on individual tasks and composite instruction-following scores.
    \textbf{Bold} results denote the best scores among medical MLLMs, and \underline{underlined} results denote the second-best scores.
   $\dag$ indicates that results were re-benchmarked using open-source model weights or model APIs within our own evaluation framework. 
    }
    \label{tab:medif_bench}
    \resizebox{\linewidth}{!}{
    \begin{tabular}{l ccccccc ccc}
        \toprule
        & \multicolumn{7}{c}{\textbf{Individual Metrics}} & \multicolumn{3}{c}{\textbf{Composite IF Scores}} \\
        \cmidrule(lr){2-8} \cmidrule(lr){9-11}
        \textbf{Models} & MCQ & \begin{tabular}[c]{@{}c@{}}MCQ with\\context\end{tabular} & Open & Prognosis & \begin{tabular}[c]{@{}c@{}}Report\\Generation\end{tabular} & Seer & Treatment & MM-IF & Text-IF & Overall-IF \\
        \midrule
        \multicolumn{11}{c}{\textit{Proprietary Models}} \\
        \midrule
        GPT-5.2 & 100.0 & 100.0 & 87.0 & 100.0 & 84.0 & 100.0 & 100.0 & 92.0 & 100.0 & 96.0  \\
        Gemini-3-Flash & 98.0 & 100.0 & 100.0 & 98.9 & 97.0 & 83.0 & 97.0 & 98.9 & 94.3 & 96.6  \\
        Claude-Sonnet-4.5 & 73.5 & 76.0 & 90.0 & 96.2 & 80.0 & 100.0 & 100.0 & 86.0 & 87.0 & 86.5 \\
        \midrule
        \multicolumn{11}{c}{\textit{General-purpose MLLMs}} \\
        \midrule
        Qwen2.5-VL-7B\textsuperscript{\dag} & 100.0 & 100.0 & 99.0 & 100.0 & 100.0 & 100.0 & 99.0 & 99.7 & 99.8 & 99.8  \\
        Qwen3-VL-8B\textsuperscript{\dag}  & 99.0 & 100.0 & 99.0 & 100.0 & 99.0 & 95.0 & 100.0 & 99.4 & 98.4 & 98.9 \\
        Qwen2.5-VL-32B\textsuperscript{\dag} & 77.0 & 83.0 & 88.0 & 100.0 & 89.0 & 20.0 & 100.0 & 87.3 & 72.7 & 79.3\\
        Qwen3-VL-32B\textsuperscript{\dag}  & 100.0 & 100.0 & 91.0 & 100.0 & 55.0 & 100.0 & 100.0 & 85.1 & 100.0 & 93.3 \\
        Qwen3-VL-235B\textsuperscript{\dag}  & 100.0 & 100.0 & 99.0 & 100.0 & 100.0 & 99.0 & 100.0 & 99.7 & 99.8 & 99.7 \\
        \midrule
        \multicolumn{11}{c}{\textit{Medical MLLMs}} \\
        \midrule
        \multicolumn{11}{l}{\textit{--- Models < 10B ---}} \\
        MedGemma-1.5-4B-IT\textsuperscript{\dag} & 50.5 & 21.0 & 97.0 & 3.8 & 0.0 & 0.0 & 0.0 & 49.9 & 8.9 & 27.4\\
        Lingshu-7B\textsuperscript{\dag} & 80.5 & 92.0 & 67.0 & 100.0 & 35.0 & 88.0 & 100.0 & 62.3 & 95.5 & 80.4  \\
        Hulu-Med-7B\textsuperscript{\dag}  & 99.0 & 100.0 & 100.0 & 98.9 & 96.0 & 69.0 & 100.0 & 98.9 & 92.0 & \underline{95.1}   \\
        \textbf{\method-8B} & 95.0 & 100.0 & 99.0 & 100.0 & 98.0 & 99.0 & 99.0 & 97.0 & 99.1 & \textbf{98.1} \\
        \midrule
        \multicolumn{11}{l}{\textit{--- Models > 10B ---}} \\
        MedGemma-27B-IT\textsuperscript{\dag}  & 79.0 & 85.0 & 100.0 & 100.0 & 100.0 & 92.0 & 96.0 & 93.4 & 89.8 & 91.4 \\
        Lingshu-32B\textsuperscript{\dag} & 82.5 & 100.0 & 73.0 & 100.0 & 31.0 & 100.0 & 91.0 & 61.7 & 99.8 & 82.6   \\
        Hulu-Med-32B\textsuperscript{\dag} & 100.0 & 100.0 & 98.0 & 100.0 & 89.0 & 100.0 & 100.0 & 96.4 & 100.0 & \underline{98.4} \\
        \textbf{\method-32B} & 100.0 & 100.0 & 91.0 & 100.0 & 100.0 & 100.0 & 100.0 & 97.5 & 100.0 & \textbf{98.9} \\ 
        \bottomrule
    \end{tabular}
    } 
\end{table*}

%% file: tables/ablation_report_generation.tex
\begin{table*}[!t]
\centering
\caption{
\textbf{Ablation study on the necessity of clinical context for report generation.}
We compare two models---\method-8B and Lingshu-7B~\cite{xu2025lingshu}---under two settings: with the extracted clinical context (``w/ Context'') and without it (``w/o Context''), where the Clinical Indication and Area of Focus are removed from the generation prompt.
}
\label{tab:context_ablation}
\resizebox{\textwidth}{!}{%
\begin{tabular}{ll ccc ccc ccc ccc}
\toprule
\multirow{2}{*}{\textbf{Model}} & \multirow{2}{*}{\textbf{Setting}} & \multicolumn{3}{c}{\textbf{AMOS Report}} & \multicolumn{3}{c}{\textbf{CT Rate Report}} & \multicolumn{3}{c}{\textbf{CheXpert-Plus}} & \multicolumn{3}{c}{\textbf{IU-XRAY}} \\
\cmidrule(lr){3-5} \cmidrule(lr){6-8} \cmidrule(lr){9-11} \cmidrule(lr){12-14}
 & & Precision & Recall & F1 & Precision & Recall & F1 & Precision & Recall & F1 & Precision & Recall & F1 \\
\midrule
\multirow{2}{*}{\method-8B} 
  & w/ Context  & {20.1} & {10.7} & {12.0} & {28.8} & {20.5} & {21.6} & {46.4} & {35.3} & {37.8} & {63.3} & {55.8} & {57.3} \\  
 & w/o Context & 0.89 & 0.39 & 0.46 & 7.99 & 3.68 & 4.71 & 13.1 & 10.7 & 10.8 & 10.6 & 8.43 & 8.87 \\
\midrule
\multirow{2}{*}{Lingshu-7B} 
 & w/ Context  & {7.96} & {4.35} & {5.04} & {11.1} & {9.61} & {9.86} & {26.2} & {17.8} & {19.9} & {43.7} & {39.3} & {40.4} \\
 & w/o Context & 0.17 & 0.12 & 0.13 & 5.83 & 4.92 & 4.91 & 11.1 & 8.58 & 8.94 & 10.7 & 8.10 & 8.76 \\
\bottomrule
\end{tabular}%
}
\end{table*}

\begin{table*}[!t]
\centering
\caption{
\textbf{Comparison of RadGraph-F1 and our LLM-as-a-Judge metric on CheXpert-Plus and IU-XRAY.}
Four model variants are evaluated, including a Qwen3-VL-8B (Long Output) variant prompted to encourage verbose generation in order to expose length bias.
RadGraph-F1 reports entity F1 and entity-relation F1, and our LLM-as-a-Judge reports Precision, Recall, and F1.
}
\label{tab:radgraph_failure}
\resizebox{\textwidth}{!}{%
\begin{tabular}{l cccc ccc cccc ccc}
\toprule
\multirow{3}{*}{\textbf{Model}}
    & \multicolumn{7}{c}{\textbf{CheXpert-Plus}}
    & \multicolumn{7}{c}{\textbf{IU-XRAY}} \\
\cmidrule(lr){2-8} \cmidrule(lr){9-15}
    & \multicolumn{2}{c}{\textbf{RadGraph-F1}} 
    & \multicolumn{1}{c}{}
    & \multicolumn{3}{c}{\textbf{LLM-as-a-Judge (Ours)}}
    & \multicolumn{1}{c}{}
    & \multicolumn{2}{c}{\textbf{RadGraph-F1}} 
    & \multicolumn{1}{c}{}
    & \multicolumn{3}{c}{\textbf{LLM-as-a-Judge (Ours)}}
    & \multicolumn{1}{c}{} \\
\cmidrule(lr){2-3} \cmidrule(lr){5-7} \cmidrule(lr){9-10} \cmidrule(lr){12-14}
    & Ent.\ F1 & Ent.-Rel.\ F1 
    &
    & Prec. & Rec. & F1
    &
    & Ent.\ F1 & Ent.-Rel.\ F1
    &
    & Prec. & Rec. & F1
    & \\
\midrule
\method-8B               & 23.2 & 21.0 & & 46.4 & 35.3 & 37.8 & & 38.7 & 36.6 & & 63.3 & 55.8 & 57.3 & \\
Lingshu-7B                & 20.0 & 17.9 & & 26.2 & 17.8 & 19.9 & & 35.1 & 34.6 & & 43.7 & 39.3 & 40.4 & \\
Qwen3-VL-8B               & 13.3 & 11.8 & & 20.0 & 16.3 & 16.8 & & 17.8 & 16.3 & & 37.0 & 37.3 & 36.7 & \\
Qwen3-VL-8B (Long Output) & 20.6 & 18.8 & & 13.8 & 18.9 & 15.1 & & 34.7 & 33.0 & & 29.4 & 42.1 & 33.6 & \\
\bottomrule
\end{tabular}%
}
\end{table*}

%% file: tables/ablation_fusion_mechanism.tex
\begin{table*}[!t]
\centering
\caption{
\textbf{Ablation study on different fusion mechanisms for the compositional vision encoder}. 
We compare the performance of a single encoder (Qwen ViT baseline) against a dual-encoder setup (Qwen ViT + DINOv2) using various fusion strategies. 
All scores are reported as accuracy (\%). 
}
\label{tab:ablation_fusion_mechanism}
\resizebox{\textwidth}{!}{%
\begin{tabular}{llccccccc}
\toprule
\textbf{Encoder Types} & \textbf{Fusion Mechanism} & \textbf{VQA-RAD} & \textbf{PMC-VQA} & \textbf{MedXQA} & \textbf{SLAKE} & \textbf{PathVQA} & \textbf{OMVQA} & \textbf{Average} \\
\midrule
Qwen ViT & - & 69.4 & 57.4 & 24.6 & 81.0 & 58.0 & 83.5 & 62.3 \\
\midrule
\multirow{4}{*}{Qwen ViT + DINOv2} & Local Cross-Attn (Ours) & 72.3 & 57.5 & 25.3 & 84.4 & 65.5 & 82.3 & \textbf{64.5} \\
& Global Cross-Attn & 69.5 & 55.0 & 24.2 & 82.6 & 61.0 & 84.1 & 62.7 \\
& Mixture of Vision Experts & 70.0 & 55.2 & 24.2 & 81.1 & 63.3 & 83.2 & 62.8 \\
& Channel Concat & 70.5 & 54.4 & 24.7 & 83.4 & 63.2 & 82.3 & 63.1 \\
\bottomrule
\end{tabular}%
}
\end{table*}


%% file: tables/ablation_one_additonal_encoder.tex
\begin{table*}[!t]
\centering
\caption{
\textbf{Ablation study on the choice of the specialist 2D encoder}. 
We augment the base Qwen ViT with one additional encoder (DINOv2, ConvNext, or Medsiglip) and evaluate performance on two task categories: VQA and report generation. 
For VQA, we report accuracy (\%). 
For report generation, we report RadGraph-F1~\cite{sellergren2025medgemma} for entities and entity-relations. 
The best average result is highlighted in \textbf{bold}.}
\label{tab:ablation_one_additional_encoder}
\resizebox{\textwidth}{!}{
\begin{tabular}{l ccccccc ccc}
\toprule
\multirow{2}{*}{\textbf{Additional Encoder}} & \multicolumn{7}{c}{\textbf{VQA Benchmarks}} & \multicolumn{3}{c}{\textbf{Report Generation}} \\
\cmidrule(lr){2-8} \cmidrule(lr){9-11}
 & VQA-RAD & PMC-VQA & MedXQA & SLAKE & PathVQA & OMVQA & Average & CheXpert-Plus & IU-XRAY & Average \\
\midrule
DINOv2 & 72.3 & 57.5 & 25.3 & 84.2 & 65.5 & 82.3 & 64.5 & 22.6 / 20.2 & 37.7 / 35.5 & \textbf{30.2 / 27.8} \\
ConvNext & 72.1 & 57.1 & 24.8 & 83.5 & 64.7 & 82.7 & 64.1 & 21.8 / 20.0 & 37.9 / 35.9 & 29.9 / 27.9 \\
Medsiglip & 70.7 & 57.8 & 25.9 & 83.5 & 66.9 & 84.2 & \textbf{64.8} & 23.1 / 20.9 & 34.6 / 32.3 & 28.9 / 26.6 \\
\bottomrule
\end{tabular}
}
\end{table*}

%% file: tables/ablation_multi_additional_encoder.tex
\begin{table*}[!t]
\setlength{\tabcolsep}{3.5pt}
\centering
\caption{
\textbf{Ablation study on the choice of multiple additional encoders, the multi-encoder fusion strategy and fusion order}.
We augment the base Qwen ViT with two additional encoders and evaluate performance on two task categories: VQA and report generation. 
For VQA, we report accuracy (\%). 
For report generation, we report RadGraph-F1~\cite{sellergren2025medgemma} for entities and entity-relations. 
The best average result is highlighted in \textbf{bold}.
}
\label{tab:ablation_multi_additional_encoder}
\resizebox{\textwidth}{!}{
\begin{tabular}{lllcccccccccc}
\toprule
\multirow{2}{*}{\textbf{Fusion Mode}} & \multirow{2}{*}{\shortstack{\textbf{Additional} \\ \textbf{Encoder 1}}} & \multirow{2}{*}{\shortstack{\textbf{Additional} \\ \textbf{Encoder 2}}} & \multicolumn{7}{c}{\textbf{VQA Benchmarks}} & \multicolumn{3}{c}{\textbf{Report Generation}} \\
\cmidrule(lr){4-10} \cmidrule(lr){11-13}
& & & VQA-RAD & PMC-VQA & MedXQA & SLAKE & PathVQA & OMVQA & Average & CheXpert-Plus & IU-XRAY & Average Report \\
\midrule
\multirow{4}{*}{Cascade} & DINOv2 & ConvNext & 70.1 & 57.1 & 25.0 & 84.7 & 66.2 & 83.4 & 64.4 & 23.2 / 21.0 & 38.7 / 36.6 & \textbf{31.0 / 28.8} \\
& ConvNext & DINOv2 & 70.3 & 54.8 & 26.2 & 84.2 & 66.5 & 84.2 & 64.4 & 20.2 / 17.9 & 39.2 / 37.4 & 29.7 / 27.6 \\
& DINOv2 & Medsiglip & 72.7 & 58.0 & 25.2 & 84.7 & 66.5 & 84.4 & \textbf{65.2} & 23.1 / 20.7 & 36.6 / 34.1 & 29.8 / 27.4 \\
& Medsiglip & DINOv2 & 69.0 & 53.6 & 25.5 & 84.6 & 68.0 & 85.9 & 64.4 & 22.4 / 20.0 & 35.1 / 32.5 & 28.8 / 26.3 \\
\midrule
Parallel & DINOv2 & ConvNext & 71.4 & 58.0 & 24.8 & 83.9 & 65.3 & 83.9 & 64.6 & 22.3 / 20.0 & 36.4 / 34.6 & 29.3 / 27.3 \\
\bottomrule
\end{tabular}
}
\end{table*}

%% file: tables/ablation_3d_encoder.tex
\begin{table*}[!t]
\centering
\caption{
\textbf{Ablation study on the necessity of the native 3D encoder during inference.}
We compare the performance of our full model against an ablated version where the 3D encoder is deactivated during inference, forcing the model to rely solely on sampled 2D slices for 3D understanding. 
For VQA tasks, we report accuracy (\%). For report generation, we present Precision/Recall/F1 scores. 
Best results are highlighted in \textbf{bold}.
}
\label{tab:ablation_3d_encoder}
\resizebox{\textwidth}{!}{%
\begin{tabular}{l c cc cc ccc ccc c c}
\toprule
\multirow{2}{*}{\textbf{Regularization}} & \textbf{AMOS} & \multicolumn{2}{c}{\textbf{3D RAD}} & \multicolumn{2}{c}{\textbf{CT Rate}} & \multicolumn{3}{c}{\textbf{AMOS Report}} & \multicolumn{3}{c}{\textbf{CT Rate Report}} & \textbf{Average} & \textbf{Average} \\
\cmidrule(lr){3-4} \cmidrule(lr){5-6} \cmidrule(lr){7-9} \cmidrule(lr){10-12}
 & MCQ & Open & MCQ & Open & MCQ & Precision & Recall & F1 & Precision & Recall & F1 & VQA & Report \\
\midrule
w/ 3D Encoder & 77.6 & 72.6 & 30.0 & 85.3 & 63.6 & 18.1 & 9.67 & 11.0 & 29.3 & 18.4 & 20.3 & \textbf{65.8} & \textbf{23.7 / 14.0 / 15.7} \\
w/o 3D Encoder & 75.8 & 76.3 & 26.4 & 79.1 & 56.4 & 12.1 & 6.82 & 7.86 & 27.0 & 13.4 & 15.5 & 62.8 & 19.6 / 10.3 / 11.7 \\
\bottomrule
\end{tabular}%
}
\end{table*}

%% file: tables/ablation_stochastic_3d.tex
\begin{table*}[!t]
\centering
\caption{
\textbf{Ablation study on stochastic residual regularization for 3D tasks.}
The ``Deterministic'' setting uses standard residual connections, while the ``Stochastic'' setting applies stochastic residual regularization. 
For VQA tasks (AMOS, 3D RAD, CT Rate), we report accuracy (\%). 
For report generation tasks, we present average Precision/Recall/F1 scores. 
Best average results are highlighted in \textbf{bold}.
}
\label{tab:ablation_stochastic_3d}
\resizebox{\textwidth}{!}{%
\begin{tabular}{l c cc cc ccc ccc c c}
\toprule
\multirow{2}{*}{\textbf{Regularization}} & \textbf{AMOS} & \multicolumn{2}{c}{\textbf{3D RAD}} & \multicolumn{2}{c}{\textbf{CT Rate}} & \multicolumn{3}{c}{\textbf{AMOS Report}} & \multicolumn{3}{c}{\textbf{CT Rate Report}} & \textbf{Average} & \textbf{Average} \\
\cmidrule(lr){3-4} \cmidrule(lr){5-6} \cmidrule(lr){7-9} \cmidrule(lr){10-12}
 & MCQ & Open & MCQ & Open & MCQ & Precision & Recall & F1 & Precision & Recall & F1 & VQA & Report \\
\midrule
Stochastic & 77.6 & 72.6 & 30.0 & 85.3 & 63.6 & 18.1 & 9.67 & 11.0 & 29.3 & 18.4 & 20.3 & \textbf{65.8} & \textbf{23.7 / 14.0 / 15.7} \\
Deterministic & 78.8 & 67.9 & 30.4 & 85.9 & 61.2 & 20.3 & 8.40 & 10.6 & 26.5 & 18.2 & 19.6 & 64.8 & 23.4 / 13.3 / 15.1 \\
\bottomrule
\end{tabular}%
}
\end{table*}

%% file: main.bbl
\begin{thebibliography}{128}
\providecommand{\natexlab}[1]{#1}
\providecommand{\url}[1]{\texttt{#1}}
\expandafter\ifx\csname urlstyle\endcsname\relax
  \providecommand{\doi}[1]{doi: #1}\else
  \providecommand{\doi}{doi: \begingroup \urlstyle{rm}\Url}\fi

\bibitem[Bai et~al.(2025{\natexlab{a}})Bai, Chen, Liu, Wang, Ge, Song, Dang,
  Wang, Wang, Tang, et~al.]{bai2025qwen2.5-vl}
Shuai Bai, Keqin Chen, Xuejing Liu, Jialin Wang, Wenbin Ge, Sibo Song, Kai
  Dang, Peng Wang, Shijie Wang, Jun Tang, et~al.
\newblock Qwen2. 5-vl technical report.
\newblock \emph{arXiv preprint arXiv:2502.13923}, 2025{\natexlab{a}}.

\bibitem[Bai et~al.(2025{\natexlab{b}})Bai, Cai, Chen, Chen, Chen, Cheng, Deng,
  Ding, Gao, Ge, Ge, Guo, Huang, Huang, Huang, Hui, Jiang, Li, Li, Li, Li, Lin,
  Lin, Liu, Liu, Liu, Liu, Liu, Liu, Lu, Luo, Lv, Men, Meng, Ren, Ren, Song,
  Sun, Tang, Tu, Wan, Wang, Wang, Wang, Wang, Xie, Xu, Xu, Xu, Yang, Yang,
  Yang, Yang, Yu, Zhang, Zhang, Zhang, Zheng, Zhong, Zhou, Zhou, Zhou, Zhu, and
  Zhu]{qwen3-vl}
Shuai Bai, Yuxuan Cai, Ruizhe Chen, Keqin Chen, Xionghui Chen, Zesen Cheng,
  Lianghao Deng, Wei Ding, Chang Gao, Chunjiang Ge, Wenbin Ge, Zhifang Guo,
  Qidong Huang, Jie Huang, Fei Huang, Binyuan Hui, Shutong Jiang, Zhaohai Li,
  Mingsheng Li, Mei Li, Kaixin Li, Zicheng Lin, Junyang Lin, Xuejing Liu,
  Jiawei Liu, Chenglong Liu, Yang Liu, Dayiheng Liu, Shixuan Liu, Dunjie Lu,
  Ruilin Luo, Chenxu Lv, Rui Men, Lingchen Meng, Xuancheng Ren, Xingzhang Ren,
  Sibo Song, Yuchong Sun, Jun Tang, Jianhong Tu, Jianqiang Wan, Peng Wang,
  Pengfei Wang, Qiuyue Wang, Yuxuan Wang, Tianbao Xie, Yiheng Xu, Haiyang Xu,
  Jin Xu, Zhibo Yang, Mingkun Yang, Jianxin Yang, An~Yang, Bowen Yu, Fei Zhang,
  Hang Zhang, Xi~Zhang, Bo~Zheng, Humen Zhong, Jingren Zhou, Fan Zhou, Jing
  Zhou, Yuanzhi Zhu, and Ke~Zhu.
\newblock Qwen3-vl technical report.
\newblock \emph{arXiv preprint arXiv:2511.21631}, 2025{\natexlab{b}}.

\bibitem[Liu et~al.(2023)Liu, Li, Wu, and Lee]{liu2023llava}
Haotian Liu, Chunyuan Li, Qingyang Wu, and Yong~Jae Lee.
\newblock Visual instruction tuning.
\newblock \emph{Advances in neural information processing systems},
  36:\penalty0 34892--34916, 2023.

\bibitem[Hurst et~al.(2024)Hurst, Lerer, Goucher, Perelman, Ramesh, Clark,
  Ostrow, Welihinda, Hayes, Radford, et~al.]{hurst2024gpt-4o}
Aaron Hurst, Adam Lerer, Adam~P Goucher, Adam Perelman, Aditya Ramesh, Aidan
  Clark, AJ~Ostrow, Akila Welihinda, Alan Hayes, Alec Radford, et~al.
\newblock Gpt-4o system card.
\newblock \emph{arXiv preprint arXiv:2410.21276}, 2024.

\bibitem[Liu et~al.(2024{\natexlab{a}})Liu, Li, Li, and
  Lee]{liu2024improved-llava}
Haotian Liu, Chunyuan Li, Yuheng Li, and Yong~Jae Lee.
\newblock Improved baselines with visual instruction tuning.
\newblock In \emph{Proceedings of the IEEE/CVF conference on computer vision
  and pattern recognition}, pages 26296--26306, 2024{\natexlab{a}}.

\bibitem[Xu et~al.(2025)Xu, Chan, Li, Aljunied, Yuan, Wang, Xiao, Chen, Liu,
  Li, et~al.]{xu2025lingshu}
Weiwen Xu, Hou~Pong Chan, Long Li, Mahani Aljunied, Ruifeng Yuan, Jianyu Wang,
  Chenghao Xiao, Guizhen Chen, Chaoqun Liu, Zhaodonghui Li, et~al.
\newblock Lingshu: A generalist foundation model for unified multimodal medical
  understanding and reasoning.
\newblock \emph{arXiv preprint arXiv:2506.07044}, 2025.

\bibitem[Jiang et~al.(2025)Jiang, Wang, Song, Hu, Zhou, Pu, Zhang, Yang, Feng,
  Zhou, et~al.]{jiang2025hulu-med}
Songtao Jiang, Yuan Wang, Sibo Song, Tianxiang Hu, Chenyi Zhou, Bin Pu, Yan
  Zhang, Zhibo Yang, Yang Feng, Joey~Tianyi Zhou, et~al.
\newblock Hulu-med: A transparent generalist model towards holistic medical
  vision-language understanding.
\newblock \emph{arXiv preprint arXiv:2510.08668}, 2025.

\bibitem[Sellergren et~al.(2025)Sellergren, Kazemzadeh, Jaroensri, Kiraly,
  Traverse, Kohlberger, Xu, Jamil, Hughes, Lau, et~al.]{sellergren2025medgemma}
Andrew Sellergren, Sahar Kazemzadeh, Tiam Jaroensri, Atilla Kiraly, Madeleine
  Traverse, Timo Kohlberger, Shawn Xu, Fayaz Jamil, C{\'\i}an Hughes, Charles
  Lau, et~al.
\newblock Medgemma technical report.
\newblock \emph{arXiv preprint arXiv:2507.05201}, 2025.

\bibitem[Li et~al.(2023{\natexlab{a}})Li, Wong, Zhang, Usuyama, Liu, Yang,
  Naumann, Poon, and Gao]{li2023llava-med}
Chunyuan Li, Cliff Wong, Sheng Zhang, Naoto Usuyama, Haotian Liu, Jianwei Yang,
  Tristan Naumann, Hoifung Poon, and Jianfeng Gao.
\newblock Llava-med: Training a large language-and-vision assistant for
  biomedicine in one day.
\newblock \emph{Advances in Neural Information Processing Systems},
  36:\penalty0 28541--28564, 2023{\natexlab{a}}.

\bibitem[Pan et~al.(2025)Pan, Liu, Wu, Liu, Zhu, Li, Chen, Ouyang, and
  Rueckert]{pan2025medvlm-r1}
Jiazhen Pan, Che Liu, Junde Wu, Fenglin Liu, Jiayuan Zhu, Hongwei~Bran Li, Chen
  Chen, Cheng Ouyang, and Daniel Rueckert.
\newblock Medvlm-r1: Incentivizing medical reasoning capability of
  vision-language models (vlms) via reinforcement learning.
\newblock In \emph{International Conference on Medical Image Computing and
  Computer-Assisted Intervention}, pages 337--347. Springer, 2025.

\bibitem[Lai et~al.(2025)Lai, Zhong, Li, Zhao, Li, Psounis, and
  Yang]{lai2025med-r1}
Yuxiang Lai, Jike Zhong, Ming Li, Shitian Zhao, Yuheng Li, Konstantinos
  Psounis, and Xiaofeng Yang.
\newblock Med-r1: Reinforcement learning for generalizable medical reasoning in
  vision-language models.
\newblock \emph{arXiv preprint arXiv:2503.13939}, 2025.

\bibitem[Chen et~al.(2024{\natexlab{a}})Chen, Gui, Ouyang, Gao, Chen, Chen,
  Wang, Zhang, Cai, Ji, et~al.]{chen2024huatuogpt-vision}
Junying Chen, Chi Gui, Ruyi Ouyang, Anningzhe Gao, Shunian Chen, Guiming~Hardy
  Chen, Xidong Wang, Ruifei Zhang, Zhenyang Cai, Ke~Ji, et~al.
\newblock Huatuogpt-vision, towards injecting medical visual knowledge into
  multimodal llms at scale.
\newblock \emph{arXiv preprint arXiv:2406.19280}, 2024{\natexlab{a}}.

\bibitem[Zhang et~al.(2024{\natexlab{a}})Zhang, Zhou, Adhikarla, Yan, Liu, Yu,
  Liu, Chen, Davison, Ren, et~al.]{zhang2024biomedgpt}
Kai Zhang, Rong Zhou, Eashan Adhikarla, Zhiling Yan, Yixin Liu, Jun Yu,
  Zhengliang Liu, Xun Chen, Brian~D Davison, Hui Ren, et~al.
\newblock A generalist vision--language foundation model for diverse biomedical
  tasks.
\newblock \emph{Nature Medicine}, 30\penalty0 (11):\penalty0 3129--3141,
  2024{\natexlab{a}}.

\bibitem[Hu et~al.(2024)Hu, Li, Lu, Shao, He, Qiao, and Luo]{hu2024omnimedvqa}
Yutao Hu, Tianbin Li, Quanfeng Lu, Wenqi Shao, Junjun He, Yu~Qiao, and Ping
  Luo.
\newblock Omnimedvqa: A new large-scale comprehensive evaluation benchmark for
  medical lvlm.
\newblock In \emph{Proceedings of the IEEE/CVF Conference on Computer Vision
  and Pattern Recognition (CVPR)}, pages 22170--22183, June 2024.
\newblock URL
  \url{https://openaccess.thecvf.com/content/CVPR2024/papers/Hu_OmniMedVQA_A_New_Large-Scale_Comprehensive_Evaluation_Benchmark_for_Medical_LVLM_CVPR_2024_paper.pdf}.

\bibitem[Liu et~al.(2021)Liu, Zhan, Xu, Ma, Yang, and Wu]{liu2021slake}
Bo~Liu, Li-Ming Zhan, Li~Xu, Lin Ma, Yan Yang, and Xiao-Ming Wu.
\newblock Slake: A semantically-labeled knowledge-enhanced dataset for medical
  visual question answering.
\newblock In \emph{2021 IEEE 18th International Symposium on Biomedical Imaging
  (ISBI)}, pages 1650--1654, 2021.
\newblock URL \url{https://ieeexplore.ieee.org/document/9434010}.

\bibitem[Hamamci et~al.(2026)Hamamci, Er, Wang, Almas, Simsek, Esirgun, Dogan,
  Durugol, Hou, Shit, et~al.]{hamamci2024CT-Rate}
Ibrahim~Ethem Hamamci, Sezgin Er, Chenyu Wang, Furkan Almas, Ayse~Gulnihan
  Simsek, Sevval~Nil Esirgun, Irem Dogan, Omer~Faruk Durugol, Benjamin Hou,
  Suprosanna Shit, et~al.
\newblock Generalist foundation models from a multimodal dataset for 3d
  computed tomography.
\newblock \emph{Nature Biomedical Engineering}, pages 1--19, 2026.

\bibitem[Hendrycks et~al.(2021)Hendrycks, Burns, Basart, Zou, Mazeika, Song,
  and Steinhardt]{hendrycks2021mmlu}
Dan Hendrycks, Collin Burns, Steven Basart, Andy Zou, Mantas Mazeika, Dawn
  Song, and Jacob Steinhardt.
\newblock Measuring massive multitask language understanding.
\newblock In \emph{International Conference on Learning Representations}, 2021.
\newblock URL \url{https://openreview.net/forum?id=d7KBjmI3GmQ}.

\bibitem[Zuo et~al.(2025)Zuo, Qu, Li, Chen, Zhu, Hua, Zhang, Ding, and
  Zhou]{zuo2025medxpertqa}
Yuxin Zuo, Shang Qu, Yifei Li, Zhang-Ren Chen, Xuekai Zhu, Ermo Hua, Kaiyan
  Zhang, Ning Ding, and Bowen Zhou.
\newblock Medxpertqa: Benchmarking expert-level medical reasoning and
  understanding.
\newblock In \emph{International Conference on Machine Learning}, pages
  80961--80990. PMLR, 2025.

\bibitem[Jin et~al.(2019)Jin, Dhingra, Liu, Cohen, and Lu]{jin2019pubmedqa}
Qiao Jin, Bhuwan Dhingra, Zhengping Liu, William Cohen, and Xinghua Lu.
\newblock Pubmedqa: A dataset for biomedical research question answering.
\newblock In \emph{Proceedings of the 2019 conference on empirical methods in
  natural language processing and the 9th international joint conference on
  natural language processing (EMNLP-IJCNLP)}, pages 2567--2577, 2019.

\bibitem[Li et~al.(2025{\natexlab{a}})Li, Chang, Yang, Wu, Chen, Bansal, Chen,
  Yang, Chen, Chen, et~al.]{sun2025holistic_3d_brain_ct}
Cheng-Yi Li, Kao-Jung Chang, Cheng-Fu Yang, Hsin-Yu Wu, Wenting Chen, Hritik
  Bansal, Ling Chen, Yi-Ping Yang, Yu-Chun Chen, Shih-Pin Chen, et~al.
\newblock Towards a holistic framework for multimodal {LLM} in {3D} brain {CT}
  radiology report generation.
\newblock \emph{Nature Communications}, 16\penalty0 (1):\penalty0 2258,
  2025{\natexlab{a}}.
\newblock \doi{10.1038/s41467-025-57426-0}.

\bibitem[Xin et~al.(2025)Xin, Ates, Gong, and Shao]{xin2025med3dvlm}
Yu~Xin, Gorkem~Can Ates, Kuang Gong, and Wei Shao.
\newblock Med3dvlm: An efficient vision-language model for 3d medical image
  analysis.
\newblock \emph{IEEE Journal of Biomedical and Health Informatics}, 2025.

\bibitem[Wu et~al.(2025{\natexlab{a}})Wu, Zhang, Zhang, Hui, Wang, and
  Xie]{wu2025RadFM}
Chaoyi Wu, Xiaoman Zhang, Ya~Zhang, Hui Hui, Yanfeng Wang, and Weidi Xie.
\newblock Towards generalist foundation model for radiology by leveraging
  web-scale 2d\&3d medical data.
\newblock \emph{Nature Communications}, 16\penalty0 (1):\penalty0 7866,
  2025{\natexlab{a}}.

\bibitem[Pai et~al.(2024)Pai, Bontempi, Hadzic, Prudente, Soka{\v{c}},
  Chaunzwa, Bernatz, Hosny, Mak, Birkbak, and Aerts]{pai2024foundation_cancer}
Suraj Pai, Dennis Bontempi, Ibrahim Hadzic, Vasco Prudente, Mateo Soka{\v{c}},
  Tafadzwa~L. Chaunzwa, Simon Bernatz, Ahmed Hosny, Raymond~H. Mak, Nicolai~J.
  Birkbak, and Hugo J. W.~L. Aerts.
\newblock Foundation model for cancer imaging biomarkers.
\newblock \emph{Nature Machine Intelligence}, 6\penalty0 (3):\penalty0
  354--367, 2024.
\newblock \doi{10.1038/s42256-024-00807-9}.

\bibitem[Shi et~al.(2025)Shi, Liu, Wang, Liao, Radhakrishnan, Zhao, Huang, Yin,
  Sapra, Yacoob, et~al.]{shi2024eagle}
Min Shi, Fuxiao Liu, Shihao Wang, Shijia Liao, Subhashree Radhakrishnan, Yilin
  Zhao, De-An Huang, Hongxu Yin, Karan Sapra, Yaser Yacoob, et~al.
\newblock Eagle: Exploring the design space for multimodal llms with mixture of
  encoders.
\newblock In \emph{International Conference on Learning Representations},
  volume 2025, pages 92628--92646, 2025.

\bibitem[Li et~al.(2024{\natexlab{a}})Li, Wei, Zhang, and Zhang]{li2025eagle-2}
Yuhui Li, Fangyun Wei, Chao Zhang, and Hongyang Zhang.
\newblock Eagle-2: Faster inference of language models with dynamic draft
  trees.
\newblock In \emph{Proceedings of the 2024 conference on empirical methods in
  natural language processing}, pages 7421--7432, 2024{\natexlab{a}}.

\bibitem[Tong et~al.(2024)Tong, Brown, Wu, Woo, IYER, Akula, Yang, Yang,
  Middepogu, Wang, et~al.]{tong2024cambrian-1}
Peter Tong, Ellis Brown, Penghao Wu, Sanghyun Woo, Adithya Jairam~Vedagiri
  IYER, Sai~Charitha Akula, Shusheng Yang, Jihan Yang, Manoj Middepogu, Ziteng
  Wang, et~al.
\newblock Cambrian-1: A fully open, vision-centric exploration of multimodal
  llms.
\newblock \emph{Advances in Neural Information Processing Systems},
  37:\penalty0 87310--87356, 2024.

\bibitem[Li et~al.(2025{\natexlab{b}})Li, Zhang, Wang, Zhong, Chen, Chu, Liu,
  and Jia]{li2025mini-gemini}
Yanwei Li, Yuechen Zhang, Chengyao Wang, Zhisheng Zhong, Yixin Chen, Ruihang
  Chu, Shaoteng Liu, and Jiaya Jia.
\newblock Mini-gemini: Mining the potential of multi-modality vision language
  models.
\newblock \emph{IEEE Transactions on Pattern Analysis and Machine
  Intelligence}, 2025{\natexlab{b}}.

\bibitem[Karamcheti et~al.(2024)Karamcheti, Nair, Balakrishna, Liang, Kollar,
  and Sadigh]{karamcheti2024prismatic_vlms}
Siddharth Karamcheti, Suraj Nair, Ashwin Balakrishna, Percy Liang, Thomas
  Kollar, and Dorsa Sadigh.
\newblock Prismatic vlms: Investigating the design space of
  visually-conditioned language models.
\newblock In \emph{Forty-first International Conference on Machine Learning},
  2024.

\bibitem[Luo et~al.(2025)Luo, Zhou, Zhang, Zheng, Sun, and
  Ji]{luo2024Mixture-of-Resolution-Adaptation}
Gen Luo, Yiyi Zhou, Yuxin Zhang, Xiawu Zheng, Xiaoshuai Sun, and Rongrong Ji.
\newblock Feast your eyes: Mixture-of-resolution adaptation for multimodal
  large language models.
\newblock In \emph{International Conference on Learning Representations},
  volume 2025, pages 84491--84506, 2025.

\bibitem[Lin(2004)]{lin2004rouge}
Chin-Yew Lin.
\newblock Rouge: A package for automatic evaluation of summaries.
\newblock In \emph{Text summarization branches out}, pages 74--81, 2004.

\bibitem[Banerjee and Lavie(2005)]{banerjee2005METEOR}
Satanjeev Banerjee and Alon Lavie.
\newblock Meteor: An automatic metric for mt evaluation with improved
  correlation with human judgments.
\newblock In \emph{Proceedings of the acl workshop on intrinsic and extrinsic
  evaluation measures for machine translation and/or summarization}, pages
  65--72, 2005.

\bibitem[Vedantam et~al.(2015)Vedantam, Lawrence~Zitnick, and
  Parikh]{vedantam2015cider}
Ramakrishna Vedantam, C~Lawrence~Zitnick, and Devi Parikh.
\newblock Cider: Consensus-based image description evaluation.
\newblock In \emph{Proceedings of the IEEE conference on computer vision and
  pattern recognition}, pages 4566--4575, 2015.

\bibitem[Zhang et~al.(2020)Zhang, Kishore, Wu, Weinberger, and
  Artzi]{Zhang2020BERTScore}
Tianyi Zhang, Varsha Kishore, Felix Wu, Kilian~Q. Weinberger, and Yoav Artzi.
\newblock Bertscore: Evaluating text generation with bert.
\newblock In \emph{International Conference on Learning Representations}, 2020.

\bibitem[Delbrouck et~al.(2022)Delbrouck, Chambon, Bluethgen, Tsai, Almusa, and
  Langlotz]{delbrouck-etal-2022-improving}
Jean-Benoit Delbrouck, Pierre Chambon, Christian Bluethgen, Emily Tsai, Omar
  Almusa, and Curtis Langlotz.
\newblock Improving the factual correctness of radiology report generation with
  semantic rewards.
\newblock In \emph{Findings of the Association for Computational Linguistics:
  EMNLP 2022}, pages 4348--4360, Abu Dhabi, United Arab Emirates, December
  2022. Association for Computational Linguistics.
\newblock URL \url{https://aclanthology.org/2022.findings-emnlp.319}.

\bibitem[Zhao et~al.(2024)Zhao, Wu, Zhang, Zhang, Wang, and
  Xie]{zhao2024ratescore}
Weike Zhao, Chaoyi Wu, Xiaoman Zhang, Ya~Zhang, Yanfeng Wang, and Weidi Xie.
\newblock Ratescore: A metric for radiology report generation.
\newblock In \emph{Proceedings of the 2024 Conference on Empirical Methods in
  Natural Language Processing}, pages 15004--15019, 2024.

\bibitem[Ostmeier et~al.(2024)Ostmeier, Xu, Chen, Varma, Blankemeier,
  Bluethgen, Md, Moseley, Langlotz, Chaudhari, et~al.]{ostmeier2024GREEN}
Sophie Ostmeier, Justin Xu, Zhihong Chen, Maya Varma, Louis Blankemeier,
  Christian Bluethgen, Arne Edward~Michalson Md, Michael Moseley, Curtis
  Langlotz, Akshay~S Chaudhari, et~al.
\newblock Green: Generative radiology report evaluation and error notation.
\newblock In \emph{Findings of the association for computational linguistics:
  EMNLP 2024}, pages 374--390, 2024.

\bibitem[Qiu et~al.(2024)Qiu, Lam, Li, Acharya, Wong, Darzi, Yuan, and
  Topol]{qiu2024agentic}
Jianing Qiu, Kyle Lam, Guohao Li, Amish Acharya, Tien~Yin Wong, Ara Darzi,
  Wu~Yuan, and Eric~J. Topol.
\newblock Llm-based agentic systems in medicine and healthcare.
\newblock \emph{Nature Machine Intelligence}, 6\penalty0 (12):\penalty0
  1418--1420, 2024.

\bibitem[Wang et~al.(2025{\natexlab{a}})Wang, Ma, Wang, Wu, Ji, Chen, Li, and
  Yuan]{wange2025baymax}
Wenxuan Wang, Zizhan Ma, Zheng Wang, Chenghan Wu, Jiaming Ji, Wenting Chen,
  Xiang Li, and Yixuan Yuan.
\newblock A survey of llm-based agents in medicine: How far are we from baymax?
\newblock \emph{Findings of the Association for Computational Linguistics: ACL
  2025}, pages 10345--10359, 2025{\natexlab{a}}.

\bibitem[Wang et~al.(2025{\natexlab{b}})Wang, Wu, Cai, Low, Yang, Li, and
  Jin]{wang2025medagentpro}
Ziyue Wang, Junde Wu, Linghan Cai, Chang~Han Low, Xihong Yang, Qiaxuan Li, and
  Yueming Jin.
\newblock Medagent-pro: Towards evidence-based multi-modal medical diagnosis
  via reasoning agentic workflow.
\newblock \emph{arXiv preprint arXiv:2503.18968}, 2025{\natexlab{b}}.

\bibitem[Li et~al.(2024{\natexlab{b}})Li, Yan, Pan, Luo, Ji, Ding, Xu, Liu,
  Dong, Lin, et~al.]{li2024mmedagent}
Binxu Li, Tiankai Yan, Yuanting Pan, Jie Luo, Ruiyang Ji, Jiayuan Ding, Zhe Xu,
  Shilong Liu, Haoyu Dong, Zihao Lin, et~al.
\newblock Mmedagent: Learning to use medical tools with multi-modal agent.
\newblock In \emph{Findings of the Association for Computational Linguistics:
  EMNLP 2024}, pages 8745--8760, 2024{\natexlab{b}}.

\bibitem[Zhu et~al.(2025)Zhu, Wang, Chen, Liu, Ye, Gu, Tian, Duan, Su, Shao,
  et~al.]{zhu2025internvl3}
Jinguo Zhu, Weiyun Wang, Zhe Chen, Zhaoyang Liu, Shenglong Ye, Lixin Gu, Hao
  Tian, Yuchen Duan, Weijie Su, Jie Shao, et~al.
\newblock Internvl3: Exploring advanced training and test-time recipes for
  open-source multimodal models.
\newblock \emph{arXiv preprint arXiv:2504.10479}, 2025.

\bibitem[Mullappilly et~al.(2025)Mullappilly, Kurpath, Pieri, Alseiari,
  Cholakkal, Aldahmani, Khan, Anwer, Khan, Baldwin,
  et~al.]{mullappilly2024bimedix2}
Sahal~Shaji Mullappilly, Mohammed~Irfan Kurpath, Sara Pieri, Saeed~Yahya
  Alseiari, Shanavas Cholakkal, Khaled Aldahmani, Fahad Khan, Rao Anwer, Salman
  Khan, Timothy Baldwin, et~al.
\newblock Bimedix2: Bio-medical expert lmm for diverse medical modalities.
\newblock In \emph{Findings of the association for computational linguistics:
  EMNLP 2025}, pages 14051--14071, 2025.

\bibitem[Lin et~al.(2025)Lin, Zhang, Li, Yuan, Yu, Li, He, Jiang, Li, Xiaohui,
  et~al.]{lin2025healthgpt}
Tianwei Lin, Wenqiao Zhang, Sijing Li, Yuqian Yuan, Binhe Yu, Haoyuan Li,
  Wanggui He, Hao Jiang, Mengze Li, Song Xiaohui, et~al.
\newblock Healthgpt: A medical large vision-language model for unifying
  comprehension and generation via heterogeneous knowledge adaptation.
\newblock In \emph{International Conference on Machine Learning}, pages
  37975--37995. PMLR, 2025.

\bibitem[He et~al.(2024)He, Nie, Wang, Yang, Wang, Cai, Chen, Xu, Luo, Xiang,
  et~al.]{he2024meddr}
Sunan He, Yuxiang Nie, Hongmei Wang, Shu Yang, Yihui Wang, Zhiyuan Cai, Zhixuan
  Chen, Yingxue Xu, Luyang Luo, Huiling Xiang, et~al.
\newblock Gsco: Towards generalizable ai in medicine via generalist-specialist
  collaboration.
\newblock \emph{arXiv preprint arXiv:2404.15127}, 2024.

\bibitem[Tanno et~al.(2025)Tanno, Barrett, Sellergren, Ghaisas, Dathathri, See,
  Welbl, Lau, Tu, Azizi, et~al.]{tanno2025flamingo_cxr}
Ryutaro Tanno, David G.~T. Barrett, Andrew Sellergren, Sumedh Ghaisas, Sumanth
  Dathathri, Abigail See, Johannes Welbl, Charles Lau, Tao Tu, Shekoofeh Azizi,
  et~al.
\newblock Collaboration between clinicians and vision--language models in
  radiology report generation.
\newblock \emph{Nature Medicine}, 31\penalty0 (2):\penalty0 599--608, 2025.
\newblock \doi{10.1038/s41591-024-03302-1}.

\bibitem[Chambon et~al.(2024)Chambon, Delbrouck, Sounack, Huang, Chen, Varma,
  Truong, Chuong, and Langlotz]{chambon2024CheXpertPlus}
Pierre Chambon, Jean-Benoit Delbrouck, Thomas Sounack, Shih-Cheng Huang,
  Zhihong Chen, Maya Varma, Steven~QH Truong, Chu~The Chuong, and Curtis~P.
  Langlotz.
\newblock Chexpert plus: Augmenting a large chest x-ray dataset with text
  radiology reports, patient demographics and additional image formats.
\newblock \emph{ArXiv}, abs/2405.19538, 2024.
\newblock URL \url{https://arxiv.org/abs/2405.19538}.

\bibitem[Gai et~al.(2025{\natexlab{a}})Gai, Liu, Li, Meng, Wu, and
  Liu]{gai2025_3D-RAD}
Xiaotang Gai, Jiaxiang Liu, Yichen Li, Zijie Meng, Jian Wu, and Zuozhu Liu.
\newblock 3d-rad: A comprehensive 3d radiology med-vqa dataset with
  multi-temporal analysis and diverse diagnostic tasks.
\newblock \emph{arXiv preprint arXiv:2506.11147}, 2025{\natexlab{a}}.

\bibitem[Ji et~al.(2022)Ji, Bai, Ge, Yang, Zhu, Zhang, Li, Zhanng, Ma, Wan,
  et~al.]{ji2022AMOS}
Yuanfeng Ji, Haotian Bai, Chongjian Ge, Jie Yang, Ye~Zhu, Ruimao Zhang, Zhen
  Li, Lingyan Zhanng, Wanling Ma, Xiang Wan, et~al.
\newblock Amos: A large-scale abdominal multi-organ benchmark for versatile
  medical image segmentation.
\newblock \emph{Advances in neural information processing systems},
  35:\penalty0 36722--36732, 2022.

\bibitem[Papineni et~al.(2002)Papineni, Roukos, Ward, and
  Zhu]{papineni2002bleu}
Kishore Papineni, Salim Roukos, Todd Ward, and Wei-Jing Zhu.
\newblock Bleu: a method for automatic evaluation of machine translation.
\newblock In \emph{Proceedings of the 40th annual meeting of the Association
  for Computational Linguistics}, pages 311--318, 2002.

\bibitem[Delbrouck et~al.(2024)Delbrouck, Chambon, Chen, Varma, Johnston,
  Blankemeier, Van~Veen, Bui, Truong, and
  Langlotz]{delbrouck-etal-2024-radgraph}
Jean-Benoit Delbrouck, Pierre Chambon, Zhihong Chen, Maya Varma, Andrew
  Johnston, Louis Blankemeier, Dave Van~Veen, Tan Bui, Steven Truong, and
  Curtis Langlotz.
\newblock {R}ad{G}raph-{XL}: A large-scale expert-annotated dataset for entity
  and relation extraction from radiology reports.
\newblock In Lun-Wei Ku, Andre Martins, and Vivek Srikumar, editors,
  \emph{Findings of the Association for Computational Linguistics ACL 2024},
  pages 12902--12915, Bangkok, Thailand and virtual meeting, August 2024.
  Association for Computational Linguistics.
\newblock URL \url{https://aclanthology.org/2024.findings-acl.765}.

\bibitem[Fedus et~al.(2022)Fedus, Zoph, and
  Shazeer]{fedus2022Switch_transformers}
William Fedus, Barret Zoph, and Noam Shazeer.
\newblock Switch transformers: Scaling to trillion parameter models with simple
  and efficient sparsity.
\newblock \emph{Journal of Machine Learning Research}, 23\penalty0
  (120):\penalty0 1--39, 2022.

\bibitem[Du and Kaelbling(2024)]{du2024compositional}
Yilun Du and Leslie~Pack Kaelbling.
\newblock Position: Compositional generative modeling: A single model is not
  all you need.
\newblock In \emph{International Conference on Machine Learning}, pages
  11721--11732. PMLR, 2024.

\bibitem[Liu et~al.(2022)Liu, Mao, Wu, Feichtenhofer, Darrell, and
  Xie]{liu2022convnet}
Zhuang Liu, Hanzi Mao, Chao-Yuan Wu, Christoph Feichtenhofer, Trevor Darrell,
  and Saining Xie.
\newblock A convnet for the 2020s.
\newblock \emph{Proceedings of the IEEE/CVF Conference on Computer Vision and
  Pattern Recognition (CVPR)}, 2022.

\bibitem[Oquab et~al.(2024)Oquab, Darcet, Moutakanni, Vo, Szafraniec, Khalidov,
  Fernandez, HAZIZA, Massa, El-Nouby, Assran, Ballas, Galuba, Howes, Huang, Li,
  Misra, Rabbat, Sharma, Synnaeve, Xu, Jegou, Mairal, Labatut, Joulin, and
  Bojanowski]{oquab2023dinov2}
Maxime Oquab, Timoth{\'e}e Darcet, Th{\'e}o Moutakanni, Huy~V. Vo, Marc
  Szafraniec, Vasil Khalidov, Pierre Fernandez, Daniel HAZIZA, Francisco Massa,
  Alaaeldin El-Nouby, Mido Assran, Nicolas Ballas, Wojciech Galuba, Russell
  Howes, Po-Yao Huang, Shang-Wen Li, Ishan Misra, Michael Rabbat, Vasu Sharma,
  Gabriel Synnaeve, Hu~Xu, Herve Jegou, Julien Mairal, Patrick Labatut, Armand
  Joulin, and Piotr Bojanowski.
\newblock {DINO}v2: Learning robust visual features without supervision.
\newblock \emph{Transactions on Machine Learning Research}, 2024.
\newblock ISSN 2835-8856.
\newblock URL \url{https://openreview.net/forum?id=a68SUt6zFt}.
\newblock Featured Certification.

\bibitem[P{\'e}rez-Garc{\'\i}a et~al.(2025)P{\'e}rez-Garc{\'\i}a, Sharma,
  Bond-Taylor, Bouzid, Salvatelli, Ilse, Bannur, Castro, Schwaighofer, Lungren,
  et~al.]{perez2025RAD-DINO}
Fernando P{\'e}rez-Garc{\'\i}a, Harshita Sharma, Sam Bond-Taylor, Kenza Bouzid,
  Valentina Salvatelli, Maximilian Ilse, Shruthi Bannur, Daniel~C Castro, Anton
  Schwaighofer, Matthew~P Lungren, et~al.
\newblock Exploring scalable medical image encoders beyond text supervision.
\newblock \emph{Nature Machine Intelligence}, 7\penalty0 (1):\penalty0
  119--130, 2025.

\bibitem[Zhang et~al.(2025{\natexlab{a}})Zhang, Xu, Usuyama, Xu, Bagga, Tinn,
  Preston, Rao, Wei, Valluri, et~al.]{zhang2025BiomedCLIP}
Sheng Zhang, Yanbo Xu, Naoto Usuyama, Hanwen Xu, Jaspreet Bagga, Robert Tinn,
  Sam Preston, Rajesh Rao, Mu~Wei, Naveen Valluri, et~al.
\newblock A multimodal biomedical foundation model trained from fifteen million
  image--text pairs.
\newblock \emph{Nejm Ai}, 2\penalty0 (1):\penalty0 AIoa2400640,
  2025{\natexlab{a}}.

\bibitem[Yang et~al.(2025{\natexlab{a}})Yang, Chen, Tian, Wang, Li, Yu, and
  Jia]{yang2025visionzip}
Senqiao Yang, Yukang Chen, Zhuotao Tian, Chengyao Wang, Jingyao Li, Bei Yu, and
  Jiaya Jia.
\newblock Visionzip: Longer is better but not necessary in vision language
  models.
\newblock In \emph{Proceedings of the Computer Vision and Pattern Recognition
  Conference}, pages 19792--19802, 2025{\natexlab{a}}.

\bibitem[Zhang et~al.(2025{\natexlab{b}})Zhang, Fan, Ma, Zheng, Huang, Cheng,
  Gudovskiy, Okuno, Nakata, Keutzer, et~al.]{zhang2024sparsevlm}
Yuan Zhang, Chun-Kai Fan, Junpeng Ma, Wenzhao Zheng, Tao Huang, Kuan Cheng,
  Denis~A Gudovskiy, Tomoyuki Okuno, Yohei Nakata, Kurt Keutzer, et~al.
\newblock Sparsevlm: Visual token sparsification for efficient vision-language
  model inference.
\newblock In \emph{International Conference on Machine Learning}, pages
  74840--74857. PMLR, 2025{\natexlab{b}}.

\bibitem[Vaswani et~al.(2017)Vaswani, Shazeer, Parmar, Uszkoreit, Jones, Gomez,
  Kaiser, and Polosukhin]{AttentionAlluNeed}
Ashish Vaswani, Noam Shazeer, Niki Parmar, Jakob Uszkoreit, Llion Jones,
  Aidan~N Gomez, {\L}ukasz Kaiser, and Illia Polosukhin.
\newblock Attention is all you need.
\newblock In \emph{NeurIPS}, pages 5998--6008, 2017.

\bibitem[Meng et~al.(2024)Meng, Yang, Tian, Dai, Wu, Gao, and
  Jiang]{meng2024deepstack}
Lingchen Meng, Jianwei Yang, Rui Tian, Xiyang Dai, Zuxuan Wu, Jianfeng Gao, and
  Yu-Gang Jiang.
\newblock Deepstack: Deeply stacking visual tokens is surprisingly simple and
  effective for lmms.
\newblock \emph{Advances in Neural Information Processing Systems},
  37:\penalty0 23464--23487, 2024.

\bibitem[Bolya et~al.(2025)Bolya, Huang, Sun, Cho, Madotto, Wei, Ma, Zhi,
  Rajasegaran, Rasheed, et~al.]{bolya2025perception_encoder}
Daniel Bolya, Po-Yao Huang, Peize Sun, Jang~Hyun Cho, Andrea Madotto, Chen Wei,
  Tengyu Ma, Jiale Zhi, Jathushan Rajasegaran, Hanoona Rasheed, et~al.
\newblock Perception encoder: The best visual embeddings are not at the output
  of the network.
\newblock \emph{arXiv preprint arXiv:2504.13181}, 2025.

\bibitem[Zhou et~al.(2022)Zhou, Chen, Zhang, Luo, Wang, and
  Yu]{zhou2022generalized_radiograph}
Hong-Yu Zhou, Xiaoyu Chen, Yinghao Zhang, Ruibang Luo, Liansheng Wang, and
  Yizhou Yu.
\newblock Generalized radiograph representation learning via cross-supervision
  between images and free-text radiology reports.
\newblock \emph{Nature Machine Intelligence}, 4\penalty0 (1):\penalty0 32--40,
  2022.
\newblock \doi{10.1038/s42256-021-00425-9}.

\bibitem[Zhang et~al.(2023{\natexlab{a}})Zhang, Chen, Jiang, Yu, Chen, Chen,
  Li, Wu, Zhiyi, Xiao, et~al.]{zhang2023huatuogpt}
Hongbo Zhang, Junying Chen, Feng Jiang, Fei Yu, Zhihong Chen, Guiming Chen,
  Jianquan Li, Xiangbo Wu, Zhang Zhiyi, Qingying Xiao, et~al.
\newblock Huatuogpt, towards taming language model to be a doctor.
\newblock In \emph{Findings of the association for computational linguistics:
  EMNLP 2023}, pages 10859--10885, 2023{\natexlab{a}}.

\bibitem[Mizrahi et~al.(2024)Mizrahi, Kaplan, Malkin, Dror, Shahaf, and
  Stanovsky]{mizrahi2024state}
Moran Mizrahi, Guy Kaplan, Dan Malkin, Rotem Dror, Dafna Shahaf, and Gabriel
  Stanovsky.
\newblock State of what art? a call for multi-prompt llm evaluation.
\newblock \emph{Transactions of the Association for Computational Linguistics},
  12:\penalty0 933--949, 2024.

\bibitem[{National Cancer Institute, Surveillance Research
  Program}(2025)]{seer_program}
{National Cancer Institute, Surveillance Research Program}.
\newblock Surveillance, epidemiology, and end results (seer) program
  (www.seer.cancer.gov), 2025.
\newblock URL \url{https://seer.cancer.gov/data/}.
\newblock SEER database; data analyzed using SEER*Stat. Visit
  https://seer.cancer.gov/data/.

\bibitem[Jian et~al.(2025)Jian, Wu, Sun, Wang, Ren, and Zhang]{jian2025look}
Pu~Jian, Junhong Wu, Wei Sun, Chen Wang, Shuo Ren, and Jiajun Zhang.
\newblock Look again, think slowly: Enhancing visual reflection in
  vision-language models.
\newblock In \emph{Proceedings of the 2025 Conference on Empirical Methods in
  Natural Language Processing}, pages 9262--9281, 2025.

\bibitem[Bai et~al.(2024)Bai, Du, Huang, Meng, and Zhao]{bai2024m3d}
Fan Bai, Yuxin Du, Tiejun Huang, Max Q.-H. Meng, and Bo~Zhao.
\newblock M3d: Advancing 3d medical image analysis with multi-modal large
  language models.
\newblock \emph{arXiv preprint arXiv:2404.00578}, 2024.

\bibitem[Blankemeier et~al.(2026)Blankemeier, Kumar, Cohen, Liu, Liu, Van~Veen,
  Gardezi, Yu, Paschali, Chen, et~al.]{blankemeier2024merlin}
Louis Blankemeier, Ashwin Kumar, Joseph~Paul Cohen, Jiaming Liu, Longchao Liu,
  Dave Van~Veen, Syed Jamal~Safdar Gardezi, Hongkun Yu, Magdalini Paschali,
  Zhihong Chen, et~al.
\newblock Merlin: a computed tomography vision--language foundation model and
  dataset.
\newblock \emph{Nature}, pages 1--11, 2026.

\bibitem[Huang et~al.(2023)Huang, Huo, Steinberg, Chiang, Langlotz, Lungren,
  Yeung, Shah, and Fries]{huang2023inspect}
Shih-Cheng Huang, Zepeng Huo, Ethan Steinberg, Chia-Chun Chiang, Curtis
  Langlotz, Matthew Lungren, Serena Yeung, Nigam Shah, and Jason Fries.
\newblock Inspect: A multimodal dataset for patient outcome prediction of
  pulmonary embolisms.
\newblock In \emph{Advances in Neural Information Processing Systems 36
  (NeurIPS), Datasets and Benchmarks Track}, pages 17742--17772, 2023.
\newblock URL
  \url{https://proceedings.neurips.cc/paper_files/paper/2023/hash/39736af1b9d87a1fddad9f84a6bcf64c-Abstract-Datasets_and_Benchmarks.html}.

\bibitem[He et~al.(2025)He, Guo, Tang, Myronenko, Nath, Xu, Yang, Zhao, Simon,
  Belue, et~al.]{he2025vista3d}
Yufan He, Pengfei Guo, Yucheng Tang, Andriy Myronenko, Vishwesh Nath, Ziyue Xu,
  Dong Yang, Can Zhao, Benjamin Simon, Mason Belue, et~al.
\newblock Vista3d: A unified segmentation foundation model for 3d medical
  imaging.
\newblock In \emph{Proceedings of the Computer Vision and Pattern Recognition
  Conference}, pages 20863--20873, 2025.

\bibitem[Wasserthal et~al.(2023)Wasserthal, Breit, Meyer, Pradella, Hinck,
  Sauter, Heye, Boll, Cyriac, Yang, et~al.]{wasserthal2023totalsegmentator}
Jakob Wasserthal, Hanns-Christian Breit, Manfred~T Meyer, Maurice Pradella,
  Daniel Hinck, Alexander~W Sauter, Tobias Heye, Daniel~T Boll, Joshy Cyriac,
  Shan Yang, et~al.
\newblock Totalsegmentator: robust segmentation of 104 anatomic structures in
  ct images.
\newblock \emph{Radiology: Artificial Intelligence}, 5\penalty0 (5):\penalty0
  e230024, 2023.

\bibitem[Li et~al.(2024{\natexlab{c}})Li, Wang, Jiang, Mao, Chen, Du, Zhang,
  Zhang, Zhang, and Liu]{li2024dmqr-rag}
Zhicong Li, Jiahao Wang, Zhishu Jiang, Hangyu Mao, Zhongxia Chen, Jiazhen Du,
  Yuanxing Zhang, Fuzheng Zhang, Di~Zhang, and Yong Liu.
\newblock Dmqr-rag: Diverse multi-query rewriting for rag.
\newblock \emph{arXiv preprint arXiv:2411.13154}, 2024{\natexlab{c}}.

\bibitem[{National Library of Medicine (US)}(2024)]{UMLS2024AA}
{National Library of Medicine (US)}.
\newblock {UMLS} knowledge sources, release 2024aa, 2024.
\newblock
  \url{http://www.nlm.nih.gov/research/umls/licensedcontent/umlsknowledgesources.html}
  (accessed 15 July 2024).

\bibitem[Chandak et~al.(2023)Chandak, Huang, and Zitnik]{Chandak2022PrimeKG}
Payal Chandak, Kexin Huang, and Marinka Zitnik.
\newblock Building a knowledge graph to enable precision medicine.
\newblock \emph{Scientific Data}, 10\penalty0 (1):\penalty0 67, 2023.
\newblock \doi{10.1038/s41597-023-01960-3}.

\bibitem[Sayers et~al.(2024)Sayers, Bolton, Brister, et~al.]{macleod2002pubmed}
Eric~W. Sayers, Evan~E. Bolton, J.~Rodney Brister, et~al.
\newblock Database resources of the national center for biotechnology
  information.
\newblock \emph{Nucleic Acids Research}, 52\penalty0 (D1):\penalty0 D33--D43,
  2024.
\newblock \doi{10.1093/nar/gkad1044}.

\bibitem[{StatPearls Publishing}(2026)]{statpearls}
{StatPearls Publishing}.
\newblock \emph{StatPearls [Internet]}.
\newblock StatPearls Publishing, Treasure Island, FL, 2026.
\newblock Available from: \url{https://www.ncbi.nlm.nih.gov/books/NBK430685/}
  (accessed 27 February 2026).

\bibitem[Jin et~al.(2021)Jin, Pan, Oufattole, Weng, Fang, and
  Szolovits]{jin2021medqa_usmle}
Di~Jin, Eileen Pan, Nassim Oufattole, Wei-Hung Weng, Hanyi Fang, and Peter
  Szolovits.
\newblock What disease does this patient have? a large-scale open domain
  question answering dataset from medical exams.
\newblock \emph{Applied Sciences}, 11\penalty0 (14):\penalty0 6421, 2021.
\newblock \doi{10.3390/app11146421}.

\bibitem[Xiong et~al.(2024)Xiong, Jin, Lu, et~al.]{xiong2024medrag}
G.~Xiong, Q.~Jin, Z.~Lu, et~al.
\newblock Benchmarking retrieval-augmented generation for medicine.
\newblock In \emph{Findings of the Association for Computational Linguistics:
  ACL 2024}, pages 6233--6251, 2024.

\bibitem[Gao et~al.(2026)Gao, Li, Chang, Du, Ye, Yeo, Xia, Guo, Zhang, Liu,
  et~al.]{gao2026multi}
Yuan Gao, Chunli Li, Wanxing Chang, Bai Du, Xianghua Ye, Yee~Hui Yeo, Yingda
  Xia, Heng Guo, Xiaoming Zhang, Wei Liu, et~al.
\newblock Multi-modal ai for opportunistic screening, staging and progression
  risk stratification of steatotic liver disease.
\newblock \emph{Nature Communications}, 17\penalty0 (1):\penalty0 1562, 2026.

\bibitem[Yu et~al.(2024)Yu, Wang, Yan, Li, Guo, Zhang, Shen, Wang, Ding, Lu,
  et~al.]{yu2025effective}
Qinji Yu, Yirui Wang, Ke~Yan, Haoshen Li, Dazhou Guo, Li~Zhang, Na~Shen, Qifeng
  Wang, Xiaowei Ding, Le~Lu, et~al.
\newblock Effective lymph nodes detection in ct scans using location debiased
  query selection and contrastive query representation in transformer.
\newblock In \emph{European Conference on Computer Vision}, pages 180--198.
  Springer, 2024.

\bibitem[Cohen et~al.(2020)Cohen, Hashir, Brooks, and
  Bertrand]{cohen2020limits}
Joseph~Paul Cohen, Mohammad Hashir, Rupert Brooks, and Hadrien Bertrand.
\newblock On the limits of cross-domain generalization in automated x-ray
  prediction.
\newblock In \emph{Medical Imaging with Deep Learning (MIDL)}, volume 121 of
  \emph{Proceedings of Machine Learning Research}, pages 136--155. PMLR, 2020.
\newblock URL \url{https://proceedings.mlr.press/v121/cohen20a.html}.

\bibitem[Yang et~al.(2025{\natexlab{b}})Yang, Xu, Zhang, Wang, Kalra, and
  Yan]{yang2025chest}
Zefan Yang, Xuanang Xu, Jiajin Zhang, Ge~Wang, Mannudeep~K. Kalra, and Pingkun
  Yan.
\newblock Chest x-ray foundation model with global and local representations
  integration.
\newblock \emph{IEEE Transactions on Medical Imaging}, 44\penalty0
  (12):\penalty0 4787--4799, 2025{\natexlab{b}}.
\newblock \doi{10.1109/TMI.2025.3581907}.

\bibitem[Chen et~al.(2024{\natexlab{b}})Chen, Varma, Delbrouck, Paschali,
  Blankemeier, Van~Veen, Valanarasu, Youssef, Cohen, Reis, Tsai, Johnston,
  Olsen, Abraham, Gatidis, Chaudhari, and Langlotz]{chexagent-2024}
Zhihong Chen, Maya Varma, Jean-Benoit Delbrouck, Magdalini Paschali, Louis
  Blankemeier, Dave Van~Veen, Jeya Maria~Jose Valanarasu, Alaa Youssef,
  Joseph~Paul Cohen, Eduardo~Pontes Reis, Emily~B. Tsai, Andrew Johnston,
  Cameron Olsen, Tanishq~Mathew Abraham, Sergios Gatidis, Akshay~S. Chaudhari,
  and Curtis Langlotz.
\newblock Chexagent: Towards a foundation model for chest x-ray interpretation.
\newblock In \emph{AAAI 2024 Spring Symposium on Clinical Foundation Models},
  2024{\natexlab{b}}.
\newblock URL \url{https://arxiv.org/abs/2401.12208}.

\bibitem[Lin et~al.(2023)Lin, Zhao, Zhang, Wu, Zhang, Wang, and
  Xie]{lin2023pmc}
Weixiong Lin, Ziheng Zhao, Xiaoman Zhang, Chaoyi Wu, Ya~Zhang, Yanfeng Wang,
  and Weidi Xie.
\newblock Pmc-clip: Contrastive language-image pre-training using biomedical
  documents.
\newblock In \emph{International Conference on Medical Image Computing and
  Computer-Assisted Intervention}, pages 525--536. Springer, 2023.

\bibitem[Pelka et~al.(2018)Pelka, Koitka, R{\"u}ckert, Nensa, and
  Friedrich]{pelka2018roco}
Obioma Pelka, Svenja Koitka, Johannes R{\"u}ckert, Felix Nensa, and
  Christoph~M. Friedrich.
\newblock Radiology objects in context (roco): A multimodal image dataset.
\newblock In \emph{Large-Scale Annotation of Biomedical Data and Expert Label
  Synthesis (LABELS) 2018, held in conjunction with MICCAI 2018}, volume 11043
  of \emph{Lecture Notes in Computer Science}, pages 180--189. Springer, 2018.
\newblock \doi{10.1007/978-3-030-01364-6_20}.

\bibitem[R{\"u}ckert et~al.(2024)R{\"u}ckert, Bloch, Br{\"u}ngel,
  Idrissi-Yaghir, Sch{\"a}fer, Schmidt, Koitka, Pelka, Abacha, G.~Seco~de
  Herrera, et~al.]{ruckert2024rocov2}
Johannes R{\"u}ckert, Louise Bloch, Raphael Br{\"u}ngel, Ahmad Idrissi-Yaghir,
  Henning Sch{\"a}fer, Cynthia~S Schmidt, Sven Koitka, Obioma Pelka, Asma~Ben
  Abacha, Alba G.~Seco~de Herrera, et~al.
\newblock Rocov2: Radiology objects in context version 2, an updated multimodal
  image dataset.
\newblock \emph{Scientific Data}, 11\penalty0 (1):\penalty0 688, 2024.

\bibitem[Chen et~al.(2024{\natexlab{c}})Chen, Gui, Ouyang, Gao, Chen, Chen,
  Wang, Cai, Ji, Wan, and Wang]{chen2024pubmedvision}
Junying Chen, Chi Gui, Ruyi Ouyang, Anningzhe Gao, Shunian Chen, Guiming~Hardy
  Chen, Xidong Wang, Zhenyang Cai, Ke~Ji, Xiang Wan, and Benyou Wang.
\newblock Towards injecting medical visual knowledge into multimodal llms at
  scale.
\newblock In \emph{Proceedings of the 2024 Conference on Empirical Methods in
  Natural Language Processing (EMNLP)}, pages 7346--7370. Association for
  Computational Linguistics, November 2024{\natexlab{c}}.
\newblock \doi{10.18653/v1/2024.emnlp-main.418}.

\bibitem[Johnson et~al.(2019)Johnson, Pollard, Berkowitz, Greenbaum, Lungren,
  Deng, Mark, and Horng]{johnson2019mimiccxr}
Alistair~EW Johnson, Tom~J Pollard, Seth~J Berkowitz, Nathaniel~R Greenbaum,
  Matthew~P Lungren, Chih-ying Deng, Roger~G Mark, and Steven Horng.
\newblock Mimic-cxr, a de-identified publicly available database of chest
  radiographs with free-text reports.
\newblock \emph{Scientific data}, 6\penalty0 (1):\penalty0 317, 2019.

\bibitem[Demner-Fushman et~al.(2016{\natexlab{a}})Demner-Fushman, Kohli,
  Rosenman, Shooshan, Rodriguez, Antani, Thoma, and
  McDonald]{demnerfushman2016openi}
Dina Demner-Fushman, Marc~D Kohli, Marc~B Rosenman, Sonya~E Shooshan, Laritza
  Rodriguez, Sameer Antani, George~R Thoma, and Clement~J McDonald.
\newblock Preparing a collection of radiology examinations for distribution and
  retrieval.
\newblock \emph{Journal of the American Medical Informatics Association},
  23\penalty0 (2):\penalty0 304--310, 2016{\natexlab{a}}.
\newblock \doi{10.1093/jamia/ocv080}.

\bibitem[Subramanian et~al.(2020)Subramanian, Wang, Bogin, Mehta, Van~Zuylen,
  Parasa, Singh, Gardner, and Hajishirzi]{subramanian2020medicat}
Sanjay Subramanian, Lucy~Lu Wang, Ben Bogin, Sachin Mehta, Madeleine
  Van~Zuylen, Sravanthi Parasa, Sameer Singh, Matt Gardner, and Hannaneh
  Hajishirzi.
\newblock Medicat: A dataset of medical images, captions, and textual
  references.
\newblock In \emph{Findings of the Association for Computational Linguistics:
  EMNLP 2020}, pages 2112--2120, 2020.

\bibitem[Siragusa et~al.(2025)Siragusa, Contino, Ciura, Alicata, and
  Pirrone]{siragusa2025medpix20}
Irene Siragusa, Salvatore Contino, Massimo~La Ciura, Rosario Alicata, and
  Roberto Pirrone.
\newblock Medpix 2.0: a comprehensive multimodal biomedical data set for
  advanced ai applications with retrieval augmented generation and knowledge
  graphs.
\newblock \emph{Data Science and Engineering}, pages 1--17, 2025.

\bibitem[Lozano et~al.(2025)Lozano, Sun, Burgess, Chen, Nirschl, Gu, Lopez,
  Aklilu, Katzer, Chiu, Rau, Wang, Zhang, Song, Tibshirani, and
  Yeung-Levy]{lozano2025biomedicaopenbiomedicalimagecaption}
Alejandro Lozano, Min~Woo Sun, James Burgess, Liangyu Chen, Jeffrey~J. Nirschl,
  Jeffrey Gu, Ivan Lopez, Josiah Aklilu, Austin~Wolfgang Katzer, Collin Chiu,
  Anita Rau, Xiaohan Wang, Yuhui Zhang, Alfred~Seunghoon Song, Robert
  Tibshirani, and Serena Yeung-Levy.
\newblock Biomedica: An open biomedical image-caption archive, dataset, and
  vision-language models derived from scientific literature.
\newblock In \emph{Proceedings of the IEEE/CVF Conference on Computer Vision
  and Pattern Recognition (CVPR)}, pages 19724--19735, 2025.
\newblock \doi{10.1109/CVPR52734.2025.01837}.

\bibitem[Liu et~al.(2024{\natexlab{b}})Liu, Li, Li, and
  Lee]{liu2023visualinstructiontuning}
Haotian Liu, Chunyuan Li, Yuheng Li, and Yong~Jae Lee.
\newblock Improved baselines with visual instruction tuning.
\newblock In \emph{Proceedings of the IEEE/CVF Conference on Computer Vision
  and Pattern Recognition (CVPR)}, pages 26296--26306, June 2024{\natexlab{b}}.

\bibitem[Deitke et~al.(2025)Deitke, Clark, Lee, Tripathi, Yang, Park, Salehi,
  Muennighoff, Lo, Soldaini, et~al.]{deitke2024molmo}
Matt Deitke, Christopher Clark, Sangho Lee, Rohun Tripathi, Yue Yang, Jae~Sung
  Park, Mohammadreza Salehi, Niklas Muennighoff, Kyle Lo, Luca Soldaini, et~al.
\newblock Molmo and pixmo: Open weights and open data for state-of-the-art
  vision-language models.
\newblock In \emph{Proceedings of the Computer Vision and Pattern Recognition
  Conference}, pages 91--104, 2025.

\bibitem[Li et~al.(2023{\natexlab{b}})Li, Wong, Zhang, Usuyama, Liu, Yang,
  Naumann, Poon, and Gao]{li2023llavamed}
Chunyuan Li, Cliff Wong, Sheng Zhang, Naoto Usuyama, Haotian Liu, Jianwei Yang,
  Tristan Naumann, Hoifung Poon, and Jianfeng Gao.
\newblock Llava-med: Training a large language-and-vision assistant for
  biomedicine in one day.
\newblock \emph{Advances in Neural Information Processing Systems},
  36:\penalty0 28541--28564, 2023{\natexlab{b}}.

\bibitem[Seyfioglu et~al.(2024)Seyfioglu, Ikezogwo, Ghezloo, Krishna, and
  Shapiro]{Seyfioglu_2024_CVPR}
Mehmet~Saygin Seyfioglu, Wisdom~O Ikezogwo, Fatemeh Ghezloo, Ranjay Krishna,
  and Linda Shapiro.
\newblock Quilt-llava: Visual instruction tuning by extracting localized
  narratives from open-source histopathology videos.
\newblock In \emph{Proceedings of the IEEE/CVF Conference on Computer Vision
  and Pattern Recognition}, pages 13183--13192, 2024.

\bibitem[Gautam et~al.(2024)Gautam, Stor{\aa}s, Midoglu, Hicks, Thambawita,
  Halvorsen, and Riegler]{gautam2024kvasirvqa}
Sushant Gautam, Andrea Stor{\aa}s, Cise Midoglu, Steven~A. Hicks, Vajira
  Thambawita, P{\aa}l Halvorsen, and Michael~A. Riegler.
\newblock Kvasir-vqa: A text-image pair gi tract dataset.
\newblock In \emph{Proceedings of the 2nd International Workshop on
  Vision-Language Models for Biomedical Applications (VLM4Bio '24)}, pages
  15--24, New York, NY, USA, 2024. Association for Computing Machinery.
\newblock ISBN 979-8-4007-0435-2.
\newblock \doi{10.1145/3689096.3689458}.

\bibitem[Bae et~al.(2024)Bae, Kyung, Ryu, et~al.]{bae2024mimicext}
Seongsu Bae, Daeun Kyung, Jaehee Ryu, et~al.
\newblock {MIMIC-Ext-MIMIC-CXR-VQA}: A complex, diverse, and large-scale visual
  question answering dataset for chest x-ray images, 2024.
\newblock PhysioNet,
  \url{https://physionet.org/content/mimic-ext-mimic-cxr-vqa/}.

\bibitem[He et~al.(2021)He, Cai, Wei, Zhang, Mou, Xing, and Xie]{he2020pathvqa}
Xuehai He, Zhuo Cai, Wenlan Wei, Yichen Zhang, Luntian Mou, Eric Xing, and
  Pengtao Xie.
\newblock Towards visual question answering on pathology images.
\newblock In \emph{Proceedings of the 59th Annual Meeting of the Association
  for Computational Linguistics and the 11th International Joint Conference on
  Natural Language Processing (Volume 2: Short Papers)}, pages 708--718.
  Association for Computational Linguistics, 2021.
\newblock \doi{10.18653/v1/2021.acl-short.90}.
\newblock URL \url{https://aclanthology.org/2021.acl-short.90/}.

\bibitem[Zhang et~al.(2024{\natexlab{b}})Zhang, Wu, Zhao, Lin, Zhang, Wang, and
  Xie]{zhang2023pmcvqa}
Xiaoman Zhang, Chaoyi Wu, Ziheng Zhao, Weixiong Lin, Ya~Zhang, Yanfeng Wang,
  and Weidi Xie.
\newblock Development of a large-scale medical visual question-answering
  dataset.
\newblock \emph{Communications Medicine}, 4\penalty0 (1):\penalty0 23,
  2024{\natexlab{b}}.
\newblock \doi{10.1038/s43856-024-00709-2}.

\bibitem[Ben~Abacha et~al.(2019)Ben~Abacha, Hasan, Datla, Liu, Demner-Fushman,
  and M{\"u}ller]{benabacha2019vqamed}
Asma Ben~Abacha, Sadid~A. Hasan, Vivek~V. Datla, Joey Liu, Dina Demner-Fushman,
  and Henning M{\"u}ller.
\newblock {VQA-Med}: Overview of the medical visual question answering task at
  {ImageCLEF} 2019.
\newblock In \emph{CLEF 2019 Working Notes, CEUR Workshop Proceedings}, volume
  2380, 2019.
\newblock URL \url{https://ceur-ws.org/Vol-2380/paper_272.pdf}.

\bibitem[Lau et~al.(2018)Lau, Gayen, Ben~Abacha, and
  Demner-Fushman]{lau2018vqarad}
Jason~J Lau, Soumya Gayen, Asma Ben~Abacha, and Dina Demner-Fushman.
\newblock A dataset of clinically generated visual questions and answers about
  radiology images.
\newblock \emph{Scientific data}, 5\penalty0 (1):\penalty0 1--10, 2018.
\newblock URL \url{https://doi.org/10.1038/sdata.2018.251}.

\bibitem[Su et~al.(2025)Su, Li, Liu, Ma, Ning, Tang, Ju, Ye, Chen, Hu,
  et~al.]{gmai_reasoning10k_arxiv}
Yanzhou Su, Tianbin Li, Jiyao Liu, Chenglong Ma, Junzhi Ning, Cheng Tang, Sibo
  Ju, Jin Ye, Pengcheng Chen, Ming Hu, et~al.
\newblock Gmai-vl-r1: Harnessing reinforcement learning for multimodal medical
  reasoning.
\newblock \emph{arXiv preprint arXiv:2504.01886}, 2025.

\bibitem[Chen et~al.(2024{\natexlab{d}})Chen, Chen, Zhang, Chen, Wu, Zhang,
  Chen, Li, Wan, and Wang]{chen2024allava}
Guiming~Hardy Chen, Shunian Chen, Ruifei Zhang, Junying Chen, Xiangbo Wu, Zhiyi
  Zhang, Zhihong Chen, Jianquan Li, Xiang Wan, and Benyou Wang.
\newblock Allava: Harnessing gpt4v-synthesized data for lite vision-language
  models.
\newblock \emph{arXiv preprint arXiv:2402.11684}, 2024{\natexlab{d}}.

\bibitem[Ben~Abacha and Demner-Fushman(2019)]{benabacha2019medquad}
Asma Ben~Abacha and Dina Demner-Fushman.
\newblock A question-entailment approach to question answering.
\newblock \emph{BMC Bioinformatics}, 20\penalty0 (1):\penalty0 511, 2019.
\newblock \doi{10.1186/s12859-019-3119-4}.

\bibitem[Chen et~al.(2024{\natexlab{e}})Chen, Cai, Ji, Wang, Liu, Wang, Hou,
  and Wang]{chen2024huatuogpto1}
Junying Chen, Zhenyang Cai, Ke~Ji, Xidong Wang, Wanlong Liu, Rongsheng Wang,
  Jianye Hou, and Benyou Wang.
\newblock Huatuogpt-o1, towards medical complex reasoning with llms.
\newblock \emph{arXiv preprint arXiv:2412.18925}, 2024{\natexlab{e}}.

\bibitem[Wang et~al.(2024{\natexlab{a}})Wang, Chen, Chen, Wang, Zhen, Zhang,
  Wu, Hu, Gao, Wan, et~al.]{wang2024apollo}
Xidong Wang, Nuo Chen, Junyin Chen, Yidong Wang, Guorui Zhen, Chunxian Zhang,
  Xiangbo Wu, Yan Hu, Anningzhe Gao, Xiang Wan, et~al.
\newblock Apollo: A lightweight multilingual medical llm towards democratizing
  medical ai to 6b people.
\newblock \emph{arXiv preprint arXiv:2403.03640}, 2024{\natexlab{a}}.

\bibitem[Zhang et~al.(2023{\natexlab{b}})Zhang, Tian, Yang, Chen, Li, and
  Petzold]{zhang2025alpacare}
Xinlu Zhang, Chenxin Tian, Xianjun Yang, Lichang Chen, Zekun Li, and Linda~Ruth
  Petzold.
\newblock Alpacare: Instruction-tuned large language models for medical
  application.
\newblock \emph{arXiv preprint arXiv:2310.14558}, 2023{\natexlab{b}}.

\bibitem[Li et~al.(2023{\natexlab{c}})Li, Li, Zhang, Dan, Jiang, and
  Zhang]{hf_healthcaremagic_100k}
Yunxiang Li, Zihan Li, Kai Zhang, Ruilong Dan, Steve Jiang, and You Zhang.
\newblock Chatdoctor: A medical chat model fine-tuned on a large language model
  meta-{AI} ({LLaMA}) using medical domain knowledge.
\newblock \emph{Cureus}, 15\penalty0 (6):\penalty0 e40895, 2023{\natexlab{c}}.
\newblock \doi{10.7759/cureus.40895}.

\bibitem[Wu et~al.(2024)Wu, Lin, Zhang, Zhang, Xie, and Wang]{wu2024pmc}
Chaoyi Wu, Weixiong Lin, Xiaoman Zhang, Ya~Zhang, Weidi Xie, and Yanfeng Wang.
\newblock Pmc-llama: toward building open-source language models for medicine.
\newblock \emph{Journal of the American Medical Informatics Association},
  31\penalty0 (9):\penalty0 1833--1843, 2024.

\bibitem[Chen et~al.(2023)Chen, Wang, Gao, Jiang, Chen, Zhang,
  et~al.]{chen2023huatuogptii}
Junying Chen, Xidong Wang, Anningzhe Gao, Feng Jiang, Shunian Chen, Hongbo
  Zhang, et~al.
\newblock Huatuogpt-ii, one-stage training for medical adaption of llms.
\newblock \emph{arXiv preprint arXiv:2311.09774}, 2023.

\bibitem[Wu et~al.(2025{\natexlab{b}})Wu, Deng, Li, Liu, Mi, Peng, Xu, Liu,
  Cho, Choi, et~al.]{wu2025medreason}
Juncheng Wu, Wenlong Deng, Xingxuan Li, Sheng Liu, Taomian Mi, Yifan Peng,
  Ziyang Xu, Yi~Liu, Hyunjin Cho, Chang-In Choi, et~al.
\newblock Medreason: Eliciting factual medical reasoning steps in llms via
  knowledge graphs.
\newblock \emph{arXiv preprint arXiv:2504.00993}, 2025{\natexlab{b}}.

\bibitem[Wei(2025)]{medthoughts2025}
Hao Wei.
\newblock Medthoughts-8k: A medical reasoning dataset distilled from
  deepseek-r1.
\newblock \url{https://huggingface.co/datasets/hw-hwei/MedThoughts-8K}, 2025.

\bibitem[Wang et~al.(2024{\natexlab{b}})Wang, Chen, Dingjie, Zhiyi, Chen, Xiao,
  Chen, Jiang, Li, Wan, Wang, and Li]{hf_openbook_cmb_exam}
Xidong Wang, Guiming Chen, Song Dingjie, Zhang Zhiyi, Zhihong Chen, Qingying
  Xiao, Junying Chen, Feng Jiang, Jianquan Li, Xiang Wan, Benyou Wang, and
  Haizhou Li.
\newblock {CMB}: A comprehensive medical benchmark in {C}hinese.
\newblock In \emph{Proceedings of the 2024 Conference of the North American
  Chapter of the Association for Computational Linguistics: Human Language
  Technologies (Volume 1: Long Papers)}, pages 6184--6205, Mexico City, Mexico,
  June 2024{\natexlab{b}}. Association for Computational Linguistics.
\newblock \doi{10.18653/v1/2024.naacl-long.343}.
\newblock URL \url{https://aclanthology.org/2024.naacl-long.343/}.

\bibitem[Zhang et~al.(2018)Zhang, Zhang, Wang, Guo, and Liu]{hf_cmedqa_v20}
Sheng Zhang, Xin Zhang, Hui Wang, Lixiang Guo, and Shanshan Liu.
\newblock Multi-scale attentive interaction networks for chinese medical
  question answer selection.
\newblock \emph{IEEE Access}, 6:\penalty0 74061--74071, 2018.
\newblock \doi{10.1109/ACCESS.2018.2883637}.

\bibitem[Wang et~al.(2025{\natexlab{c}})Wang, Gao, Yang,
  et~al.]{wang2025citrus}
Guoxin Wang, Minyu Gao, Shuai Yang, et~al.
\newblock Citrus: Leveraging expert cognitive pathways in a medical language
  model for advanced medical decision support.
\newblock \emph{arXiv preprint arXiv:2502.18274}, 2025{\natexlab{c}}.

\bibitem[Kim et~al.(2025)Kim, Hwang, Lee, Park, Kim,
  et~al.]{kim2024medbooks18cot}
Hyunjae Kim, Hyeon Hwang, Jiwoo Lee, Sihyeon Park, Dain Kim, et~al.
\newblock Small language models learn enhanced reasoning skills from medical
  textbooks.
\newblock \emph{npj Digital Medicine}, 8\penalty0 (1):\penalty0 240, 2025.
\newblock \doi{10.1038/s41746-025-01653-8}.

\bibitem[Teknium(2023)]{openhermes25}
Teknium.
\newblock Openhermes 2.5: An open dataset of synthetic data for generalist llm
  assistants, 2023.
\newblock Hugging Face Datasets,
  \url{https://huggingface.co/datasets/teknium/OpenHermes-2.5}.

\bibitem[Zhang et~al.(2024{\natexlab{c}})Zhang, Wu, Zhao, Lei, Zhang, Wang, and
  Xie]{zhang2024radgenomechestct}
Xiaoman Zhang, Chaoyi Wu, Ziheng Zhao, Jiayu Lei, Ya~Zhang, Yanfeng Wang, and
  Weidi Xie.
\newblock Radgenome-chest ct: A grounded vision-language dataset for chest ct
  analysis.
\newblock \emph{arXiv preprint arXiv:2404.16754}, 2024{\natexlab{c}}.

\bibitem[amo(2024)]{amosmm2024_zenodo}
{AMOS-MM}: Abdominal multimodal analysis challenge dataset, 2024.
\newblock Zenodo, \url{https://zenodo.org/doi/10.5281/zenodo.10992154}
  (accessed 23 February 2026).

\bibitem[Gai et~al.(2025{\natexlab{b}})Gai, Liu, Li, Meng, Wu, and
  Liu]{gai2025_3drad}
Xiaotang Gai, Jiaxiang Liu, Yichen Li, Zijie Meng, Jian Wu, and Zuozhu Liu.
\newblock 3d-rad: A comprehensive 3d radiology med-vqa dataset with
  multi-temporal analysis and diverse diagnostic tasks.
\newblock In \emph{Advances in Neural Information Processing Systems},
  volume~38, 2025{\natexlab{b}}.

\bibitem[Yu et~al.(2025)Yu, Wang, Wu, Xie, and Zhou]{yu2025MedFrameQA}
Suhao Yu, Haojin Wang, Juncheng Wu, Cihang Xie, and Yuyin Zhou.
\newblock Medframeqa: A multi-image medical vqa benchmark for clinical
  reasoning.
\newblock \emph{arXiv preprint arXiv:2505.16964}, 2025.

\bibitem[Ye et~al.(2024)Ye, Wang, Li, Deng, Li, Li, Duan, Huang, Su, Wang,
  et~al.]{ye2024Gmai-mmbench}
Jin Ye, Guoan Wang, Yanjun Li, Zhongying Deng, Wei Li, Tianbin Li, Haodong
  Duan, Ziyan Huang, Yanzhou Su, Benyou Wang, et~al.
\newblock Gmai-mmbench: A comprehensive multimodal evaluation benchmark towards
  general medical ai.
\newblock \emph{Advances in Neural Information Processing Systems},
  37:\penalty0 94327--94427, 2024.

\bibitem[Pal et~al.(2022)Pal, Umapathi, and Sankarasubbu]{pal2022medmcqa}
Ankit Pal, Logesh~Kumar Umapathi, and Malaikannan Sankarasubbu.
\newblock Medmcqa: A large-scale multi-subject multi-choice dataset for medical
  domain question answering.
\newblock In Gerardo Flores, George~H Chen, Tom Pollard, Joyce~C Ho, and
  Tristan Naumann, editors, \emph{Proceedings of the Conference on Health,
  Inference, and Learning}, volume 174 of \emph{Proceedings of Machine Learning
  Research}, pages 248--260. PMLR, Apr 2022.
\newblock URL \url{https://proceedings.mlr.press/v174/pal22a.html}.

\bibitem[Chen et~al.(2025)Chen, Fang, Singla, and Dredze]{chen2025medbullets}
Hanjie Chen, Zhouxiang Fang, Yash Singla, and Mark Dredze.
\newblock Benchmarking large language models on answering and explaining
  challenging medical questions.
\newblock In \emph{Proceedings of the 2025 Conference of the Nations of the
  Americas Chapter of the Association for Computational Linguistics: Human
  Language Technologies (Volume 1: Long Papers)}, pages 3563--3599,
  Albuquerque, New Mexico, April 2025. Association for Computational
  Linguistics.
\newblock URL \url{https://aclanthology.org/2025.naacl-long.182/}.

\bibitem[Du et~al.(2025)Du, Yao, Ma, Wang, Zheng, Zhu, Liu, Liang, Jin, Wei,
  Zheng, Deng, Guo, Jia, Jiang, Liao, Li, Li, Li, Li, Li, Ma, Ni, Que, Wang,
  Wen, Wu, Xing, Xu, Yang, Wang, Zhou, et~al.]{pteam2025supergpqa}
Xinrun Du, Yifan Yao, Kaijing Ma, Bingli Wang, Tianyu Zheng, King Zhu, Minghao
  Liu, Yiming Liang, Xiaolong Jin, Zhenlin Wei, Chujie Zheng, Kaixin Deng,
  Shuyue Guo, Shian Jia, Sichao Jiang, Yiyan Liao, Rui Li, Qinrui Li, Sirun Li,
  Yizhi Li, Yunwen Li, Dehua Ma, Yuansheng Ni, Haoran Que, Qiyao Wang, Zhoufutu
  Wen, Siwei Wu, Tianshun Xing, Ming Xu, Zhenzhu Yang, Zekun~Moore Wang,
  Junting Zhou, et~al.
\newblock Supergpqa: Scaling llm evaluation across 285 graduate disciplines.
\newblock In \emph{Advances in Neural Information Processing Systems 38
  (NeurIPS), Datasets and Benchmarks Track}, 2025.
\newblock URL \url{https://openreview.net/forum?id=6WgflzYQpf}.

\bibitem[Demner-Fushman et~al.(2016{\natexlab{b}})Demner-Fushman, Kohli,
  Rosenman, Shooshan, Rodriguez, Antani, Thoma, and McDonald]{demner2016IUXRay}
Dina Demner-Fushman, Marc~D Kohli, Marc~B Rosenman, Sonya~E Shooshan, Laritza
  Rodriguez, Sameer Antani, George~R Thoma, and Clement~J McDonald.
\newblock Preparing a collection of radiology examinations for distribution and
  retrieval.
\newblock \emph{Journal of the American Medical Informatics Association},
  23\penalty0 (2):\penalty0 304--310, 2016{\natexlab{b}}.
\newblock URL \url{https://doi.org/10.1093/jamia/ocv080}.

\bibitem[Selvaraju et~al.(2020)Selvaraju, Cogswell, Das, Vedantam, Parikh, and
  Batra]{selvaraju2020grad_cam}
Ramprasaath~R Selvaraju, Michael Cogswell, Abhishek Das, Ramakrishna Vedantam,
  Devi Parikh, and Dhruv Batra.
\newblock Grad-cam: visual explanations from deep networks via gradient-based
  localization.
\newblock \emph{International journal of computer vision}, 128\penalty0
  (2):\penalty0 336--359, 2020.

\end{thebibliography}
